\documentclass{article}

\usepackage{jmlr2e_custom}
\usepackage{natbib}
\usepackage{color}

\usepackage{changepage}
\usepackage{graphicx}
\usepackage{subcaption}
\usepackage{lscape}
\usepackage{verbatim}
\usepackage{afterpage}
\usepackage{multirow}
\usepackage{enumerate}
\usepackage{adjustbox}
\usepackage{amssymb, amsmath}
\usepackage{tikz}
\usepackage[super]{nth}
\usepackage{hyperref} 
\usetikzlibrary{shapes.misc}

\newcommand*\circled[1]{\tikz[baseline]{
  \node[
    shape=rounded rectangle,
    rounded rectangle arc length=180, 
    draw, 
    inner sep=+.223em,
    text depth=+.1ex,
    anchor=base
    ] (char) {\scriptsize{#1}};}}

\newcommand*\circledtbl[1]{\tikz[baseline]{
  \node[
    shape=rounded rectangle,
    rounded rectangle arc length=180, 
    draw, 
    inner sep=+.223em,
    text depth=+.1ex,
    ] (char) {\scriptsize{#1}};}}

\newcommand*\circledtext[1]{\tikz[baseline=(char.base)]{
  \node[
    shape=rounded rectangle,
    rounded rectangle arc length=180, 
    draw, 
    inner sep=+.333em,
    text depth=+.1ex,
    anchor=base
   ] (char) {\scriptsize{#1}};}}

\newcommand*\circledone[1]{\tikz[baseline]{
  \node[
    shape=rounded rectangle,
    rounded rectangle arc length=180, 
    draw, 
    inner sep=+.223em,
    text depth=+.1ex,
    anchor=base
    ] (char) {\scriptsize{#1}};}}

\definecolor{darkblue}{rgb}{0.1,0.1,0.35}

\hypersetup{colorlinks,breaklinks,
            linkcolor=darkblue,urlcolor=darkblue,
            anchorcolor=darkblue,citecolor=darkblue}

\usepackage{array}
\newcolumntype{L}[1]{>{\raggedright\let\newline\\\arraybackslash\hspace{0pt}}p{#1}}
\newcolumntype{C}[1]{>{\centering\let\newline\\\arraybackslash\hspace{0pt}}m{#1}}
\newcolumntype{R}[1]{>{\raggedleft\let\newline\\\arraybackslash\hspace{0pt}}m{#1}}

\newcolumntype{T}[2]{%
    >{\adjustbox{angle=#1,lap=\width-(#2)}\bgroup}%
    l%
    <{\egroup}%
}
\newcommand*\rot{\multicolumn{1}{T{25}{1em}}}

\ShortHeadings{ML for Movement Generation}{O. Alemi and P. Pasquier}
\begin{document}

\title{Machine Learning for Data-Driven Movement Generation: a Review of the State of the Art}

\author{\name Omid Alemi \email oalemi@sfu.ca \\
       \addr School of Interactive Arts and Technology\\Simon Fraser University\\
       Vancouver, Canada
       \AND
       \name Philippe Pasquier\email pasquier@sfu.ca  \\
       \addr School of Interactive Arts and Technology\\Simon Fraser University\\
       Vancouver, Canada
}

\editor {}

\maketitle


\begin{abstract}
The rise of non-linear and interactive media such as video games has increased the need for automatic  movement animation generation.
In this survey,  we review and analyze different aspects of  building  automatic movement generation systems using machine learning techniques and motion capture data. We cover topics such as high-level movement characterization, training data,  features representation,   machine learning models, and evaluation methods. 
We conclude by presenting a discussion of the reviewed literature and outlining the research gaps and remaining challenges for future work.

\end{abstract}

\section{Introduction}
\label{sec:intro}

The shift from linear media (e.g., music, books, movies, etc.) to non-linear media (e.g., video games, interactive installations, etc.), along with the proliferation of interactive storytelling mediums such as web and virtual and augmented reality has resulted in an increase in the need for creating diverse content, ranging from sound and music to graphical textures and virtual agent animation.

In particular, the dynamic and interactive nature of non-linear applications lead to a need for the animation of anthropomorphic virtual agents with a wide range of behaviours, actions, expressions, and personalities. 
This increase in demand is changing the practice of creating movement animation, as the traditional methods are too costly and time consuming to be used in non-linear applications \citep{tomlinson_linear_2005,pejsa_state_2010}. 
As a result, a body of research around building automatic movement generation models has been growing over the past two decades.

Automatic movement generation can be applied to both physical agents (robots) and animated software agents. 
In this paper, we focus on animated software agents, although we note that the techniques used for both types of agents are not mutually exclusive and 
 similar models can be used in the motor controllers of physical robots and software agents, e.g., \citep{herzog_parametric_2008,Matsubara:bp,Kulic:2011hw} \circledone{13} \circledone{15} \circledone{19}\footnote{Numbers in \circledtext{num} refer to the items in the tables.}. 

There are three broad groups of approaches that are employed in the computational models of movement to generate new animation: physics-based, data-driven, and hybrid.

\textit{Physical simulations} are used to model and generate movement animation \citep{safonova_synthesizing_2004,Agrawal:2013bg}. 
Incorporating physical laws allows such models to create movements that are physically valid, are proportionate to the physical dimensions of the body, and react to the other forces in the environment such as gravity, friction, and external push or pull. This group includes models that use reinforcement learning techniques in which an agent interacts with an environment and iteratively learns how to move through a rewarding system that is enforced by the physical laws of the environment.
Despite its powerful properties, relying solely on physical simulation can be insufficient in producing natural looking movements and modelling the expressive variations of movements \citep{pejsa_state_2010,wei_physically_2011} \circledone{17}.

\textit{Data-driven animation techniques} use pre-recorded movements of real human actors.
The movement segments are concatenated, e.g., \citep{tanco_realistic_2000}, blended, e.g.,  \citep{kwon_motion_2005,hsu_style_2005}, or used as the training data of machine learning models, e.g.,  \citep{brand_style_2000,wang_multifactor_2007,taylor_factored_2009,tilmanne_stylistic_2012} \circledone{1} \circledone{9} \circledone{14} \circledone{23}. Compared to physics-based approaches, the resulting movements are more natural looking and expressive. However, they are susceptible to artifacts such as foot skating or lack of balance. In addition, while physics-based approaches allow for modelling a wide variety of  movement types and creatures, 
not every movement can be generated using most data-driven methods as most data-driven methods can only generate a movement if the dataset it is based on contains movement that is, in one way or another, similar to the desired movement. 
Machine learning and statistical approaches can overcome this limitation by learning a generalized model of movement to create movements that do not have an explicit example in their training dataset.

\textit{The hybrid approaches}, while less explored in the literature, combine both the physics and the data-driven approaches in an attempt to create movement animations that are simultaneously physically valid and natural looking, e.g.,  \citep{wei_physically_2011} \circledone{17}.  This integration of the laws of physics and learning from observations and experiences resembles has overlaps with the processes that are behind human motor control.
While moving, we humans (and animals) respond to the feedbacks about our physical environment through our perception. 
We learn movement through the neural plasticity property of the brain, in which  special regions of the brain change in ways that are determined by the personal experience, including movement \cite{Rosenbaum:2009ta}.  Therefore, a hybrid architecture that allows for learning movement from experiences, enforced by the physical laws of the environment  could lead to more powerful generative models than  purely physics-based or data-driven approaches. 

In the rest of this paper, we focus on data-driven techniques, and in particular on machine learning and statistical methods applied on motion capture data.
Compared to the other data-driven methods, statistical models are not confined to the variations that exist in the training data and can be used to learn a generalized space of movements, fill the missing data, or generate continuous streams of movements.
The potential for the application of using statistical models in movement animation generation can be demonstrated by the success of such models in generating patterns of data in other fields, such as speech synthesis, e.g., \citep{Zen:2012dz,Ling:2013dg}, computer music, e.g., \citep{Dubnov:2003hn,Schulze:2011jn,GEDMAS},  and visual textures, e.g., \citep{Kivinen:2012ut}.

Review studies have covered the physics-based and data-driven techniques for movement generation.
For example, \citet{Wang:2014eo} present a general overview of the field of 3D human movement editing and generation. 
\citet{geijtenbeek_interactive_2012}  provide an extensive survey of different components of physics-based models and review the literature. In another study, \citet{pejsa_state_2010} review the literature on data-driven methods for creating graph-like structures, motion planning, and parametric movement synthesis using interpolation techniques. \citet{Karg:2013fd} review the recognition and generation techniques in the domain of affect-expressive movements. 
While statistical movement generation has been actively researched over the past two decades, to the best of our knowledge, there is no study presenting a comprehensive survey of the literature on applying machine learning models on movement data for animation generation. 

In this paper, we identify the key goals, challenges, and gaps in the research on statistical movement generation. 
We first review the typical architecture of a statistical movement generation system and its components, and outline the goals, challenges, and the design choices that are involved, in Section~\ref{sec:movement}.
We summarize our findings on the characterization of movement in the literature and detail our framework in Section~\ref{sec:charac}.
We discuss recording,  processing, and representations of movement data in the literature, and review the publicly available movement databases in Section~\ref{sec:data}.
We survey the application of using machine learning techniques with motion capture data for learning and generating movement animation
 in Section~\ref{sec:synthesis}.  We look at the evaluation methods in Section~\ref{sec:eval}. We summarize our findings and provide a discussion of the gaps and remaining challenges in Section~\ref{sec:discussion}.   Finally, we present the conclusions in Section~\ref{sec:conc}.

\section{Background and Fundamentals}
\label{sec:movement}

This section presents the fundamentals of statistical movement generation. We first lay out the definitions and assumptions that we use throughout this paper. Next, we describe a typical architecture for capturing, learning, and generating human movement. We discuss the common themes and the research goals motivating the field,  followed by a description of the applications of movement generation.

\subsection{Definition of Key Concepts}
\begin{description}
\item[- \textbf{Skeleton:}] In modelling and animating full-body movement, it is common to use a skeleton to represent the body structure. 
Each body pose is described by a set of the rotations of the joints (or equivalently the bones), as well as the orientation and position of the agent in the global space (typically called the \textit{root}). The hierarchy of the joints and their rotations are constrained by a pre-defined skeleton structure.

\item[- \textbf{Posture / Pose:}] Posture or pose refer to a static state of the body,  described  by the positions or orientations of the body parts as a whole. Numerically, a pose is represented by a single frame of data.

\item[- \textbf{Pose Space:}]  We use the term pose space to refer to the space of all possible body poses. A pose is therefore a single point in this space.

\item[- \textbf{Gesture:}] Gesture is the movement of a subset of body parts, often performed to communicate  information \citep{lamb1965posture}. 

\item[- \textbf{Movement:}] By {movement} we refer to the animation of a full-body representation of an anthropomorphic skeleton through time.

\item[- \textbf{Motion:}] We make a distinction between human \textit{motion} and human \textit{movement}. While mostly used interchangeably in the literature, we use \textit{motion} to refer to the changes of the position of the  body as a whole rather than individual body parts. In contrast, we use human \textit{movement} to refer to the coordinated motion of individual body limbs .

\item[- \textbf{Movement Primitives:}] The notion of \textit{movement primitive} is used to represent basic segments of human movements that constitute longer movements \citep{Schaal:2003ic}.

\item[- \textbf{Factor Space:}] We use the term factor space to refer to a high-level space of movement descriptors, such as those describing actions and emotions.

\item[- \textbf{Agent:}] We use the term agent to refer to an abstract model of a mover. Although the agent can refer to a human, physical robot, or a software, throughout the paper we use agent to specifically describe software agents.

\item[- \textbf{Mover:}] Throughout this paper, we use the terms  \textit{actor}, \textit{mover}, \textit{dancer}, \textit{performer}, and \textit{subject}, interchangeably to refer to the person or agent moving.

\item[- \textbf{Personal Movement Signature/Style:}] An individual's distinguishable movement patterns that is influenced by a combination of factors  such as the individual's physical build and cultural background \citep{bartenieff_body_1980, Serlin_2007}.


\end{description}

\subsection{Research Directions}
\label{sec:move:goals}

We categorize the directions that the body of research on movement generation follows in three themes: (a) achieving believability, (b) controlling  and  manipulating the characteristics of the generated movements, and (c) supporting real-time and continuous generation.
Each of the themes bring challenges that justify the design and development of movement generation models.

\begin{enumerate}
[(a)]
\item\textit{Believability}:  Believability is one of the fundamental notions in virtual agent animation \citep{Lasseter:1987ks,Bates:1994ja,pejsa_state_2010,Mori:2012is}.    
Even non-movement-expert humans notice  the smallest details that make  movement look unnatural \citep{pejsa_state_2010,geijtenbeek_interactive_2012}. It is challenging to manually create a believable animation that looks appealing to the audience from scratch.   

The animation industry has employed motion capture technologies in order to record the movements of human actors.  The recordings preserve the realism and expressive details of the movement, and is used as the basis of animations \citep{Menache:2000ue}.  Data-driven movement generation methods also take advantage of motion capture data in order to create natural-looking animations \citep{pina_computer_2000}. The challenge facing such methods is the often unwanted noise or artifacts that are introduced as a result of the computational manipulations of the recorded data.  Generating natural-looking movements is therefore one of the intrinsic goals of data-driven movement generation techniques.

\begin{figure}[t]
\centering
\includegraphics[width=.5\linewidth]{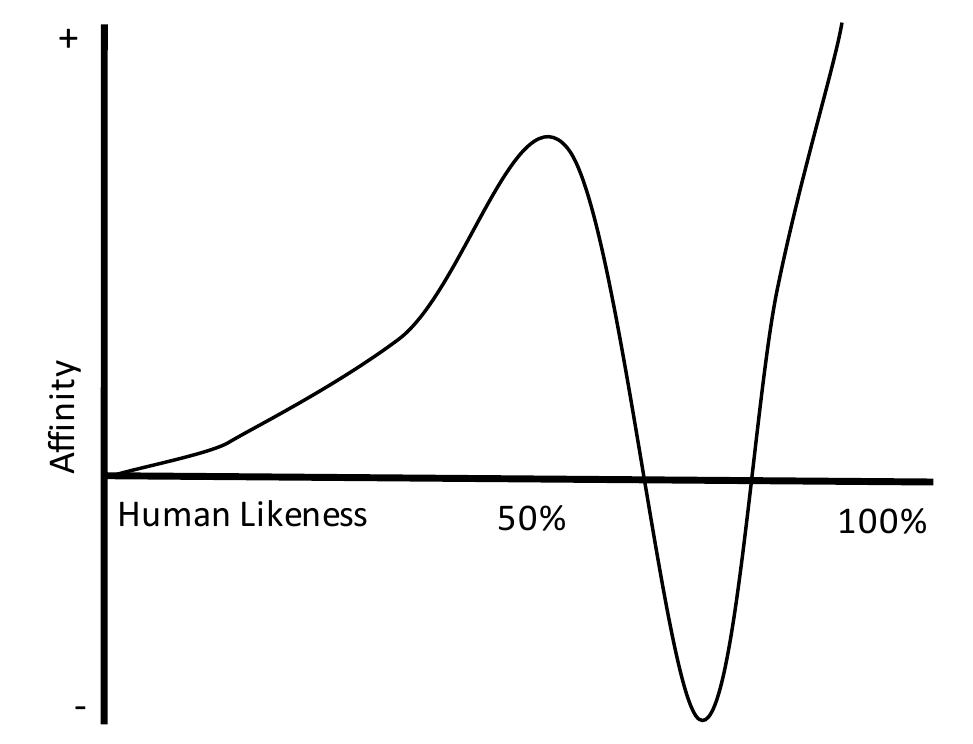}
\caption{The uncanny valley: the relationship between people's affinity towards human-like agents, as they approach human likeness.}
\label{fig:uncanny}
\end{figure}

There are two ingredients that are essential in achieving higher levels of believability in movements of an agent:

\begin{itemize}

\item[$-$]  \textit{Physical Validity}: As in reality, humans move in a physical world, their movements are constrained by the laws of physics. The movements that are generated  also need to follow the laws of physics and the biomechanics that are involved in human movement. Note that in humans, the notion of physics is implicit. The brain does not explicitly solve physical equations in order to produce the movement patterns. However, trough feedbacks from the physical environment, the brain learns to adapt to the laws of physics.   

As most of the data-driven methods approach the modelling problem without any prior assumptions about the mechanisms that produce the data, they are limited in guaranteeing obeying the laws of physics. Hybrid approaches, e.g., \citep{wei_physically_2011} \circledone{17}, combine data-driven methods with physical models to generate movements that are physically valid and look natural. 

The majority of the data-driven studies do not address the problem of physical validity. Although by learning the movements from real data and imitating their qualities, the generated movements uphold some of the physical properties of human movements, there is no constraint to enforce such rules and react to dynamic changes of forces.

\item[$-$]  \textit{Expressivity}: The expressiveness of the movements of agents plays an important role in their believability \citep{Bates:1994ja}. Movement is a form of non-verbal communication and conveys affective qualities that reflect the inner state of the agent. A generative model  of movement, should therefore be able to exhibit a variety of expressions, and allow controlling those expressions according to some high-level descriptors. In particular, the literature on data-driven models has addressed modelling expressive walks \citep{tilmanne_expressive_2010, Tilmanne:2014tx,Alemi:2015kg} \circledone{16} \circledone{26} \circledone{28} and hand movements \citep{Taubert:2011jp,Taubert:2012gl,Samadani:2012ki} \circledone{22} \circledone{22} \circledone{25}. Modelling the expressivity of movement is discussed in more details in Section~\ref{sec:charac}.
Expressivity is one of the advantages that data-driven models have over physics-based movement modelling approaches. 
\end{itemize}

\vspace{1mm}

Note that achieving a high level of realism may not always results in the agent being perceived as natural, which is a concept known as the \textit{uncanny valley} introduced by \citet{Mori:2012is}.  
As shown in Figure~\ref{fig:uncanny}, people's affinity to humanlike animated agents or robots increases as the similarities to real humans increases, but it abruptly diminishes as the similarities reach to near humanlike, but fail to reach too close. 
With respect to data-driven movement generation techniques, one could argue that achieving the same level of realism as the original recorded data is a satisfying criteria for evaluating the naturalness of the movement.
We discuss the evaluation methods of movement generation systems in Sections~\ref{sec:eval}~and~\ref{sec:discussion}.

\item \textit{Control and Manipulation}:
Automatic movement generation is fully utilized when provided with a level of control over the characteristics of the movements being generated \citep{Lee:2002ik,pejsa_state_2010,geijtenbeek_interactive_2012}.  
 The ability to control and manipulate the movement is one of the main elements that give data-driven, and statistical generation techniques in particular, benefit over using just the recorded movements. 
A single model can generate many variations of the same movement, while one would otherwise need to capture all those variations individually, and blend or sequence them manually.

An agent's movements portray its personality, emotional state, goals, and intentions,  while corresponding to its reactions to the external stimuli from the environment and other agents surrounding it  \citep{bartenieff_body_1980,Studd:2013uc}.  One can therefore use high-level cognitive attributes and states in order to control and manipulate the generated movements.

These many sources of influence  result in a large combinatorial space of possible movements.  Consequently, the problem of manipulation and control is nontrivial and it brings challenges and requirements that we detail below and throughout this paper.


\begin{itemize}

\item[$-$] \textit{Movement Parameterization}: 
Directly manipulating movement at the level of raw data (joint rotations) is cumbersome and inefficient, mostly due to the low-level, high-dimensional, dense in time, and non-linear space of movement data.  
It is easier to manipulate a high-level representation that is sparse in time and has fewer dimensions than the raw data.
In addition, it is easier to associate a high-level representation with the meaningful characteristics of the agent and its movement.  
This has motivated the research on learning a mapping from a low-dimensional control space to the high-dimensional pose space, as well as performing operations such as interpolation and extrapolations on the parameters.  Techniques for addressing these are described in details in  Section~\ref{sec:synthesis}.




\item[$-$] \textit{Characterization of Movement}: 
In order to properly integrate the movement generation process into an agent with physiological and psychological properties, the high-level parameters used to control the movement have to correspond to, directly or indirectly, the agent's physiological, mental,  emotional states,  components, as well as the properties of the environment in which the agent resides . As will be discussed in further details in Sections~\ref{sec:charac} and~\ref{sec:data},  most studies have not adopted  a characterization framework that refers to meaningful concepts and either model a single pattern of movement, or model arbitrary variations of  a movement pattern. 


\item[$-$] \textit{Motor Variability:} 
Various studies have shown that variability is a fundamental characteristic of the movement of biological entities including humans \citep{Davids:2006vw,Muller:2009ch}. Humans never exactly repeat the same movement even when they try to do so. In other words, although multiple repetitions of the same movement can have the exact same functional, planning, and expressive descriptors, the execution dimension will always differ for each try. 
Thus, models that replicate the same execution will be perceived as more mechanical than natural. 
For example, Motion Graphs \citep{Kovar:2002dq} and similar approaches use exact copies of the recorded motion capture segments (except some of the transitions), which replicate the same execution over and over. 
\end{itemize}

\item{\textit{Interactive and Real-Time Animation:}}
Generating movement animation interactively, as in applications such as video games, requires two conditions to be present: 

\begin{itemize}
\item[$-$] \textit{Computational Constraints:}
A model that generates new samples in real-time given a set of parameters makes it possible to be used in interactive applications, in which real-time generation of the contents is desired. 
Real-time generation brings challenges in both time and space complexities of the generation algorithms of a statistical model.  The model should be able to generate new frames according to the  frame-rate of the system, while leaving enough processing power and memory for other computations needed in the system. 

\item[$-$] \textit{Generate Transitions}: Interactive animation requires making a large number of transitions between consecutive movements segments. Due to the dynamic nature of the scenarios, the exact timing  and occurrences of such transitions cannot be defined and authored by the animator a priori. Therefore, the transitions need to be generated in real-time.  While generating a transition can be seen as simply blending the source and target movements, a statistical model that has learned a general model of movement can be able to generate transitions the same way it generates any movement.
Creating smooth and believable transitions is more challenging than generating movement segments with fixed characteristics and remains an open problem, which is discussed in Sections~\ref{sec:charac} and \ref{sec:synthesis}.

\end{itemize}
\end{enumerate}

As described above, automatic movement generation research follows the goals along three general themes of believability, control and manipulation, and interactive  animation.
In the next part, we argue that the aforementioned research goals are shaped by two types of applications for automatic movement generation.

\subsection{Statistical Movement Generation Architecture}

\begin{figure}[t]
\centering
\includegraphics[width=0.7\linewidth]{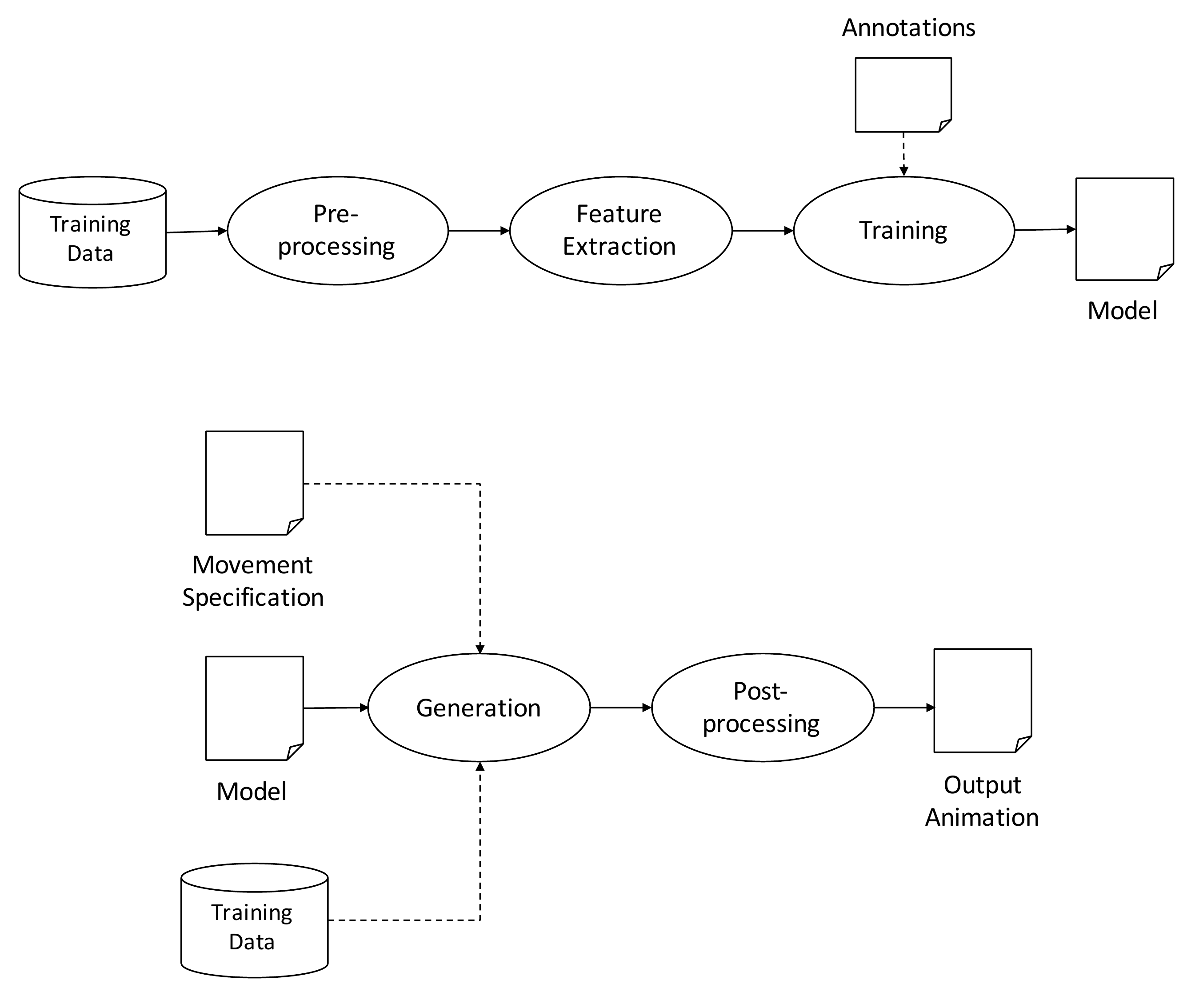}
\caption{Training the model (top), and generating new movements (bottom).}
\label{fig:pipeline}
\end{figure}


Statistical movement generation can be described as synthesizing new movements by learning a movement model from a group of recorded movement segments. 
The typical architecture of a movement learning and generation model is shown in  Figure~\ref{fig:pipeline}. In this section, we briefly highlight different parts of the architecture, while leaving their detailed discussions in the proceeding sections.

The first choice in designing a generative model is the type of the movements to generate.
In data-driven approaches, the repertoire of the movements that a model can generate relies on the diversity of the movements that exist in the training data, the available variations of each movement type,  the number of samples for each variation, and the number of human actors performing these samples.   
The size of the training dataset is important. If trained on a relatively small dataset, a machine learning model might closely imitate the only movements it has seen, but fail to learn a more general space of possible movements. 
In case of supervised learning, it is required to annotate the training data based on some descriptors  (Section~\ref{sec:data:annot}).

Depending on the choice of the movements, one can use the data available from a public database (Section~\ref{sec:data:base}), or record the movements using a motion capture system (Section~\ref{sec:data:capture}). The latter is costly and time-consuming, but could produce a more desirable set of data, while the former data are ready to use, but might not directly fit the desired requirements of a particular study.

In the optional pre-processing and feature extraction stage, the raw training data is transformed to a set of features to make them more suitable to be used by the learning algorithm.  The pre-processing stage can include changing the rotational representation of the data (Section~\ref{sec:data:rep});  calculating  the joint speed and acceleration, or learning more suitable representations of the raw input data using feature learning or extraction algorithms (Section~\ref{sec:data:feat}); or reducing the dimensionality of the data (Section~\ref{sec:data:dim}).
The training sequences might also be divided into shorter segments. 

If multiple databases are combined, each data source might use a different skeleton with a different shape, size, and number of body joints. In most cases, one needs to re-target the data to a uniform representation of the data from all sources so that they can be interpreted by the machine learning model in the same way.


In the training stage, depending on the machine learning technique and the type of the features used, a learning algorithm is employed to determine the generation function. In some cases, more that one learning algorithm might be used for different parts of the system. 
The learning can be supervised, unsupervised, or semi-supervised. 
In supervised learning, the training process involves learning the correlation between the movement data and a set of labels. In unsupervised learning, the data is not labeled and the model learns the underlying patterns that generate the data. In semi-supervised learning, only a subset of the training data is labeled and the training process involves both supervised and unsupervised techniques. 
The choice of which learning method to use is determined by the problem being addressed, as well as the approach designed  to tackle the problem. We will discuss the examples and implications of each method throughout the paper.

A generation algorithm (Section~\ref{sec:synthesis}) uses the learned model to create new samples. A group of models are able to generate new data based on a given description of the movements, which allow controlling the qualities of the generated movements (Section~\ref{sec:move:goals}). Some machine learning techniques such as Gaussian Process Models)need to retain the training data to be able to generate new samples while others require to keep only a few initialization frames (such as some artificial neural networks) or do not need any data for the generation (such as Hidden Markov Models).

The raw output of the model goes through the post-processing stage to be converted to a movement representation that can be used for animation. It is often the case that the post-processing involves reversing the steps performed in the pre-processing stage.

The quality of the output of the system is then evaluated formally or informally, as discussed in (Section~\ref{sec:eval}).

\section{Characterization  of Movement}
\label{sec:charac}

Movement is multifaceted. Multiple elements influence the movement: the internal state of the agent performing the movement (e.g., emotions and intentions), as well as the external stimuli that shape the environment surrounding the agent (e.g., objects, gravity, friction, etc.).  

We use the term ``factors'' to describe \textit{the sources of influence} on the agent movement. 
Each factor has a specific domain, which can be continuous or discrete. Choosing different values for a factor results in movements with different characteristics.
For example, if we consider the position of the hand as a factor, the factor space would be the 3D space that is within the reach of the hand. Or if we consider the affective state of the agent as a factor and follow a categorical representation of affect, the categories such as happy, sad, or afraid would be within the space of the factor.

The interaction and combination of the factors across multiple dimensions (e.g., affective state, actions, etc.) result in the endless varieties of movement that humans can perform. 
As building movement generation systems that can understand and generate all of this endless variety is not yet feasible (see the discussion in Section~\ref{sec:discussion}), researchers choose a subset of movements to model, and only a few factors to describe these movements (if any.)

In this section, we first review and criticize the characterization of movement and its factors in the literature.
Next, we present a  framework to characterize movement based on the factors that are meaningful to agents. 

\begin{table*}[th!]
\begin{adjustwidth}{-0.5cm}{-0cm}
\centering
\sffamily
\bgroup
\def\arraystretch{1.5}

\caption{
    An overview of characterization of movement in the reviewed literature.
}
\label{tbl:stylesum}
\scriptsize
\tiny
\begin{tabular}{@{}p{0.3cm} L{2cm}L{1.6cm}L{1.4cm}L{0.4cm}L{2.5cm}L{0.5cm}L{0.5cm}L{0.5cm}L{2.6cm}@{}}
& \rot{\textbf{Ref}} &	
\rot{\textbf{ML Technique}}	&	
\rot{\textbf{Characterization}} &	
\rot{\textbf{Control Support}} & 
\rot{\textbf{Control Technique}} &	
\rot{\textbf{Discrete / Continuous Factors}} & 
\rot{\textbf{Supervised / Unsupervised Learaning}} & 
\rot{\textbf{Single / Multi Factor(s)}} & 
\rot{\textbf{Modelled Movements}} 
\\ \hline

\circledtbl{1} & \cite{brand_style_2000}& Stylistic HMM& Expression - Personal Signature & Yes   & Parametric Gaussian Distribution &  C & U  & M & Gender, weight distribution, grace, energy, and formal dance styles \\ \hline

\circledtbl{2} & \cite{tanco_realistic_2000}& HMM & Function  & Yes   & Hierarchical architecture and clustering  &  D  & U  & S & Standing up, walking, running  \\ \hline

\circledtbl{3} & \citep{li_motion_2002}  & LDS   & Expression  & No& -   & - & -  & - & Disco dance\\ \hline

\circledtbl{4} & \cite{yamazaki_human_2005} & HSMM + Multiple Regression & Expression & Yes   & Parametric Gaussian Distribution & C & S  & M & Walking with different speed and stride length \\ \hline

\circledtbl{5} & \cite{wang_learning_2005} & HMM & Function  & Yes& Hierarchical architecture & D & U  & S & Regular walk, chopping a tree, ballet walk, ballet roll, disco, and complex disco\\ \hline

\circledtbl{6} & \cite{wang_key-styling:_2006} & SOMN  & Planning & Yes   & Parametric Gaussian Distribution&  C & S  & M & Boxing, with varying body height and the distance of punch target  \\ \hline

\circledtbl{7} & \citep{wang_learning_2006}& HMM/Mix-SDTG &Planning& Yes   & Parametric Gaussian Distribution& C & S  & M & The height of the right arm \\ \hline

\circledtbl{8} & \cite{taylor_modeling_2007}   & CRBM  & - & No & -   &  - & -  & - & Walking and running \\ \hline

\circledtbl{9} & \citep{wang_multifactor_2007}  & Multifactor GPLVM   & Expression  & Yes   & Mapping points in a low-dimensional latent space to movement & C & Semi S & M & Identity and gait in walking \\ \hline

\circledtbl{10} & \citep{Wang:ia}  & GPDM   & -& No   & -  &  - & - & - & Walking  \\ \hline

\circledtbl{11} & \citep{lin_applying_2008} & MLP  &   Planning  & Yes & Regression  &  C & S  & M  & Humanoid arm movements  \\ \hline

\circledtbl{12} & \cite{qu_motion_2008}  & Isomap + LDS & Function and Expression & No & -   & - & -  & - & Boxing, Indian dance\\ \hline

\circledtbl{13} & \citep{herzog_parametric_2008,herzog_recognition_2009} & Parametric HMM &  Planning  & Yes   & Interpolating individual models for each parameter value and using parametric Gaussian distribution &  C & S  & M & the position of the pointing target\\ \hline

\circledtbl{14} & \cite{taylor_factored_2009}   & Factored CRBM& Expression and Planning & Yes   & Modulating network weights   &  C & S  & M & Walking styles + walking speed and stride length  \\ \hline

\circledtbl{15} & \cite{Matsubara:bp}& Nonlinear Dynamical Systems & Planning & Yes   & Dynamical systems & C & S  & M & Table tennis \\ \hline

\circledtbl{16} & \cite{tilmanne_expressive_2010}  & PCA   & Expression   & Not explicit & Principal Components & - & -  & - & Walking styles \\ \hline

\circledtbl{17} & \citep{wei_physically_2011}& GP + Physics & -  & No &- & - &- &  - & Running, walking, and jumping under different physical forces \\ \hline

\circledtbl{18} & \citep{liu_human_2011}& Multilinear ICA& Planning and Expression& Yes & Optimization   &   C & S  & M  & Sideways stepping, reaching, and striding over obstructions, with multiple actors   \\ \hline

\circledtbl{19} & \citep{Kulic:2011hw}& HMM   & Function  & Yes& Hierarchical model & - & -  & - & Arm raising, bending, walking, squating, kicking\\ \hline

\circledtbl{20} & \cite{Chiu:2011do} & Hierarchical FCRBM  & Execution& Yes &  Modulating network weights &  C & S & M  &Prosody of speech in gestures   \\ \hline

\circledtbl{21} & \cite{Chiu:2011uv} & Hierarchical FCRBM  & Expression  & Yes   & Modulating network weights + blending networks & D& S  & S & Walking styles \\ \hline

\circledtbl{22} & \cite{Taubert:2011jp,Taubert:2012gl}  & Hierarchical GPLVM + HMM   & Expression& Yes   & Model interpolation &  D & S  & S & Handshake   \\ \hline

\circledtbl{23} & \cite{tilmanne_stylistic_2012}& HMM + Transformation & Expression   & No & Model interpolation + Transformation algorithms &  - & S  & S & Walking styles \\ \hline

\circledtbl{24} & \cite{Min:2012jy} & Functional PCA + Gaussian Mixture Model + Gaussian Process & Function and Planning & Yes   & Graphs + Optimization & C & S  & M & Function transitions: walking, sitting, picking, placing; Controling the movement: direction, end-position, speed \\ \hline

\circledtbl{25} & \cite{Samadani:2012ki} & Functional PCA& Expression& Not explicit & Principal components + clustering    &  - & U  & -  & Hand movement   \\ \hline

\circledtbl{26} & \cite{Tilmanne:2014tx} & HSMM  & Expression   & Yes   & Model interpolation  & D & S  & S & Walking with different emotions, morphology personifications, or situations \\ \hline

\circledtbl{27} & \cite{Fragkiadaki:2015vx} & Recurrent Neural Networks & Function   & No & - & -  & - & -& Waling, eating, smoking \\ \hline

\circledtbl{28} & \cite{Alemi:2015kg}   & Factored CRBM&Expression & Yes   & Modulating network weights & C & S  & M & Valence and arousal dimensions of affect  \\ \hline

\circledtbl{29} & \citep{CrnkovicFriis:2016vx}& LSTM RNN&   Expression& No & -   &- & -  & -  & Contemporary Dance\\ \hline

\circledtbl{30} & \citep{Holden:2016bv} & Convolutional Autoencoders + Feedforward Networks & Planning & Yes & Using a \textit{control} neural network &  C & Semi S & M  & Navigation, punching and kicking, factor transfer, crowd animation \\ \hline 

\end{tabular}
\egroup
\end{adjustwidth}
\end{table*}


\begin{table*}[th!]
    \begin{adjustwidth}{-0.5cm}{-0cm}
    \renewcommand\thetable{I}
    \centering
    \sffamily
    \bgroup
    \def\arraystretch{1.5}

    \caption{An overview of characterization of movement in the reviewed literature - \textit{Continued}.}
    \label{tbl:stylesum2}
    \scriptsize
    \tiny
    \begin{tabular}{@{\vspace{0.3em}\hspace{1em}}m{0.3cm} m{2cm}L{1.6cm}L{1.4cm}L{0.4cm}L{2.5cm}L{0.5cm}L{0.5cm}L{0.5cm}L{2.6cm}}
    & \rot{\textbf{Ref}} &	
    \rot{\textbf{ML Technique}}	&	
    \rot{\textbf{Characterization}} &	
    \rot{\textbf{Control Support}} & 
    \rot{\textbf{Control Technique}} &	
    \rot{\textbf{Discrete / Continuous Factors}} & 
    \rot{\textbf{Supervised / Unsupervised Learaning}} & 
    \rot{\textbf{Single / Multi Factor(s)}} & 
    \rot{\textbf{Modelled Movements}} 
    \\ \hline


    
    \circledtbl{31} & \cite{Wang:2017bs} & Adverserial Learning & Function and Expression & Yes   & Factors as conditional inputs & C & S  & M & Experiments \\ \hline
    
    \circledtbl{32} & \cite{Herrmann:2017hj} & Functional PCA + Gaussian Mixture Model + Gaussian Process + kMeans Trees & Function, Planning, and Expression & Yes   & Graphs + Optimization & C & S  & M & Walking with different emotions, lifting, sitting, moving books, knocking on the door, throwing \\ \hline
    

\circledtbl{33} & \cite{Martinez:2017ta} & Recurrent Neural Networks & Function & No   & - & - & -  & - & Mean angle error for prediction of a variety of actions \\ \hline


    \circledtbl{34} & \cite{Alemi2017_WalkNet} & Factored FCRBM & Planning and Expression & Yes   & Modulating network weights  & C & S  & M & Valence and arousal dimensions of affect, walking direction, movement signature \\ \hline

    \circledtbl{35} & \cite{Alemi2017_GrooveNet} & Factored FCRBM & Expression & Yes   & Modulating network weights  & C & S  & M & Dance movements for a given song\\ \hline

\end{tabular}
\egroup
\end{adjustwidth}
\end{table*}

\subsection{Movement Factors in the Literature}
We present a summary of the dimensions of movement that are characterized, the definitions and application of the factors, and the controlling abilities of the reviewed works in Table~\ref{tbl:stylesum}. 

The majority of the works address movement characterization from a perceptual perspective, i.e., how an arbitrary factor changes the perceived movement, rather than from an agency perspective, i.e, how factors that characterize an agent's internal state change the movement.

Research on perceptual systems has identified the notion of \textit{content} and \textit{style} as two factors of a perceptual system \citep{Tenenbaum:2006eo}. For instance, the same word (the content) can be spoken in different accents (the style), or the same letter (the content) can be written with different fonts (the style). 
Although movement is not merely a perceptual system,  style and content separation is  applied to the domain of human movement analysis. For instance, \textit{walking} from point $A$ to point $B$ in an environment, the \textit{content},  can be performed in different \textit{styles}, such as taking different paths, exhibiting distinct movement signatures, or expressing different emotions. 

Consequently, the research on statistical movement generation has adopted the concept of \textit{style and content separation} as a method for controlling the characteristics of the generated movements, .e.g., \cite{wang_key-styling:_2006,taylor_modeling_2007,herzog_recognition_2009,Tilmanne:2014tx,Alemi:2015kg}  \circledone{6} \circledone{8} \circledone{13} \circledone{26} \circledone{28}, creating new styles for movements, e.g., \citep{Chiu:2011uv,Tilmanne:2014tx} \circledone{21} \circledone{26}, or transferring  the style of one movement to another, e.g.,\citep{brand_style_2000,wang_multifactor_2007} \circledone{1} \circledone{9}. 

Employing the two-dimensional style and content characterization of movement brings out the question of what is considered the content of a movement and what is considered its style. 
A majority of studies have considered gaits variations during locomotion, or anthropomorphizations of non-human creatures as the \textit{style} factor of movement, treating the locomotion as the \textit{content} factor, e.g.,  \citep{taylor_factored_2009,tilmanne_expressive_2010,tilmanne_stylistic_2012,Chiu:2011uv,Tilmanne:2014tx} \circledone{14} \circledone{16} \circledone{23} \circledone{21} \circledone{26}.
Other studies consider gender or the personal movement signature as the stylistic factors, e.g,  \citep{brand_style_2000,wang_multifactor_2007} \circledone{1} \circledone{9}, or characteristics such as the position of a body part, walking speed, and stride length as the style of movement, e.g., \citep{yamazaki_human_2005,wang_key-styling:_2006,taylor_factored_2009, Matsubara:bp} \circledone{4} \circledone{6} \circledone{14} \circledone{15}.
A few studies model movement factors by  using a more specific characterization scheme than style and content. For instance,  \citet{Taubert:2012gl} \circledone{22} and \citet{Alemi:2015kg} \circledone{28} use factors that represent the categories of emotions or the valence and arousal dimensions of affect, respectively.

The review of the literature  reveals two main issues on movement characterization:

1) The definition of the \textit{style} factor varies across the literature, and there is no consensus on what \textit{style} represents. 
While many studies do not provide any definition of \textit{style}, some refer to it as the quality of movement that changes across the training data. The most viable definition of \textit{style} is used by \citet{brand_style_2000} \circledone{1} as the variations of the same movement type.

2) Simply using the two dimensions of content and style as the influential factors is insufficient in describing the multifaceted nature of agent movement. A framework with a broader range of dimensions is required to distinguish adequately between various movement qualities, and to better connect those qualities to the internal state of the agent. 
Only a few studies have addressed associating the controlling factors with the internal state of an agent, e.g., \citep{herzog_parametric_2008,Matsubara:bp,Taubert:2012gl,Alemi:2015kg} \circledone{13} \circledone{15} \circledone{22} \circledone{28}. 

In the next section, we make the case for a characterization framework tailored for the integration with an agent model that addresses the above shortcomings.

\subsection{Movement Characterization for Agents}
\label{sec:char_agent}
When describing the internal state of an agent and its surrounding environment, we can often put things into multiple semantic dimensions.
As the agent movement is influenced by these factors, we can also describe the qualities of its movement across multiple semantic dimensions. 
The way that these dimensions are laid out and incorporated into a movement generation system plays an important role the application of the movement generation model. For example, movement generation for video games would benefit from a high-level interface that corresponds directly to an agent model and generates the movement according to the changes in the agent's internals. 
Note that, although inspired by humans,  such framework may not exactly mirror how human internal state and movement work together.


\begin{figure}[t]
\centerline{\includegraphics[width=0.4\linewidth]{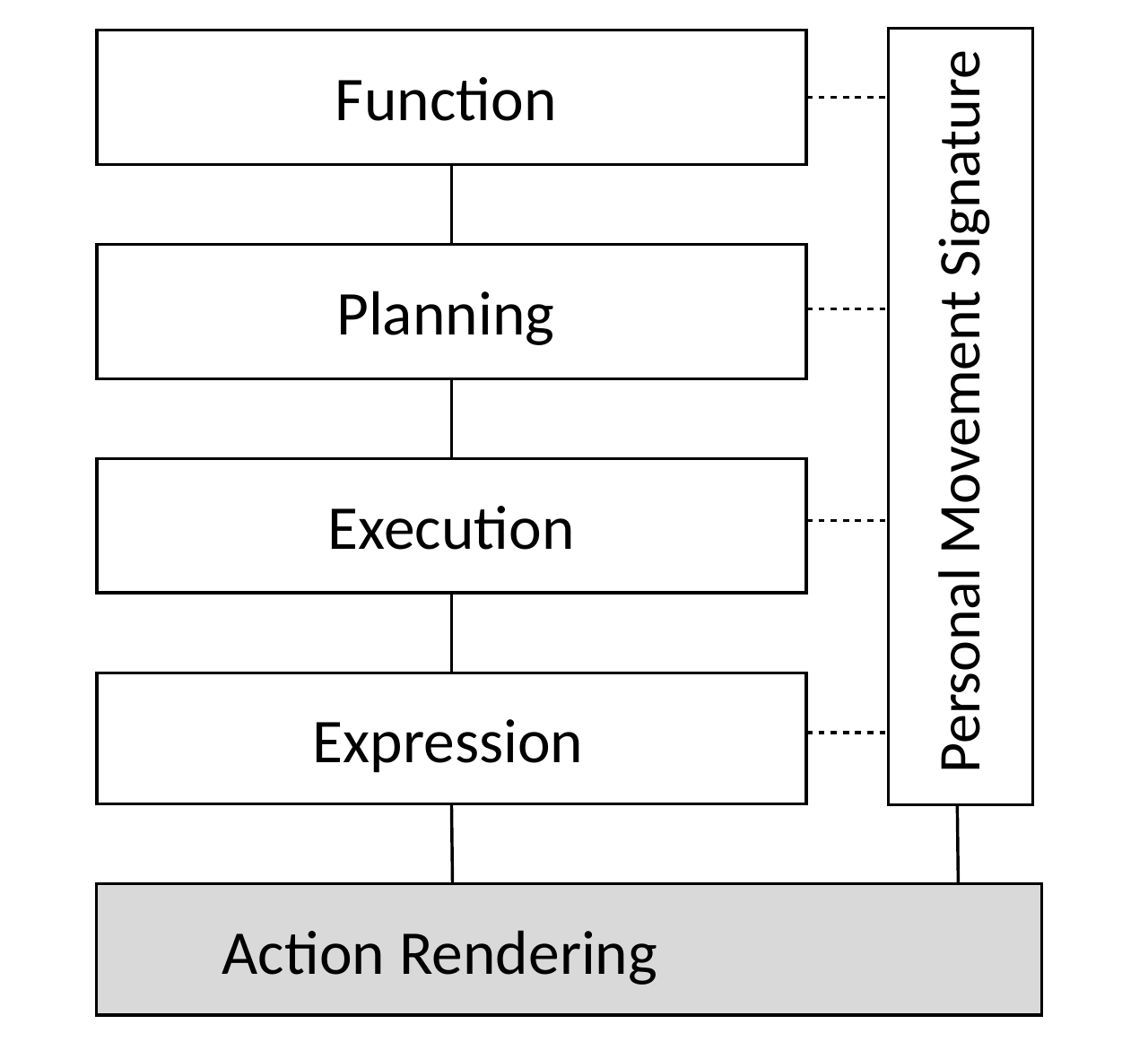}}
\caption{The multi-dimensional framework for movement characterization of animated agents.} 
    \label{fig:fpeel}
\end{figure}

Here, we present a framework for characterizing movement that is tailored for integration with virtual agent models. We use this framework throughout this paper to provide a coherent analysis of the literature. The proposed framework consists of six semantic dimensions that characterize an agent's movement : \textit{function},  \textit{planning}, \textit{execution}, \textit{expression}, \textit{personal movement signature}, and \textit{limb motion} (Fig.~\ref{fig:fpeel}):

\begin{description}
\item[Function] According to the goal-oriented behaviour of an agent and at the cognitive level,  the {functional} dimension of movement  corresponds to the task that the agent is performing through its movement: e.g., reaching the destination from its current location or picking up an object from the table. The functional dimension is perfunctory and does not communicate expression \cite{CruzGarza:2014dna}.
Note that the function may not always explicitly
be present in some movements, such as dancing or abstract movements \citep{Karg:2013fd}.



In modelling the functional dimension of movement in generative models, the common practice is to build a different model for each function. Upon generation, controlling the function of the movement is done by selecting and switching between the available models. A more challenging approach is to build a single model that is capable of generating a variety of functions, e.g., \citep{li_motion_2002,qu_motion_2008} \circledone{3} \circledone{12}. 


\item[Planning] The {planning} dimension is concerned with the sequencing of the fully-body movement and  limb motions in order to realize a desired task.
For example, moving between two points in a room requires planning the movement in a way to avoid any obstacles. Another example would be choreography, which lays out the sequence of movements in a dance. 




Different sequencing and timing of movements used for planning are often implemented by modelling parametrized movements. For example, a model can be parameterized by the  position of end-effectors \citep{wang_learning_2006,herzog_parametric_2008,herzog_recognition_2009} \circledone{7} \circledone{13} or the trajectory of end-effectors \citep{Matsubara:bp} \circledone{15} in order to capture different planning sequences that can achieve the tasks.

\item[Execution] This dimension encapsulates the patterns that result from the coordinated motion of individual limbs in order to realize the higher-level dimensions of movement in the physical or virtual world. For example, walking (the function) is executed through a locomotion pattern. 
In other words, the patterns that are defined across the execution dimension act as templates for realizing function and plan variations.


During the training process, the main task of the learning algorithm is to learn the underlying patterns that produce the movements. Any machine learning model would learn one or more execution templates. In addition, some models learn how these templates are modulated by other dimensions, such as expression or planning.

\item[Expression] The \textit{expressional}  dimension refers to the exhibition of affect through body movement, 
including the emotions and mood. The expression can be understood as modulation of the execution pattern.  
For example, for most people, walking while being angry looks different from walking the same path while enjoying it.  

The expression in movement can be described using a variety of representations, such as Laban Movement Analysis (LMA)  \citep{bartenieff_body_1980} for describing movement qualities, or categorical and dimensional representations of affect for emotions \citep{Karg:2013fd}. 
Controlling the expressivity of the movement is done by learning separate models  for each state, e.g., \cite{Tilmanne:2014tx} \circledone{26},  or learning movements that are parameterized by factors describing the expressivity  e.g., \citep{Taubert:2011jp,Alemi:2015kg} \circledone{22} \circledone{28}.


\item[Personal Movement Signature]

This dimension encapsulates the qualities across the other dimensions that together make the movements of one individual distinguishable from the movements of others.

\item[Action Rendering] At the lowest semantical level, the positions of the individual body limbs are manipulated through space and time, as defined by the execution patterns and modulated by the expressive factors. Action rendering is defined in the pose space, in which we only deal with the configuration of body parts.


\end{description}


Now that we have discussed how to provide a high-level, semantical representation of movement,  we look at how to capture and represent movement at a lower level in the next section.

\section{Movement Data}
\label{sec:data}

Data-driven and statistical movement generation systems do not incorporate any prior knowledge of human movement into their models. 
As a result, 
the collection of the movement data that are used in creating such models plays an important role in the generative capabilities and of them. 

Table~\ref{tlb:data} summarizes the characteristics of the training data used in the reviewed works.
There are a number of choices involved in acquiring a training data set for movement generation, including the type of sensors by which the movement is captured, the way movement data is represented numerically, whether annotations and labels are needed, and the number of human subjects that are available in the data.  
In addition, there are a number of processing operations that are often performed on the raw data to make them more suitable for a particular machine learning model.
In this section, we discuss and review each of these aspects with respect to common practices as reported in the literature. We also include the review of the freely available movement databases that can be used for movement generation. 

\begin{table*}[tbph!]
\begin{adjustwidth}{-0.9cm}{-0cm}
\sffamily
\bgroup
\def\arraystretch{1.5}
\tiny
 \centering
 \renewcommand\thetable{II}
 \caption{Characteristics of the training data used in the reviewed models }
 \label{tlb:data}
 \begin{tabular}{m{0.3cm} L{1.5cm} L{1.8cm} L{3.1cm} L{3.8cm} L{4.55cm} }
 \hline
 &\textbf{Ref} &
  \textbf{Data Source } &
  \textbf{Capturing \& Input Format} & 
  \textbf{Content} & 
  \textbf{Pre-processing \& Feature Extraction} \\ 
   \hline

 \circledtbl{1} &  \citep{brand_style_2000} & 
 Various sources / \textit{Unspecified} &
 
Mocap Markers: 20

Framerate (FPS): 60

Raw Dimensions: \textit{Unspecified}

Features Dimensions: reduced to $<10$ with PCA
&
- Locomotion

- Amateur and professional ballet moves and modern dance

\textit{Unspecified number of subjects}
&
- Reducing dimensions using PCA

\\
  \hline 
\circledtbl{2} &  \citep{tanco_realistic_2000} & 

Their own (not shared)
 
  &
FPS: \textit{unspecified}

Raw Dimensions: 70

Features Dimensions: reduced to 15 with PCA

&
- Standing up from the floor

- Walking

- Running

(465 frames total)

Single subject
&
- Aligning data to be invariant to orientation

- Converting the rotations to the angle-axis representation adapted from \citep{Pennec:1997kq}

- Reducing dimensions using PCA


\\
  \hline
\circledtbl{3}  &  \citep{li_motion_2002} & 

Their own (not shared)
 
  &
FPS: 60

Features Dimensions: 60

&
- Disco dance (49,800 frames)

Single subject
&
- Converting the rotations to exponential maps
\\
  \hline
\circledtbl{4}  &  \citep{yamazaki_human_2005} & 

 Their own (not shared) &
 
Markers:  19

Raw Dimensions: 60

Features Dimensions: 180

FPS: 33

&
- Walking, varied pace and stride length, 66 sequences

Single subject
&
- Manually labeling each frame with motion primitives

- Calculating 1st and 2nd derivatives of the raw rotations

- Calculating the walking pace and stride length

\\
  \hline
\circledtbl{5}  &  \citep{wang_learning_2005} & 

\textit{Unspecified}
 
  &
FPS: \textit{unspecified}

Features Dimensions: \textit{unspecified}

&
- Walk (191 frames)

- Chop tree (700 frames)

- Ballet walk (146 frames)

- Ballet roll (169 frames)

- Disco (600 frames)

- Complexer disco (600 frames)

\textit{Unspecified number of subjects}
&
- Segmenting the elementary behaviour 

- Converting the rotations to exponential maps

- Calculating the first-order derivatives

\\
  \hline
\circledtbl{6}  &  \citep{wang_learning_2006} & 
 Their own (not shared) & 
 
Markers:  19

Features Dimensions:  120 

FPS: 33.3

&
- Normal walking of a male actor
 
- Cat walk of a female character with right arm raised

Two subjects
&
 - Converting the rotations to exponential maps

\\
  \hline
\circledtbl{7}  &  \citep{wang_key-styling:_2006} & 

 Their own (not shared)
 
  &
FPS: 66

Features Dimensions:  \textit{unspecified}

&
- Boxing (3 minutes)

- Single subject
&
- Manually labeling the motion

- Calculating the style values

\\
  \hline
\circledtbl{8}  &  \citep{taylor_modeling_2007} & 

 CMU, \citep{hsu_style_2005} 
 
  &
FPS: 30

Markers:  30 (CMU), 17 (\citep{hsu_style_2005})

Features Dimensions:  62 (CMU), 49 (\citep{hsu_style_2005})

&
- Waling and running

- Stylistic walks

Single subject
&
- Converting the rotations to exponential maps

- Feature standardization

- Removing constant zero dimensions


\\
  \hline
\circledtbl{9}  &  \citep{wang_multifactor_2007} & 

CMU
 
  &
FPS: 30

Features Dimensions: 89  




&
- Locomotion

Three subjects

&
- Calculating rotation and translation velocities

- Converting the rotations of joints with 3 DOFs and the global orientation to exponential maps

- Labelling the gait type

\\
  \hline
\circledtbl{10} &  \citep{Wang:ia} & 

CMU
 
  &
FPS: 60 \& 30

Features Dimensions: 50

&
- Single-subject walk (260 frames)

- Four-subjects walk (1146 frames) 

- Golf club swing (four samples, 1015 frames)

&
\textit{N/A} 


\\
  \hline
\circledtbl{11}  &  \citep{lin_applying_2008} & 

Their own (not shared)
 
  &

Features Dimensions: 21 positions  (5 arm joints + initial positon + final postion in 3D)

&
- Lifting arm movements (four repetitions)
&
\textit{N/A}
\\
  \hline
\circledtbl{12}  &  \citep{qu_motion_2008} & 

 CMU
 
  &
FPS: 60

Raw Dimensions: \textit{unspecified}

Features Dimensions: 5 \& 8

&
- Boxing, Indian dancing

Single subject
&
\textit{N/A}
- Reducing dimensions with Isomap
\\
  \hline
\circledtbl{13}  &  \citep{herzog_parametric_2008,herzog_recognition_2009} & 

 Their own (not shared) &
 
Markers:  7

Dimensions:  \textit{unspecified}

FPS:  \textit{unspecified}

Using 3D position of markers instead of rotations
&
- Grasping, pointing

Single subject
&
\textit{N/A}

\\
  \hline
\circledtbl{14}  &  \citep{taylor_factored_2009} & 

 CMU, \citep{hsu_style_2005} (quantitative analysis), Their own (not shared) (multiple style variables)
 
  &
FPS: 60

Markers:  30 (CMU), 17 (\citep{hsu_style_2005})

Features Dimensions:  62 (CMU), 49 (\citep{hsu_style_2005})

&
%
%

- Stylized walks 

 - Multiple style variables: cross-product of (slow, normal, fast) speed and (short, normal, long) stride length. 6000 frames.

- Quantitative analysis: seven types of walking, each at three different speeds

Single subject

&
- Converting the rotations to exponential maps

- Feature Standardization

- Removing constant zero dimensions

- Labelling the data

\\
  \hline
\circledtbl{15}  &  \citep{Matsubara:bp} & 

Their own (not shared)
 
  &
FPS: \textit{unspecified}

Features Dimensions: 12

Using markers positions
&
- Table tennis swings (15 sequences)

- Reaching (15 sequences)

Single subject

&
\textit{N/A} 


\\
  \hline
 \circledtbl{16} &  \citep{tilmanne_expressive_2010} & 

Their own (not shared)

  &
FPS: 30

Features Dimensions: 54

&
- Stylized walks (247 cycles)

Single subject
&
- Removing the root translation

- Re-sampling the data to have a fixed length

- Converting the rotations to quaternions for re-sampling

- Converting the rotations  exponential maps for PCA

\\
  \hline
 \circledtbl{17} &  \citep{wei_physically_2011} & 

\textit{Unspecified}
 
  &
FPS: \textit{Unspecified}

Features Dimension: 19, 22, and 19

&
- Waling variations: step sizes, turning angles, walking speeds, and walking slopes

- Stylized walking

- Locomotion

\textit{Unspecified number of subjects}
&
- Reducing dimensions using PCA

\\
  \hline
\circledtbl{18}  &  \citep{liu_human_2011} & 

Their own (not shared)

 &
FPS: \textit{unspecified}

Features Dimensions: \textit{unspecified} 

&
- Sideways stepping (75 sequences)

- Reaching (70 sequences)

- Stride over obstructions (78 sequences)

12 subjects
&
- Time warping all movement to a reference movement

- Manually specifying key-frames for each sequence

- Reducing dimensions using PCA

\\
 \hline

 \hline
\end{tabular} 
\egroup
\end{adjustwidth}
\end{table*}

\begin{table*}[tbph!]
\begin{adjustwidth}{-0.9cm}{-0cm}
\renewcommand\thetable{II}
\sffamily
\bgroup
\def\arraystretch{1.5}
\tiny
 \centering
 \caption{Characteristics of the training data used in the reviewed models - \textit{Continued} }
 \label{tlb:data2}
 \begin{tabular}{L{0.3cm} L{1.5cm} L{1.8cm} L{3.1cm} L{3.8cm} L{4.55cm} }
 \hline
 &\textbf{Ref} &
  \textbf{Data Source } &
  \textbf{Capturing \& Input Format} & 
  \textbf{Content} & 
  \textbf{Pre-processing \& Feature Extraction} \\ 
   \hline

\circledtbl{19}  &  \citep{Kulic:2011hw} & 

Their own (not shared)
  &
FPS: 30

Markers: 34

Features Dimensions: 90 and 120

Using markers positions

&
- Walking, squating, kicking, raising an arm

- Single subject
&
\textit{N/A}
\\
  \hline
\circledtbl{20} &  \citep{Chiu:2011do} & 

Data from Human Sensitivity study \citep{Ennis:2010gj}
 
  &
FPS: \textit{unspecified}


Features Dimensions of the arm joints: 21

&
- Debate conversations (1140 frames)

Three subjects
&
- Converting the rotations to exponential maps


\\
  \hline
\circledtbl{21}  &  \citep{Chiu:2011uv} & 

CMU
 
  &
FPS: \textit{unspecified}

Features Dimensions: 96

&
- Stylized walks 

Single subject
&
-  Removing the global translation

- Converting the rotations to exponential maps

\\
  \hline
\circledtbl{22} &   \citep{Taubert:2011jp,Taubert:2012gl} & 

Their own (not shared)

  &
FPS: \textit{unspecified}

Features Dimensions: 38 and 159

&
- Handshakes with four emotions 

Two subjects
&
\textit{N/A} 
\\
  \hline
 \circledtbl{23}   &  \citep{tilmanne_stylistic_2012}
 & 
Proprietary:

Mockey \citep{tilmanne_expressive_2010} 

eNTERFACE'08 3D \citep{Tilmanne2008eNTERFACE}
&
Inertial sensors:  18

Raw dimensions:  54

Dimensions with speed and acceleration: 162

FPS: 60 (Mockey), 30 (eNTERFACE'08 3D)

&
- Mockey: Walks in 11 styles: proud, decided, sad, top model, drunk, cool, afraid, tiptoeing, heavy, in a hurry, and manly.

- eNTERFACE'08 3D: Neutral walking sequences of 41 actors 
&
- Aligning the direction of all the walking sequences.

- Manually segmenting the sequences into left and right steps

- Converting to exponential maps

- Calculating the speed and acceleration of the rotations

\\
  \hline
\circledtbl{24}  &  \citep{Min:2012jy} & 

 Their own (not shared) &
 
Markers:  \textit{Unspecified}

Features Dimensions:  \textit{Unspecified}

FPS: \textit{Unspecified}

&
- Standing, walking, running, two-feet jumping,
stepping-stone jumping, sitting down, standing up, climbing up,
climbing down, left punching, right punching, picking, placing,
kneeling down, kneeling up, backward walking

- Transitions between the above

&
- Extracting keyframes

- Segmentation

- Segment registration for each primitive

- Functional decomposition of each segment


\\
  \hline
\circledtbl{25}  &  \citep{Samadani:2012ki} & 

Data from \citep{Samadani:2011bd}

  &
FPS: 84

Features Dimensions: 54

&
- Closing and opening the hand with happy, sad, and angry emotions
&
- Re-sampling the data to be aligned and have a fixed length

- Modelling the data with Basis Function Expansion method

\\
  \hline
\circledtbl{26} &   \citep{Tilmanne:2014tx} & 

Mockey Database \citep{tilmanne_expressive_2010} 

  &
FPS: 30

Markers: 34

Features Dimensions: 54

&
- Stylized walks

Single subject
&
- Automatically annotating frames with left and right steps

- Converting the rotations to exponential maps
\\
  \hline
\circledtbl{27}  &  \citep{Fragkiadaki:2015vx} & 

 H3.6M \citep{Ionescu:kt} &
 
Markers:  30

FPS: 50

Features Dimensions:  54

&
- Walking, eating, smoking

&
- Converting the rotations to exponential maps

- Calculating the speed of the roations and global translation

- Feature standardization


\\
  \hline


   %
\circledtbl{28} & \citep{Alemi:2015kg} & 

Their own (shared) \citep{MODADB}

  &
FPS: 30

Markers: 53

Features Dimension: 52

&
- Affect-expressive walks (36,000 frames)

Two subjects
&
- Converting the rotations to exponential maps

- Feature standardization

- Removing constant zero dimensions

- Labelling each training sequence with the valence and arousal values
\\
  \hline
\circledtbl{29}  &  \citep{CrnkovicFriis:2016vx} & 

Their own (not shared)

  &
FPS: 30

Features Dimensions: 75

Using 3D positions

&
- Contemporary Dance
&
\textit{N/A}
\\
  \hline

\circledtbl{30}  &  \citep{Holden:2016bv} &

 \citep{CMUDB} +

 \citep{HDM05DB} +

 \citep{Ofli:ky} +

 \citep{Xia:2015ep}
  &

FPS: 60

Features Dimensions: 70

&
All of the contents of the used databases, retargetted into a uniform skeleton, resulting in around six million frames (at 60 FPS)

&
- Converting the joint rotations to posittions with respect to a body-centric coordinate system

- Calculating the body orientation and global velocities

- Applying Gaussian filters to reduce noise

- Finding foot contact points

- Feature standardization

- Segmenting the data into overlapping windows  

\\
  \hline 
\circledtbl{31}  &  \citep{Wang:2017bs} & 

Emilya Dataset \citep{FOURATI14.334} &

FPS: 120

Features Dimensions:  \textit{Unspecified}

&
- 12 performers

- 8 activities

- 8 emotions

&
\textit{N/A}

\\
  \hline

\circledtbl{32}  &  \citep{Herrmann:2017hj} & 

 Their own (not shared) &
 
Joints:  20

FPS: \textit{Unspecified}

Features Dimensions: 79

&
- Walking

- Picking

- Placing

- Screwing

&
- Converting the rotations to quaternions

- Segmentation

- Temporal and spatial alignment

- Feature standardization

- Smoothing


\\
  \hline

\circledtbl{33}  &  \citep{Martinez:2017ta} & 

 Human 3.6M &
 
Markers:  30

FPS: 50

Features Dimensions:  54

&
- Seven performers 

- Walking, smoking, engaging in a discussion, taking pictures, and
talking on the phone

- Two different trials for each performer/actions

&
- Converting the rotations to exponential maps


\\
\hline

   \circledtbl{34}  &  \citep{Alemi2017_WalkNet} & 

   Their own (shared) \citep{MODADB}

   &
 FPS: 30
 
 Markers: 53
 
 Features Dimension: 52

  &
- Affect-expressive walks (36,000 frames)

- Two subjects

  &
  - Converting the rotations to exponential maps

  - Feature standardization
  
  - Removing constant zero dimensions
  
  - Labelling the valence and arousal values for  each training sequence

  - Labelling each frame with the orientation of the body
  
  \\
  \hline

    \circledtbl{35}  &  \citep{Alemi2017_GrooveNet} & 

    Their own (shared) \citep{MODADB}

    &
  FPS: 30
  
  Markers: 53
  
  Features Dimension: 52

  &
- Dance movements

- Audio features

  &
  - Converting the rotations to exponential maps

  - Feature standardization
  
  - Removing constant zero dimensions
  
  - Labelling each frame with corresponding audio features
  
  \\
  \hline


  















\hline
 
 \hline
\end{tabular} 
\egroup
\end{adjustwidth}
\end{table*}

\setcounter{table}{2} \renewcommand{\thetable}{\Roman{table}}

\subsection{Capturing Human Movement}
\label{sec:data:capture}

A number of sensor systems are used to capture the movements of human actors.
Depending on the application, one or more sensor system might be used to capture movement data.
These systems vary based on the areas of the body they capture, such as hand movements, full-body movements, expansion of the lungs through breathing,  and muscle contractions among others. They also capture different quantities  such as position, acceleration, biometrics, energy, etc. Other factors such as the setup requirements (e.g., indoors, outdoors, capturing volume, mobility), the precision and reliability of the measurements, and the sampling rate play a role in choosing the sensor system.
 
The quantities that sensors capture are summarized in the following categories: 

\begin{itemize}	
 \item[$-$]  Joint positions and rotations: motion capture systems
 \item[$-$]  Joint acceleration and orientation: accelerometer and gyroscope 
 \item[$-$]  Biometric features: electromyography, electroencephalography, breath, heart rate
 \item[$-$]  Location of the body: Radio Frequency ID (RFID), Global Positioning System (GPS), and Mobile Networks
\end{itemize}

Throughout the rest of the paper, we only focus on motion capture data, as none of the generative models reviewed use other sensor systems. 

\subsubsection{Motion Capture}
\label{sec:data:format} 
Motion Capture (Mocap) is a popular approach for recording movement and is widely  used in the movie, video game, sports, and health care industries.  Mocap systems use marker-based or marker-less techniques to capture the trajectories of body limbs in a 3D coordinate system. 
Mocap markers can be acoustic, inertial (as in \citep{tilmanne_expressive_2010} \circled{16}), magnetic, reflective (as in \cite{CMUDB,Alemi:2015kg} \circled{28}) or a combination of these.
Marker-less systems use computer vision techniques to track the optical flow of the pixels in a 2D video stream of movement (RGB and infrared), as used in Microsoft Kinect. Special motion capture systems for capturing the movements of hands and fingers can be worn like a glove, e.g., \citep{Lu:bv}.

Motion capture systems are often used to capture whole body movements. However, it is also commonly used for capturing detailed limb movements, e.g., \citep{Samadani:2011bd}, as well as facial expression. 

Regardless of the capturing techniques, the trajectories of the markers or pixels are often mapped to a virtual skeleton, defined by a hierarchy of joint angle rotations that ensures that the body limbs have fixed lengths. While most of the approaches use joint rotations, the trajectories are also directly used for modelling movement  \citep{Kulic:2011hw,CrnkovicFriis:2016vx} \circledone{  8} \circled{29}.

Mocap data are provided in different formats such as C3D (3D marker positions); Acclaim, Biovision (BVH), and Vicon  (both the skeleton and the motion data); text (comma or space delimited); and more general 3D asset formats such as COLLADA and FBX. 


C3D format \citep{C3DSpec} is a public-domain format which represents movement using the position of the limbs or markers in a 3D coordinate system. It does not include a skeleton or a hierarchical representation. Most of the raw data in movement database are provided in the C3D format.

The Acclaim format \citep{AcclaimSpec} is a skeleton-based  format for mocap data developed by the video game company Acclaim. It consists of two files, one for describing the skeleton information and one containing the movement data.
Typically, the same skeleton file is used for multiple movement sequences of the same subject.
Unlike other formats, in the Acclaim format the movement is represented by rotations of bones rather than the rotations of joints.

Biovision Hierarchical data (BVH) \citep{BVHSpec} is a widely used skeleton-based mocap format developed by Biovision.
Unlike acclaim, BVH contains both the skeleton and the content in a single file and uses the rotations of the joints to represent the movement. Most of the reviewed public databases provide their mocap data in the BVH  format.

COLLADA \citep{arnaud2006collada, COLLADASpec} and FBX \citep{FBXSpec} are standard asset exchange formats used in computer graphics and 3D animation software. 
They are designed to provide an extensible and flexible container for animation assets such as motion capture data, geometry, shaders, physics, lighting, animation, and kinematics as well as any custom data.  Thus, they can be used to integrate multiple data describing the movement such as annotations associated with the mocap.

\subsubsection{Motion Capture Data Representation}
\label{sec:data:rep}

Each frame of motion capture data consists of a root node which defines the body's \textit{absolute} position and orientation with respect to a global Cartesian coordinate system, and a set of nodes each representing a  joint's or bone's orientation. Each node, depending on what part of the body it corresponds to, can be represented by 1 (e.g. knee), 2 (e.g. wrist), or 3 (e.g. arm)  parameters, also called Degrees Of Freedom or DOF. Each parameter or DOF describes the rotation of the joint/bone along one of the axes of a 3D coordinate system \textit{relative} to its parent joint/bone. These parameters constitute  the input of the machine learning pipelines.


There are different parametrization schemes to represent the aforementioned joint angle rotations.
 None of these representations is perfect and depending on the application one might be chosen over another. 
One of the basic formalisms to represent rotational DOFs is to use {rotation matrices}. Using a rotation matrix, rotating a point can be implementing by a matrix multiplication.  A representation based on a $3\times3$ rotation matrix requires nine components plus three orthogonality constraints which require a larger space compared to other techniques. Because of this large number of components and the need for imposing constraints on various operations, it is not efficient to use rotation matrices for most of the applications. Four other common parametrization techniques are discussed below.

\textit{Euler Angles} is one of the most common representations for orientations in movement data. Euler angles describe a one, two or three DOFs of orientation by a sequence of rotations around each axis in the global or local coordinate system using a vector in $\mathbb{R}^3$. 
While widely being used by animators, they cannot be interpolated and are susceptible to the loss of degrees of freedom in which different combinations of its three components can lead to the same 3D rotation (also known as a \textit{gimbal lock}). Thus, few computational models use Euler angles in practice \citep{wang_multifactor_2007} \circledone{9}.

\textit{Quaternions} represent rotational DOFs using 4 components. Gimbal locks do not occur with quaternions representation, and interpolation is well supported with them. A minor shortcoming of quaternions is the extra \nth{4} component they use to represent the rotations compared to the Euler angles.  Despite this, the Quaternions representation is  commonly used in the  movement generation systems \citep{pejsa_state_2010,tilmanne_expressive_2010} \circledone{16}.

\textit{Exponential map} is another technique that is applied to motion capture representations.
``The exponential map maps a vector in $\mathbb{R}^3$ describing the axis and magnitude of a three DOF rotation to the corresponding rotation'' \citep{Grassia1998}. Exponential map representation has a number of benefits over Euler angles including the support for interpolation and being less susceptible to gimbal lock when used for modelling human movement. As a result, exponential map is the most commonly used representation for machine learning purposes \citep{li_motion_2002,wang_multifactor_2007,taylor_factored_2009,tilmanne_expressive_2010,Tilmanne:2014tx,Alemi:2015kg} \circledone{3} \circledone{9} \circledone{14} \circledone{16} \circledone{26} \circledone{28}.

\subsection{Training Data for Movement Modelling}


In the following, we discuss those aspects of modelling movement data that are relevant to creating generative movement models.

\textbf{Sampling Rate}
\label{sec:data:fps}
The sampling rate (frame rate) of the data represents how many measurements are recorded by the sensors in a window of time. 
In order to capture fast-paced movements, a high sampling rate is needed to produce a smooth recorded movement. The sampling rate of the training data might be adjusted to comply with the space and computational complexities of the statistical models, as well as to combine data from different sources that have different sampling rates. While the original data might be recorded in higher frame rates (e.g., 120HZ), most approaches down-sample the data (e.g., to 30HZ) to reduce the size of the training dataset.
Most of the motion capture formats use a fixed sampling rate when recording the data. This ensures that the frames are linearly sampled. 

\textbf{Pre-Processing \& Feature Extraction}
\label{sec:data:feat}
A generative system might require the raw mocap data to go through a series of processes to make the data usable for the learning algorithm. Some common processes include, but not limited to:

\begin{itemize}
\item[$-$] \textit{Data Representation:}
The representation of the data can be changed in a number of different ways in the pre-processing stage.
The motion capture files typically use the Euler formalism to represent the joint rotations. However, many approaches convert the data into exponential maps or quaternions before performing other processes on them. In addition, approaches that use functional statistics transform the motion capture data using basis functions \citep{Samadani:2011bd}. 

\item[$-$] \textit{Segmentation:} 
Some approaches use segmentation to break down long sequences or to organize the system into a hierarchal structure.
The segmentation can be done based on identification of elementary movements, which is discussed in more details in Section~\ref{sec:synthesis}, or based on choosing windows of fixed length, as done by \citet{Holden:2016bv} \circledone{30}.

\item[$-$] \textit{Alignment and Length Normalization:} Some approaches require the training data to have fixed lengths in such a way that similar movements (e.g., each walking cycle) are aligned. The alignment and resampling can been done using the SLERP algorithm \citep{Shoemake:1985gp}, or using piecewise linear re-sampling.

\item[$-$] \textit{Rotational or Positional Velocity and Acceleration:} Some studies calculate the velocity and acceleration of each DOF of the movement and add the extra features to the training data, as in \citep{wang_learning_2005,yamazaki_human_2005, wang_multifactor_2007,tilmanne_stylistic_2012} \circledone{5} \circledone{4} \circledone{9} \circledone{23}.

\item[$-$] \textit{Derived Movement Features:}
 It is also possible to derive other features from the movement data using analytical formulas. For example, the stride length is directly extracted from the data  and used as labels to annotate the data \citep{yamazaki_human_2005} \circledone{4}. 

\item[$-$] \textit{Dimensionality Reduction:} 
The curse of dimensionality, a concept in machine learning, states that the higher the dimensionality of the data, the more difficult it becomes to learn and model the data. 
Therefore, some statistical models are trained more effectively when given a fewer number of dimensions.  
Often many of the dimensions of the data do not carry much information about the underlying patterns.
Dimensionality reduction techniques, such as Principal Component Analysis (PCA) \citep{jolliffe2002principal}, are applied to the data to identify the dimensions that cause the most variations in the data and eliminate the ones that do not carry much information.
A reduced feature vector is then used for training the model, e.g., \citep{brand_style_2000, liu_human_2011,wei_physically_2011} \circledone{1} \circledone{18} \circledone{17}.

\item[$-$] \textit{Feature learning:}
Instead of solely using the raw movement data or the features that are derived from analytical approaches (also known as feature engineering), one can derive features through an unsupervised learning processes. In such learning process, a machine learning algorithm is used to learn a new representation of the data that could posses characteristics that are more efficient for learning a generative model that the raw data or analytical features. This is more common in deep learning applications in which a neural network is first trained on a large amount of training data in an unsupervised way, and then a second model is trained on a possibly smaller dataset that is used directly for the generation \citep{Holden:2016bv} \circledone{30}. 


\item[$-$] \textit{Normalization:} For approaches that use artificial neural networks, e.g., \citep{taylor_factored_2009, Alemi:2015kg,Holden:2016bv}  \circledone{14} \circledone{28} \circledone{30}, the training process converges more efficiently when the training data vectors are normalized to have zero mean and unit standard deviation. The normalization is often the latest stage of the pre-processing and is applied to each dimension of the data independently. 
\end{itemize}


The above operations can be discussed under the field of movement signal processing \citep{FeatSOA}. Unlike other forms of data such as audio and images that have a  well-established signal processing body of literature, studies on movement signal processing are still limited and scarce. 

\textbf{Dimensionality of Movement Data}
\label{sec:data:dim}
The dimensionality is another important characteristic of training data. The number of dimensions of each frame of the training data (i.e., the feature vector) is determined by the type of the rotation parameterization, the number of data points corresponding to the markers, joints, or bones, and any extra movement features that might be added to the feature vector such as the velocity or the acceleration of the joint rotations. 




\begin{figure}[t]
\centerline{\includegraphics[width=0.8\linewidth]{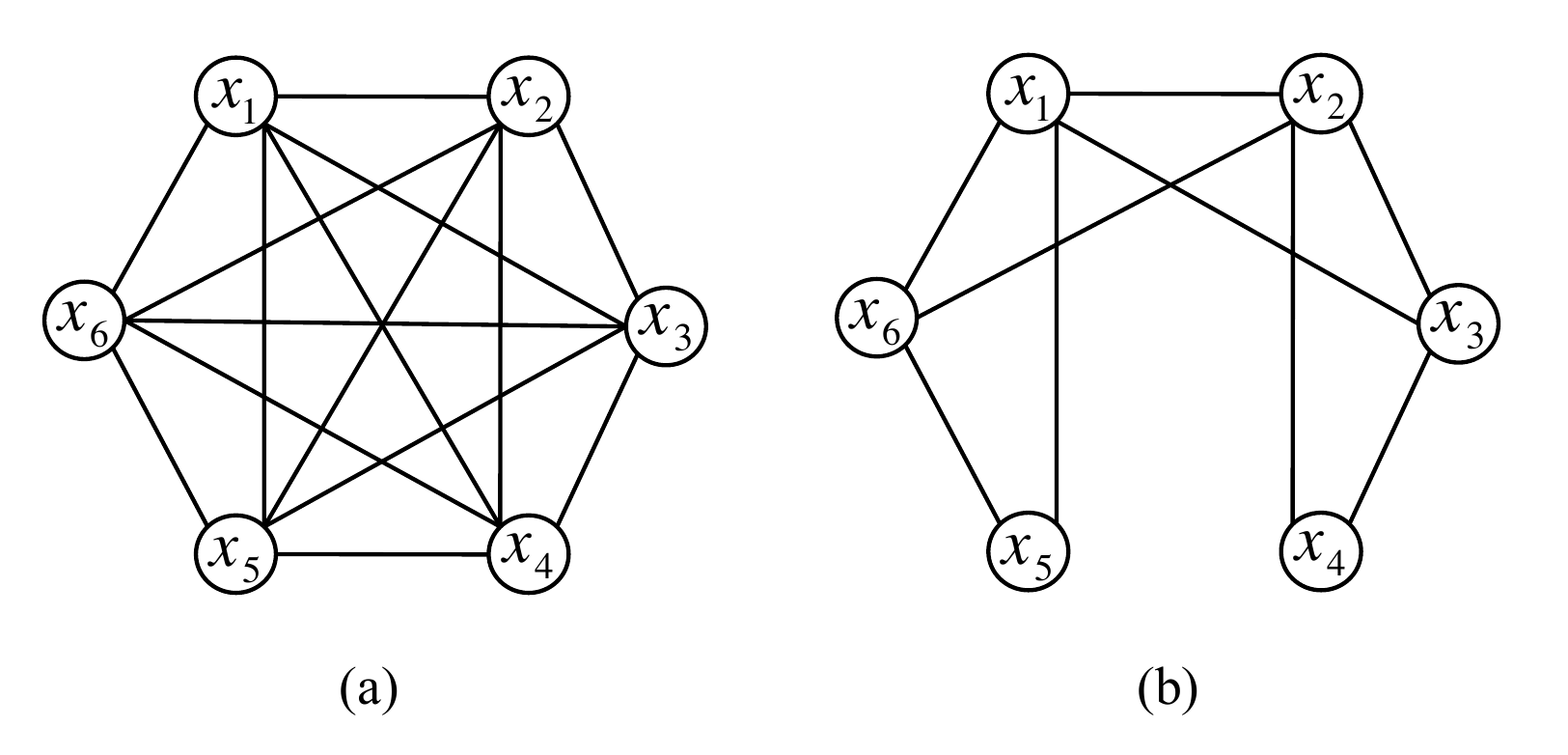}}
\caption{Comparison of (a) a Gaussian distribution  and (b) a DTG distribution with three components (the triangles). Each node represents a dimension of the data. Edges represent the dependencies between the dimensions. The DTG components are $\{x_1, x_5, x_6\}$, $\{x_1, x_2, x_3\}$, and $\{x_2, x_3, x_4\}$.}
    \label{fig:dtg}
\end{figure}

\textbf{Probability Distribution}
\label{sec:data:probdist}
The multivariate Gaussian distribution is the most commonly used probability distribution to model the  multi-dimensional movement data. 
The Gaussian distribution represents the correlations between the dimensions of the data with the assumption that each dimension conditionally depends on all other dimensions (see Fig.~\ref{fig:dtg}(a)). With this configuration, if several pairs of the data dimensions are not significantly conditionally dependent (as it is in the mocap data), the covariance matrix can become singular, which in turn prevents from sampling the distribution. 
However,  the high-dimensional human movement data can be modelled more efficiently with the assumption that each dimension only depends on two other dimensions \citep{song_2003}, and can be represented by a decomposable triangulated graph (DTG), as shown in Fig.~\ref{fig:dtg}(b)). 
Each triangle in the DTG is modelled by a 2-dimensional or 3-dimensional Gaussian distribution. DTGs are applied in a few movement generation studies, e.g., \citep{wang_learning_2005,wang_learning_2006} \circledone{5} \circledone{7}. Refer to the article by \citet{song_2003} for more information and discussion of the differences between the Gaussian and DTG distributions.


\textbf{Data Annotations}
\label{sec:data:annot}
Using supervised learning techniques requires the training data to be annotated (labeled) according to some descriptors. The annotations then allow the model to learn the associations between the given descriptors, and the underlying mechanisms that generate the movement patterns. The annotations are often marked manually by human observers.  

The annotations can be discrete to represent categorical data such as the mover's identity or gait type, or continuous to represent real-valued measurements such as stride length or hand position. Depending on the semantics of the descriptors and the learning mechanisms, the annotations can be associated to whole sequences of the training data for to each frame individually. 

Some of the descriptors that are used in the literature include \textit{the right and left steps} of a walking cycle, e.g., \cite{yamazaki_human_2005,tilmanne_stylistic_2012,Tilmanne:2014tx} \circledone{4} \circledone{23} \circledone{26}; the body height and the distance of a punching action \citep{wang_key-styling:_2006} \circledone{6}; the affective qualities of the movement, in terms of categorical emotions, e.g., \citep{Samadani:2012ki} \circledone{25}, or in terms of the valence and arousal dimensions, e.g., \citep{Alemi:2015kg} \circledone{28};  the walking speed and stride length \citep{taylor_factored_2009,yamazaki_human_2005} \circledone{14} \circledone{4}; or arbitrary class labels, e.g., the gait styles during walking \citep{taylor_factored_2009} \circledone{14}.


\begin{table*}[tbph!]
\begin{adjustwidth}{-1cm}{-0cm}
\centering
\sffamily
\bgroup
\def\arraystretch{1.3}
\caption{A summary of the motion capture databases}
\label{tbl:dball}
\scriptsize
\tiny
\begin{tabular}{@{}L{1.2cm} L{2.2cm} L{2.8cm} L{1.4cm} L{1.6cm} L{1.4cm} L{.9cm} L{1.2cm} L{1cm}@{}}
{\textbf{DB}} &{\textbf{Purpose}} & {\textbf{Content and Size}} & {\textbf{Characterization}} & 
{\textbf{Recording Tech}} & {\textbf{Modalities}} &  {\textbf{Subjects}} & {\textbf{Redundancy}} &
{\textbf{Annotation}} \\ \hline

Ohio State University's ACCAD \citep{ACCADDB}   &  Video Games \& Animation    & 300  sequences

- Locomotion

- Gestures

- Martial Arts
  &   Function with Planning Variations 
&       - Vicon       &          Marker Data, Skeletal Data  &      3     &  One instance per performer &    -       
  \\ \hline

AffectMe  \citep{AFFECTMEDB}   &      Study  of body posture as an indicator of human affective states, pain, and immersion        & Collection of datasets: 

- Acted emotions

- Non-acted affective states in computer game settings  &   Expression 
& 

- Vicon 

- Gypsy5  

&          -  &  -  &    -     &    Categorical Emotions 
\\ \hline

CMU Mocap  \citep{CMUDB} &   General Research   & 2605 sequences in five categories:

- Human Interaction

- Interaction with Environment

- Locomotion

- Physical Actitivties \& Sports

- Situations \& Scenarios
& Function with Planning and Expression Variations   
&
   - Vicon with 41 markers        &         Marker Data, Skeletal Data, Video, Animation  &                144 (including duplicates) &   Varies for different functions

    &       -      

\\ \hline

Cologne DB\citep{KOELNDB}   &   \textit{Unspecified}   &   
- Locomotion

- Arm Gestures 

- Skydiving, Swimming, Climbing  &       Expression with variations in Function    

&         26 joints       &         Marker Data, Skeletal Data  &  29                &     Different roles, moods, genders, and actors     &    -

\\ \hline

Dance Motion Capture Database of the University of Cyprus.
\citep{CYPRESSDB}   &   Digital Archive of Dance
     &  - Greek and Cypriot Dances  &    Expression  
     &         - Phasespace Impulse X2 with 38 markers
      &          Marker Data, Skeletal Data, Video  &        3 & -         &    -
      \\ \hline

HDM05 \citep{HDM05DB}   &  General Research  & 3 hours of motion captures in 70 different classes

- Locomotion

- Grabbing and Depositing

- Sports

- Sitting and Lying Down

- Dance
&   Function with variations in Planning and Performer  
&         - Vicon with 40 markers and 24 joints       &          Marker Data, Skeletal Data  &     5     & 10 to 50 instances of the same function performed by different subjects &   -
\\ \hline

HumanEva-I  \citep{Sigal:2006uw}   &  Human movement and pose estimation from video data     & 

- Walking

- Jogging

- Gestures

- Throw/Catch

- Combinations of the above
  &   Function 
  &         - ViconPeak       &          Synchronized Marker Data  and Video &  4      &    2 instances per function     &    - 
  \\ \hline

IEMOCAP \citep{Busso:2008dx}   &  
- Recognition and Analysis of Emotional Expression

- Analysis of Human Dyadic Interactions

- Design of Emotion-Sensitive Human Computer Interfaces and Virtual Agents    &  

12 hours

- Facial expression and head and hand movements + the audio recordings of the conversations  &  Function with variations in Expression  

&         - Vicon with 53 facial markers, 3 markers for each hand, and 2 markers on the head      &          Facial Mocap, Head Movement and Orientation, Video, Speech, Dialog Transcript  &   10 &   -     &    Categorical and dimensional emotions

\\ \hline

MoDa \citep{MODADB}   &  
    - Studying movement and meaning based on Laban Movement Analysis
    
   -  Creating motion graphs
    
   -  Affect-expressive movement generation
    &   
   Collection of Datasets: 

- Affect-Expressive Motion Graph

- LMA Basic Effort Actions 

- Knocking and Direction Gestures with Variation form LMA

- Grooving
&  Function, Expression, Planning    
&        -  Vicon with 53 markers and 26 joints      &   Marker Data, Skeletal Data, Video, (Audio and Physiological, for some datasets) &         Varies (1 to 3 performers)          & 1 to 4 repeatitions across function and expression  &    Laban  Movement Analysis, Dimensional Emotions

\\ \hline

NUS DB \citep{NUSDB}  &   \textit{Unspecified}  & 

- Locomotion

- Interaction with Obstacles

- Martial Arts

- Dance

- Yoga   &   Function with planning variations 
&         - Vicon       &          Marker Data, Skeletal Data  &   8    &    -     &    -
\\ \hline

UPenn DB \citep{UPENDB}  &  

- Multi-Actor behaviours

- Diverse personalities

- The effects of posture and dynamics on the perception of emotion

- Study human fatigue     &  Collection of  multimodal datasets:

- Walking

- Emotional Actions

- Emotional Body Language

- Exercise  &    Function and Expression     

&         \textit{Unspecified}      &          Marker Data, Skeletal Data, ForcePlate, Biological Data  &  Varies             &    -     &    Categorical Emotions

\\ \hline

\citep{Yingliang:B9Rp6CQI}  &   
Study of identity, gender, and emotion perception 
    & 
    4080 sequences 
    
    - Walking
    
    - Knocking
    
    - Lifting
    
    - Throwing   &     Function and Expression    
    &         Falcon Analog, 35 markers, 15 joints       &       Marker Data, Skeletal Data  &   30 & 5 repetitions     &    Categorical Emotions
    \\ \hline

NTU RGB+D \citep{Shahroudy:ch}    &   
RGB+D human action recognition 
    & 60 action classes within daily actions, health-related actions, and inter-personal actions  &     
    Function  
    &        - Three Microsoft Kinect V2 devices       &       Video, Joint Positions  &   40 & 
    2 instances for each actions (two different angles from the camera)     &    
    Action labels
    \\ \hline



Berkeley MHAD \citep{Ofli:ky}  &   
RGB+D human action recognition 
    & 11 actions with high dynamics in:
    
    - Both upper and lower body

    - Upper body

    - Lower body
    &     
    Function  
    &  - Mocap: Impulse
    
    - Video: 12 Dragonfly2 cameras
    
    - Depth: 2 Microsoft Kinect V2
    
    - Acceleration: 6 three-axis wireless accelerometers on wrists, ankles, and hips       &      
     Joint positions, depth, multi-angle video, audio, acceleration  &   
     5 female, 7 male & 
     5 repetitions for each action    &    
     Action labels
    \\ \hline

Human 3.6M \citep{Ionescu:kt}   &   
    Human pose estimation

    & 15 actions within upper body movement, full-body upright variations, walking variations, sitting on the floor, and miscellaneous movements  &     
    Function  
    &        
     - Mocap: Vicom T40
     
     - TOF: Mesa SR4000
     
     - Video: Basler piA1000
     
     - Body Scan: Vitus Smart LC3       &      
     RGB video, depth, joint positions, 3D volumetric models of subjects  &   
     5 female and 6 male subjects & 
     One repetition     &    
     Action labels, body parts (for video)
    \\ \hline    
                        
\end{tabular}
\egroup
\end{adjustwidth}
\end{table*}

\textbf{Number of Subjects}
\label{sec:data:numb}
Human movement, in terms of its modulations and variations, depends on the personal movement signature of the performer. Each person has a different movement signature, which is influences by her or his genetics, habits, attitudes, values, and life history \citep{Studd:2013uc}. 
To learn a generalized model of movement, which is invariant to the performer-specific styles while recognizing the ``personal factor'' of movement, requires training the model using the data from multiple subjects.  

The majority of studies use a single subject in their training data, while some studies use two, e.g., \citep{wang_learning_2006,Taubert:2011jp,Taubert:2012gl,Alemi:2015kg}  \circledone{7} \circledone{22} \circledone{22} \circledone{28}, three subjects, e.g., \citep{wang_multifactor_2007,Chiu:2011do} \circledone{9} \circledone{20},  and in other cases 12 and 41 subjects, e.g., \citep{liu_human_2011} \circledone{18} and \citep{tilmanne_stylistic_2012} \circledone{23}, respectively. 

The number of subjects used in the studies is, to some extent, limited by the availability of the data from multiple actors, performing the same type of movements. This number varies in the publicly available training databases, which is discussed in the following section. 

\subsection{Movement Databases}
\label{sec:data:base}

The majority of the reviewed studies use the data from the Carnegie Mellon University Motion Capture Database (CMU mocap) or have captured their own data.  In addition to the CMU database, there are a number of other databases that are publicly available for research purposes and potentially can be used for movement generation. 
Key details of these movement databases are presented in  Table~\ref{tbl:dball} and are discussed in the following.

\subsubsection{Curation and Purpose}
Most of the databases are created and tailored towards a set of particular research questions. For example, the primary goal of the IEMOCAP, University of Glasgow, and AffectMe databases is towards the analysis of emotional expression,  while the University of Pennsylvania database is tailored towards modelling multi-actor behaviours. 
Databases such as the CMU, ACCAD, and HDM05 provide a wider and more general set of contents that are created to provide freely available motion capture data to the research community for a variety of purposes. 
MoDa, an open-source movement database, is a repository of multiple databases that address a range of movement-related research questions such as affect-expressive motion graphs, data tailored for Laban Movement Analysis research, and dance and music studies.

Some databases, such as Berkeley MHAD, NTU RGB+D, and Human 3.6M, are created for the research on human movement analysis and action recognition from image and video (RGB) data in the context of every-day human activities. However, they often include reference motion capture data which can be used for training generative movement models.

\subsubsection{Content and Size}
The size of the presented databases varies extensively both in terms of the length of the content and the diversity of the movements. 
While some databases provide a relatively large amount of motion captured data (e.g., Human 3.6M, CMU, IEMOCAP, and MoDa), others only have a few sequences (e.g., Cypress DanceDB).



\subsubsection{Characterization Diversity}
To effectively create systems that model and control variations across different dimensions of movement (as described in Section~\ref{sec:charac}), one has to have access to a training dataset that contains the desired variations.
The majority of the databases contain movements that vary across the function and expression dimensions, while a fewer number of databases contain variations across the planning dimension.

\subsubsection{Recording Technology and Files Formats}
The majority of the database use Vicon motion capture systems with reflective markers, some use mocap systems with infrared markers, and a few use inertia-based capturing systems. 

All of the databases provide the raw marker data in C3D format, and most provide skeletal data (joint angle rotations) in the form of BVH or AMC files. Multi-modal data sets are also available in some databases and provide video, audio, or physiological recordings that accompany the movement. 

\subsubsection{Capturing Modalities} 
Most databases provide the raw marker data, as well as the skeletal data. Some also accompany a video recording of the motion capture session for each movement as a reference. Few databases such as MoDa, IEMOCAP, and Berkeley MHAD provide other modalities such as voice, text, facial expression, and physiological measures. 

\subsubsection{Human Subjects}
Every human has a distinct movement signature and style \citep{Studd:2013uc}. The more the number  of subjects in a training dataset, the better is the model’s ability to distinguish between these personal modulations in the data and the underlying patterns that is common among the movements of all performers. 

The number of subjects varies from 1 to 144 in the reviewed databases. 
Note that in some cases, the same movement may not be available for all of the subjects.
For example, in the CMU database the same movement is only repeated by very few subjects, rather than the whole 144 movers. On the other hand, databases such as IEMOCAP or MoDa ensure that the same movements are consistently performed by all subjects.

\subsubsection{Repetitions and Motor Variation} 
In most machine learning problems, including learning generative movement models, having more variations of the data increases the robustness of the model towards the variations that the model faces in real-world applications and avoids overfitting the model to a limited set of input. While many databases provide no or very few repetitions, databases such as HDM05, MoDa, and the University of Glasgow provide multiple repetitions of the same movement.

\subsubsection{Annotations}
In the reviewed databases, annotations mostly include the categorical emotions (as in IEMOCAP,  University of Pennsylvania, University of Glasgow, and AffectMe) and the dimensional affect representations (as in IEMOCAP and MoDa). Databases in MoDa  also include annotations based on the Laban Movement Analysis \citep{bartenieff_body_1980}.


\section{Learning and Generation}
\label{sec:synthesis}

In this section,  we analyze the learning and generation methods that are applied to the motion capture data.
We organize our analysis based on the machine learning families,  namely dimensionality reduction techniques, Gaussian processes, hidden Markov models,  artificial neural networks, as well as a few other machine learning approaches.

\begin{center}

\begin{table*}[t!]
\begin{adjustwidth}{-1.5cm}{-0cm}
\bgroup
\def\arraystretch{1.6}
\scriptsize
 \centering
 \caption{Machine Learning Methods for Movement Learning and Generation }
 \label{tlb:ml}
 \begin{tabular}{L{3cm} L{0.2cm} L{5cm} L{1.5cm} L{3cm} L{3cm} }
 \hline
 Machine Learning Family & & Model & Details & Factorization Technique & Remarks  \\ 
   \hline
   
   \multirow{3}{*}{Dimensionality Reduction}

	& \circled{12} & Isomap embedding \citep{qu_motion_2008} & Modelling dynamics with LDS &  & \\

	& \circled{16} & Principal Component Analysis (PCA) \citep{tilmanne_expressive_2010} & &  Principal Components &    \\

  &\circled{24} & Functional PCA + Gaussan Mixture Model  + Gaussian Process \citep{Min:2012jy}  &  & Graphs + Optimization &  \\  

  & \circled{25} & Functional PCA \citep{Samadani:2012ki} &  & Principal components + clustering & \\

  &\circled{32} &  Functional PCA + Gaussian Mixture Model + Gaussian Process + kMeans Trees \citep{Herrmann:2017hj}  &  & Graphs + Optimization &  \\
   \hline

   \multirow{3}{*}{Gaussian Process Models}

 &\circled{9} & Multifactor Gaussian Process Models \citep{wang_multifactor_2007} & Dynamic model & Latent space \\

  &\circled{10} & Gaussian Process Dynamical Models (GPDM) \citep{Wang:ia} &  & Dynamic model &  \\

  &\circled{22} & Guassian Process Latent Variable Models (GPLVM) \citep{Taubert:2011jp,Taubert:2012gl}  &  Modelling dynamics with HMM & Individual models & Modelling two-character handshake \\

 
 \hline
  
   \multirow{3}{*}{Hidden Markov Models} 

 &\circledone{1} & Stylistic HMM \citep{brand_style_2000} & 75 states & Parametric Gaussian & Unsupervised learning of movement factors \\ 


 &\circled{4} & Multiple Regression Hidden Semi-Markov Models \citep{yamazaki_human_2005}& 5 states & Parametric Gaussian & Modeling walking pace and stride length\\

 &\circled{5} & Hierarchical HMM \citep{wang_learning_2005} &  &   & Hierarchical model - Using DTG \\

 &\circled{7} & HMM/Mix-SDTG \citep{wang_learning_2006} & 4 states  & Parametric Gaussian & Using mixture of SDTGs   \\ 

 &\circledone{13} & Parametric HMM \citep{herzog_parametric_2008} &  20 states & 1.Interpolating individual models 

2. Parametric Gaussian & \\

 &\circled{23} & Hidden Semi-Markov Models \citep{tilmanne_stylistic_2012} & 5 states &Average model + Individual stylized models + Transformation algorithms  & Learns a neutral walking model which can be adapted to different styles \\

 \hline

 \multirow{3}{*}{Others}

&\circledone{3} & Linear Dynamic System \citep{li_motion_2002} &‌ &  &  Motion Texture\\

&\circled{18} & Multilinear Independent Component Analysis \citep{liu_human_2011} & & Optimization  & \\

 \hline
\end{tabular} 
\egroup
\end{adjustwidth}
\end{table*}

\end{center}


\begin{center}
  
  \begin{table*}[t!]
    \begin{adjustwidth}{-1.5cm}{-0cm}
    \renewcommand\thetable{IV}
  \bgroup
  \def\arraystretch{1.6}
  \scriptsize
   \centering
   \caption{Machine Learning Methods for Movement Learning and Generation  - \textit{Continued}}
   \label{tlb:ml2}
   \begin{tabular}{L{3cm} L{0.2cm} L{5cm} L{1.5cm} L{3cm} L{3cm} }
   \hline
   Machine Learning Family & & Model & Details & Factorization Technique & Remarks  \\ 
     \hline
     \multirow{3}{*}{Artificial Neural Networks}

     &\circled{6} & Self-Organizing Mixture Network or SOMN \citep{wang_key-styling:_2006}   & 1 layer & Parametric Gaussian &  \\

     &\circled{8} & Conditional RBM (CRBM)  \citep{taylor_modeling_2007}& 1 and 2 layers &  & Unsupervised learning\\

     &\circled{11} & Feed-Forward Network \citep{lin_applying_2008}  & 1 layer & Regression &  Lifting movement\\

     &\circled{14} & Factored CRBM  \citep{taylor_factored_2009}& 1 layer & Controlling network weights & Supervised learning and control \\

     &\circledone{20} & Hierarchical FCRBM  \citep{Chiu:2011uv}& 2 layers & Controlling network weights &  Interpolation\\ 

     &\circledone{21} & Hierarchical FCRBM  \citep{Chiu:2011do}& 2 layers & Controlling network weights &  Gestures controlled by audio\\ 

     &\circled{27} & Encoder-Recurrent-Decoder \citep{Fragkiadaki:2015vx}  & 1 encoder, 2 recurrent, and 1 decoder layers  &  & Learning the representation of posture using fully-connected layers at the same time as training the recurrent layers \\

     &\circledone{28} & Factored CRBM  \citep{Alemi:2015kg}& 1 layer & Controlling network weights &  Supervised learning and control of affect expression\\

     &\circledone{29} & LSTM - RNN \citep{CrnkovicFriis:2016vx} & 3 layers &  &  Kinect Data / Unsupervised dance generation \\

     &\circledone{30} & Convolutional Autoencoders + Feed-Forward Network \citep{Holden:2016bv}  & 5 layers & Training a control network & Semi-supervised learning \\

     &\circled{31} & Seq2Seq with Adverserial Learning \citep{Wang:2017bs}  &  1 layer  & Conditional inputs & Adverserial Learning \\

     &\circled{33} & Seq2Seq with GRU RNN \citep{Martinez:2017ta}  &  1 layer &  & Learning velocities using a residual architecture \\

     &\circledone{34} & Factored CRBM  \citep{Alemi2017_WalkNet}& 1 layer & Controlling network weights &  Supervised learning and control of affect expression and navigation\\

     &\circledone{35} & Factored CRBM  \citep{Alemi2017_GrooveNet}& 1 layer & Controlling network weights &  Music-driven dance generation\\

    

    \hline
  \end{tabular} 
  \egroup
\end{adjustwidth}
  \end{table*}

  \end{center}

\subsection{Generative Dimensionality Reduction}
\label{sec:synth:dr}

As we saw in Section \ref{sec:data:dim}, dimensionality reduction (DR) techniques are used to derive a smaller feature vector from a high-dimensional dataset. Smaller feature vectors are effective to reduce the redundancies in the data, reduce the memory usage during the training process, and  increase the learning speed of other machine learning models (e.g., HMMs). In this section, we look at another application of DR techniques that aims to directly generate movement animation.

Dimensionality reduction techniques map high-dimensional movement data, either in the form of single poses or windows of consecutive poses, to a lower-dimensional representation.
Depending on the DR technique, this mapping can be bi-directional: not only we can transform movements to the DR space, it is also possible to reconstruct movements, with some information loss, from a given representation in the DR space. As different points in the DR space correspond to different movement qualities, by choosing a point in the DR space and mapping it back to the high-dimensional space we can generate movements.  In the following, we review the application of Isomap and variations of the  Principal Component Analysis in generating movements.

\subsubsection{Isomap} 
Isometric feature mapping (Isomap) is a non-linear dimensionality reduction technique that constructs a graph connecting the nearest data points. The graph is then used to create a lower-dimensional representation that preserves 
the geodesic distance\footnote{Determined by the number of nodes on the shortest path between two nodes on the graph.} between all data point pairs.


A method proposed by \citet{qu_motion_2008} \circledone{12} combines Isomap and linear dynamic systems (LDS). After  low-dimensional representations of the training data are extracted using Isomap, they are segmented using dynamic models that are trained on the same data. The resulting segments are considered as the  basic units of movement, which can be assembled to create longer sequences.
The method calculates the transition matrix between the segments with the assumption that the segments satisfy a first-order Markov chain constraint.
For generating new sequences, first, a low-dimensional representation is created by making noise-driven transitions between the segments. The resulting sequence is then mapped back to the high-dimensional pose space, which produces the generated movement segment.

\subsubsection{Principal Component Analysis}

\citet{tilmanne_expressive_2010} \circledone{16} use Principal Component Analysis (PCA) to generate expressive walking movements using a dataset of walking cycles with different gait styles.
First, the walking sequences are segmented into individual walking cycles, and cycles are normalized using the SLERP interpolation algorithm \citep{Shoemake:1985gp} to have the same length.  
To allow for interpolation between each frame, joint rotations parameterized by Euler angles are first converted to quaternions.

The resulting fixed-length vectors are converted into exponential maps, which are locally linear and more suitable for PCA. After performing PCA, 
empirical experiments by the authors show that the first 23 principal components account for $90\%$ of the variations in the data, in such a way that the reconstruction data is visually similar to the original data. 


Generation is performed for each gait style individually. First, the values of new data points for the Principal Components (PCs) are calculated for each cycle of the walk. Cycles are concatenated and smoothed in the PC subspace. The sequence is then transformed back from the PC subspace to the pose space, parameterized in exponential maps. The generated movement, after conversion to the quaternions, is resampled to the correct duration using the SLERP algorithm.

\subsubsection{Functional Principal Component Analysis}
\citet{Samadani:2012ki} \circledone{25} use functional statistics for extracting a set of movement features that are most salient to the expression of affect. 
First, similar movement segments are aligned with each other and converted into fixed-length vectors using piece-wise linear resampling. These fixed-length vectors of movement are then decomposed into temporal functions using basis function expansion (BFE) \citep{Ramsay:2013vf}. 
Next, using functional principal component analysis (FPCA), the BFE representation is transformed into a set of low-dimensional features that are suitable for discriminative analysis of affective movements. resulting in a low-dimensional  representation that is used for both tasks of affect recognition and movement generation.

Affect recognition can be implemented by using any simple classifier trained on the low-dimensional subspace that is the product of the FPCA. Movement generation is performed by using the centroids of the clusters (that represent the classes of movement) in the FPCA subspace as the representatives of each class. Next, the FPCA functional feature transformations are used to reconstruct the high-dimensional features in the pose space. These features are then linearly resampled to the average length of the original movements in the corresponding movement class.


\citet{Min:2012jy} \circledone{24}  propose an approach that creates a finite directed graph of generative  movement models. It differs from similar graph-bass approaches for movement generation \citep{Kovar:2002dq, Arikan:2002fg, Heck:2007ea} in that this approach uses generative models rather than the recorded data for each node or transition.

Building the graph follows a series of procedures: first, the data is decomposed into segments representing movement primitives.  The segments that represent the same movement primitive are aligned using dynamic time warping.  Functional PCA is then applied on the aligned representations of each primitive.  Next, output of the FPCA is modelled using Gaussian Mixture Model (GMM) to create a generative model for each primitive (graph nodes). Finally, the transitions between each node is learned using a Gaussian Process (GP) model.

The approach requires each movement primitive to be annotated with the environmental contact information, which are used in the generation processes to constrain the model to generate movements that follow a user-defined contact specification.

For movement generation without using any control parameters,  a two-step procedure is followed. First, the high-level structure of the movement is generated through a random walk over the graph. Next, movement segments for each node (model) is generated by probabilistic sampling over the movement parameters. The transitions between each node/segment are created using a blending approach introduced by \citet{Rose1998} to reduce any discontinuities around the transition points. 

For control over the generation, an approach based on graph walks, probabilistic sampling, and gradient-based optimization is devised by formulating the problem as a Maximum A Posteriori (MAP) framework to find a 
posteriori distribution defined over three terms: transition, contact-awareness, and control at the kinematic and semantic levels. 

Given the initial node, the generation algorithm first evaluates each  possible transitions  from the current node in the graph using the GP model that is trained before. The contact term measures the distance between the generated contact points and the target contact point. To support semantic control,  semantic commands such as ``picking up an object at a particular location'' are mapped to the proper graph node (e.g., `pick-up action'), and the proper kinematic parameters (e.g., the contact location). The control term also allows defining kinematic control parameters by using the likelihood of a forward kinematic function that maps the control parameters to their corresponding movements. 


\citet{Herrmann:2017hj} \circledone{32} extend the work by \citet{Min:2012jy} \circledone{24} by using a k-Means tree to speed-up the optimization process.
Similar to \citet{Min:2012jy} \circledone{24}, each pre-processed movement segment is mapped into a fixed-length, low-dimensional representation by applying Functional PCA. These fixed-length segments are then modelled by a Gaussian Mixture Model (GMM).
For generating a new movement segment, one has to draw a new sample from the GMM and back-project the low-dimensional representation into the movement space.

To speed up the optimization step during the generation process, rather than performing a brute force search over the position, orientation, and pose constraints, this approach takes advantage of the observation that the data in the latent space created by the GMM forms clusters of similar movements. 
Based on the described clusters, the latent space is partitioned hierarchically by recursively applying the k-Means++ algorithm \citep{Arthur2007}  on the data.

To generate movements, the tree is traversed to find the optimal sample using an objective function that is specified by the user-defined constraints. 


\subsection{Gaussian Processes}
\label{sec:synth:gp}

\subsubsection{Guassian Process Latent Variable Model}

\begin{figure}[t]
\centerline{\includegraphics[width=0.7\linewidth]{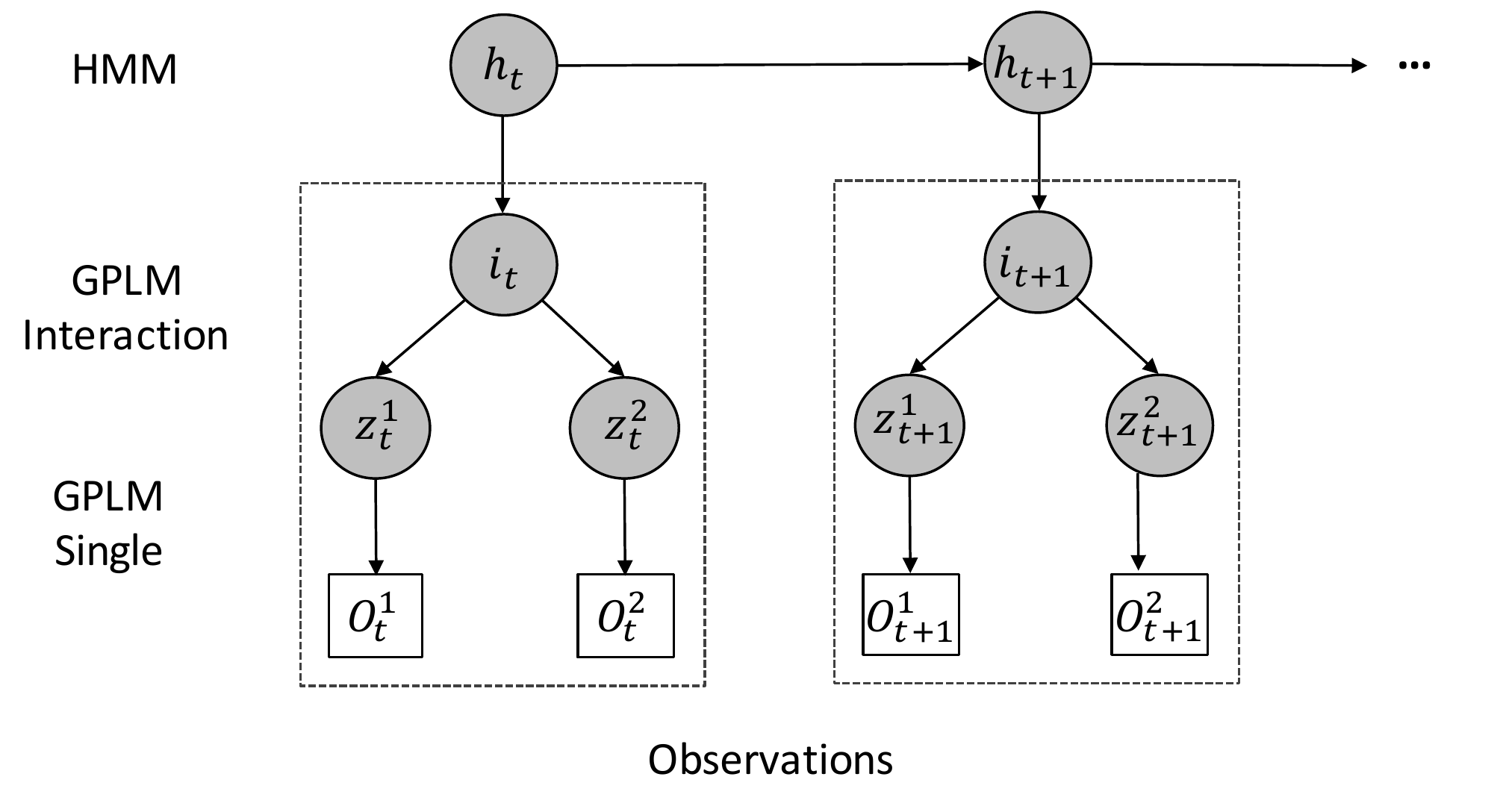}}
\caption{The three-layer model described in \citep{Taubert:2012gl}. In the static model, shown inside the box, at each frame $t$, the handshake data of two different actors ($O^1$ and $O^2$) are mapped into two separate latent spaces ($z^1$ and $z^2$). The latent representation of both actors are then combined and mapped into another latent space $i$, which represents the interactions between the two actor. In the dynamic model, a hidden Markov model with hidden unit $h$ is trained on the representation in $i$ to learn the dynamics of movement.}
    \label{fig:gplvm_hmm}
\end{figure}

Gaussian Process Latent Variable Models (GPLVM) learn the joint distribution of observations and their low-dimensional representation in a latent\footnote{Also called \textit{hidden.}} space. GPLVM can also be described as a non-linear dimensionality reduction method that generalizes the probabilistic PCA \citep{Tipping:1999uo}. 
In the method proposed by \citet{Taubert:2012gl} \circledone{22}, the model generates hand-shake movements for a chosen category (neutral, fearful, happy, angry, and sad). 
First, a hierarchical Gaussian process latent variable model (GP-LVM) maps the motion capture data of handshakes into a low-dimensional space. Next, a standard hidden Markov model (HMM) learns the dynamics of the handshakes from the low-dimensional space encoded by movement categories. 

As shown in Fig.~\ref{fig:gplvm_hmm}, the resulting model consists of three layers: the bottom layer is the GP-LVM-single, in which the movements of one individual actor are mapped onto a 3-dimensional latent variable while capturing the variations with respect to parameters such as actors, trials, emotional category, and time. The interaction layer (GP-LVM-interaction)  learns a 3-dimensional latent variable from a 6-dimensional observation variable that is created by the learned bottom-layer model for each pair of interacting actors. In the top layer (HMM-dynamic), a left-to-right HMMs with seven states learns the temporal evolution of the latent variable in the interaction layer. For each emotion category, a different HMM is learned. 

\subsubsection{Gaussian Process Dynamical Models}

\begin{figure}[t]
\centerline{\includegraphics[width=0.6\linewidth]{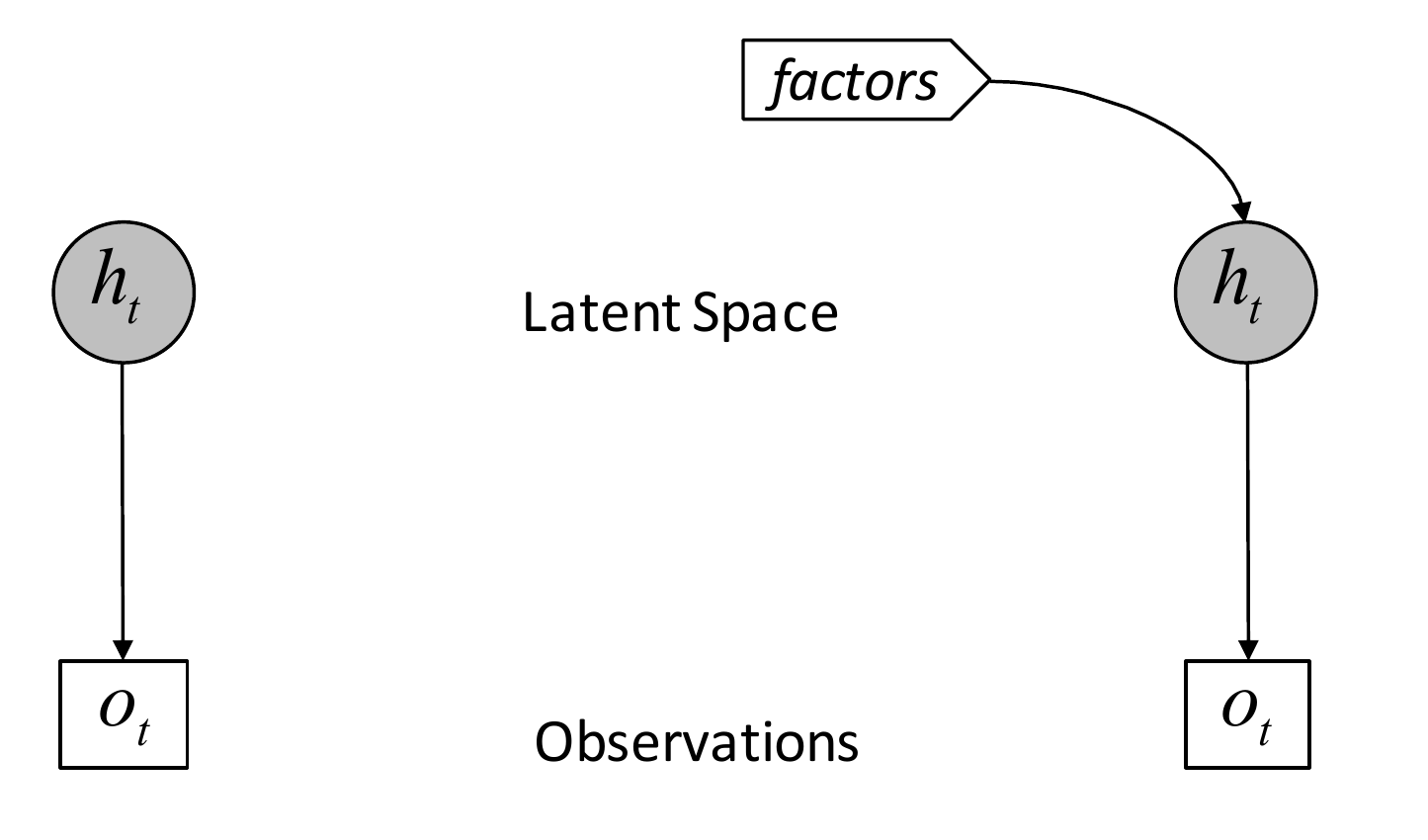}}
\caption{GPDM (left) and MF-GPM (right).}
    \label{fig:gp}
\end{figure}

\citet{Wang:ia} \circledone{10} extend the GPLVM with a dynamical model over its latent space to learn the temporal structures of sequences.  The resulting model provides generative mappings between the input observations  and a low-dimensional and sequential latent space (Fig.~\ref{fig:gp}).  GPDM can model first-order, as well as higher-order Markov chains, which can be used to learn the speed and acceleration of movement as well. Learning GPDMs involves performing numerical optimization to estimate the model parameters. Four different learning algorithms are examined by \citet{Wang:ia} \circledone{10}: Maximum A Posteriori (MAP), Balanced GPDM, Hand Tuning, and two-stage MAP, from which the Balanced GPDM algorithm performs the best.

\subsubsection{Multifactor Gaussian Process Model}
Multifactor Gaussian Process Model (MF-GPM), a special case of Gaussian process latent variable model, is proposed by \citet{wang_multifactor_2007} \circledone{9} to learn and generate cyclic locomotion.
The MF-GPM includes a low-dimensional latent space of multiple 	movement factors, as well as a mapping from the latent space to the high-dimensional observations (Fig.~\ref{fig:gp}). 
The MF-GPM  is capable of learning a factorized model, which allows it to generate movements with factor combinations that do not exist in the training data.

In this model, each pose in a movement segment is generated based on the combination of three independent factors: (1) the identity of the subject, (2) the gait (walk, stride, run), and (3) the current state of the movement (e.g., the walking phase). 
These factors are learned in a semi-supervised manner from the training data.
The movement is modelled based on the assumption that the identity and gait variables are fixed for each sequence, and only the movement state changes. While this assumption does not allow to observe transitions in the training data, transitions can be generated by interpolating the gait factors during the generation.


\subsection{Hidden Markov Models}
\label{sec:synth:hmm}
Hidden Markov Models (HMMs) are widely used in sequence modelling and speech synthesis \citep{Zen:2009ea}, as well as in learning and generating human movement.
A variety of statistical models are derived from HMMs, among them, left-to-right HMM, Hidden Semi-Markov Model (HSMM), Parametric HMM, and Hierarchical HMM are used for movement generation.

Classical Hidden Markov Models are trained using variations of the Expectation-Maximization (EM) algorithm, such as the Baum-Walch algorithm \citep{Rabiner:1989hs}. 
Generating new movements using an HMM involves creating a sequence of hidden states, and then generating the pose for each state by sampling from its probability distribution. Creating the hidden state sequence can be done manually, e.g., if they are associated with some meaningful notions such as a particular walking phase, or can be done automatically using the Viterbi algorithm \citep{Rabiner:1989hs}. 
The former approach allows for authoring the content of the movement explicitly whereas the latter generates a random sequence.


There are two approaches to using HMMs for learning and  generating movement: by learning a parameterized model of movement, and by modelling individual movement primitives. In the following, we review studies in each approach.

\subsubsection{Parameterized Movement Models}
Learning a single classical HMM is only efficient in learning an average model over the whole training dataset. Consequently, it does not allow modelling any variations of movement factors.
However, HMMs can be adapted to learn and generate parameterized data. We review how parameterized movements are learned by HMMs, followed by discussing the parameterized generation methods as used in the literature.

\begin{figure}[t]
\centerline{\includegraphics[width=0.7\linewidth]{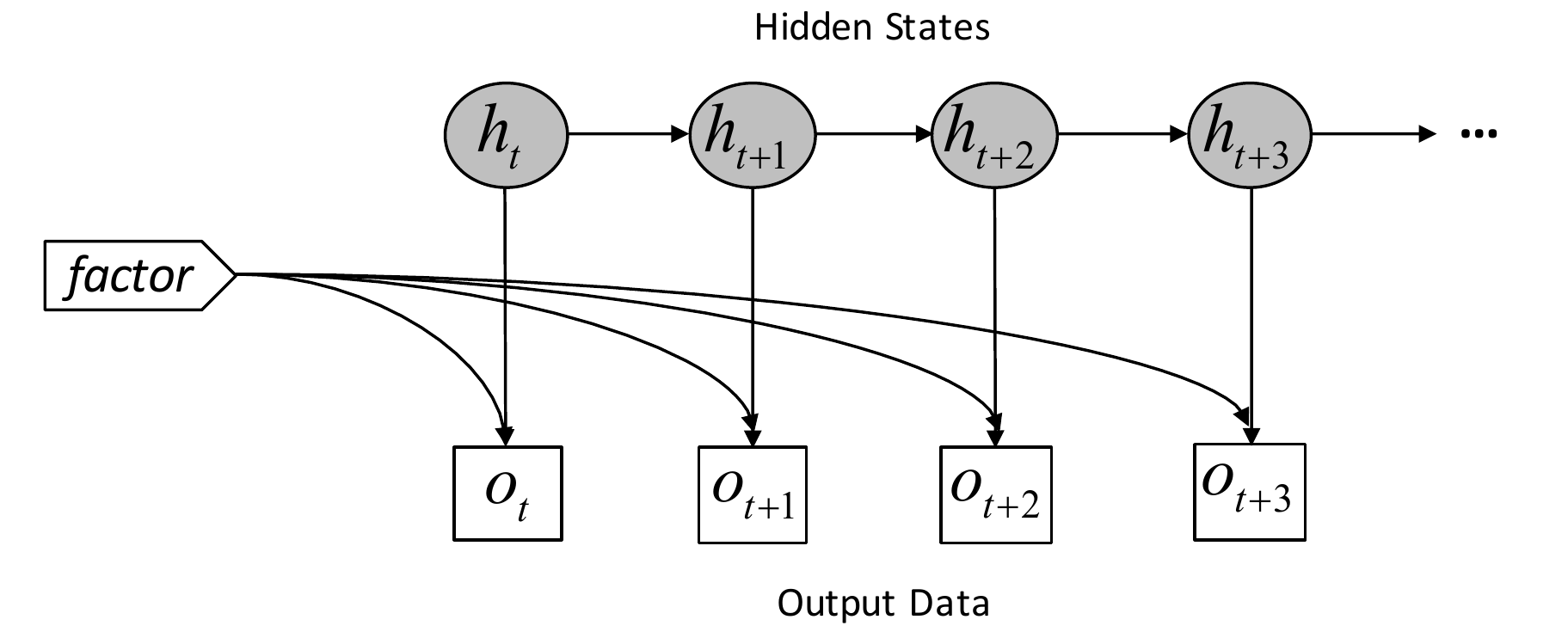}}
\caption{A Parametric HMM. The outputs are conditioned on the factor variable.} 
    \label{fig:phmm}
\end{figure}

Parametric Hidden Markov Model (PHMM) captures the variations in movement using a designated \textit{factor parameter} variable (Fig.\ref{fig:phmm}).
\citet{herzog_recognition_2009} \circledone{13} propose two  approaches to learn a Parametric HMM. The first approach uses Gaussian distributions with their means conditioned on the factor parameter. These PHMMs are trained using an extension of the Baum-Welch algorithm \citep{Wilson:1999dka}. 

The second approach trains a group of individual classic HMMs, one for each factor value in the training data \citep{herzog_parametric_2008} \circledone{13}, and uses component-wise linear interpolation of the means and covariance matrices of the observation distributions to derive a new HMM that generates movements based on the desired factors. 
In order for the interpolation to work, the states of all of the HMMs should be synchronized. This means that for each HMM state, there is an equivalent state in all other HMMs. As a result, the movements in the training data are aligned so that all the individual HMMs have the same number of states that point to the same parts of movement.

\begin{figure}[t]
    \centerline{\includegraphics[width=0.7\linewidth]{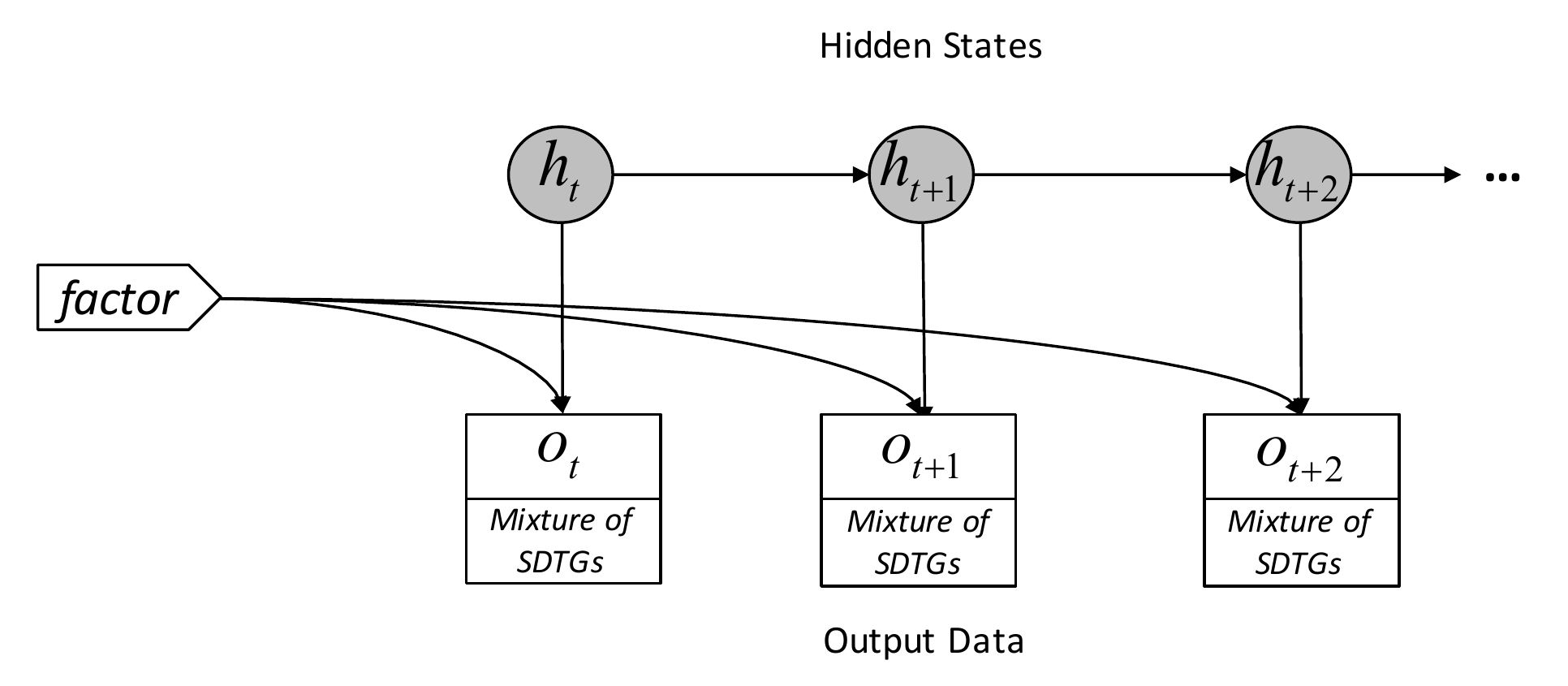}}
    \caption{The HMMIMix-SDTG model. 
    SDTGs are linear functions of the factor variables.} 
        \label{fig:hmmdtg}
    \end{figure}

\citet{wang_learning_2006} \circledone{7} introduce HMM/Mix-SDTG (Mixtures of Stylized Decomposable Triangulated Graph), an extension of HMM that uses Mix-SDTGs instead of the Gaussian distribution to model movement data. Similar to the Parametric HMMs,  SDTG incorporates a supervised variable to model the variations in the data, shown in (Fig.~\ref{fig:hmmdtg}). In HMM/Mix-SDTG, each observation (i.e., movement data) is conditioned based on the parameter variable, and the model is trained using a modified version of the EM algorithm. 

In a similar approach, the Stylistic Hidden Markov Model (SHMM) used in \textit{Style Machines}\citep{brand_style_2000} \circledone{1} is an HMM with its parameters (e.g., the means and covariances of the observation probability distributions) being functionally dependent on a factor  variable. In contrast to the Parametric HMMs, Style Machines simultaneously learns a generic HMM as well as a group of style-specific HMMs using an entropy minimization algorithm in an unsupervised manner. The generic model captures the movement mechanisms that are shared among all the styles. Each style-specific model then only captures a variation of the generic movement.
Using an optimization method via Expectation-Maximization (EM), the Style Machines automatically segment the data into movement primitives. Using this segmentation, Style Machines is able to learn similar primitives that are performed in different styles.

Style Machine uses a multidimensional style (i.e., factor)  variable that represents continuous values. Each dimension of the style variable is automatically discovered from the variations in the training data through an unsupervised learning algorithm as follows. First, a style space is constructed from the means vector of the generic HMM states, their square-root covariances, as well as their state dwell times (how long the model stays in one state before transiting to another state). Principal Component Analysis (PCA) is then applied on this style space to identify the dimensions of the space that explain the most variations in the data. Each of these dimensions represents a space of variation that exists in the training data and is used as the dimensions of the style variable. 


To generate new movements with HMMs that use a conditional probability distribution for their observations, first a sequence of hidden states is determined. Next, given the desired conditions a new pose is sampled from the probability distribution of each hidden state.

To generate new movements with Parametric HMMs \citep{herzog_recognition_2009} \circledone{13} that are built from a group of individual models, first the set of local HMMs that are closest to the desired parameters are identified. Next, a new HMM for the desired parameters is derived by interpolating the model parameters of the group of the chosen HMMs. Once we have the new HMM, the movements are generated using the Viterbi algorithm.

In the method proposed by \citet{wang_learning_2006} \circledone{7}, the first step to generate movements with an HMM/Mix-SDTG involves creating a sequence of hidden states based on the most likely state transition probabilities. Next, based on the given parameter, the output values of each hidden state is calculated. A $B$-spline curve is constructed based on the mean vectors of the joint rotations and the global position as the control points. New poses are generated by interpolating the points along the curve. The mean vectors of the dynamic features (i.e., the global and angular velocities) are used as the constraints for the local derivatives on the control points to ensure that the generated movement is smooth and continuous. 

Unlike other HMM-based models that use between 4-5 hidden states, Style Machines uses a relatively larger number of hidden states, about 69 states in the experiments performed by \citet{brand_style_2000} \circledone{1}. More hidden states allow the model to also learn a more diverse set of movements. During the training, the Style Machines automatically maps movement primitives to sub-sequences of the hidden states. One can then manually arrange these sub-sequences to choreograph new movements. One can also perform a random walk on the graph of the learned hidden state transitions in order to generate new movements. This is similar to the concept of motion graphs that was later introduced in the literature \citep{Kovar:2002dq}. 
With style machines, one can also regenerate existing movements with a different style by taking the state sequence identified from the existing movement and use a new style value to reproduce the same movement with the new style.


\subsubsection{Movement Primitive Learning}
\label{sec:synth:primitive}

The notion of \textit{movement primitive} is used to represent basic segments of human movements that constitute longer movements \citep{Schaal:2003ic}. The studies discussed in the following use this notion to break down the learning process by training the statistical models on shorter movement primitive segments rather than the whole movement. By concatenating these shorter segments differently, one can create different longer movements.

\begin{figure}[t]
\centerline{\includegraphics[width=0.7\linewidth]{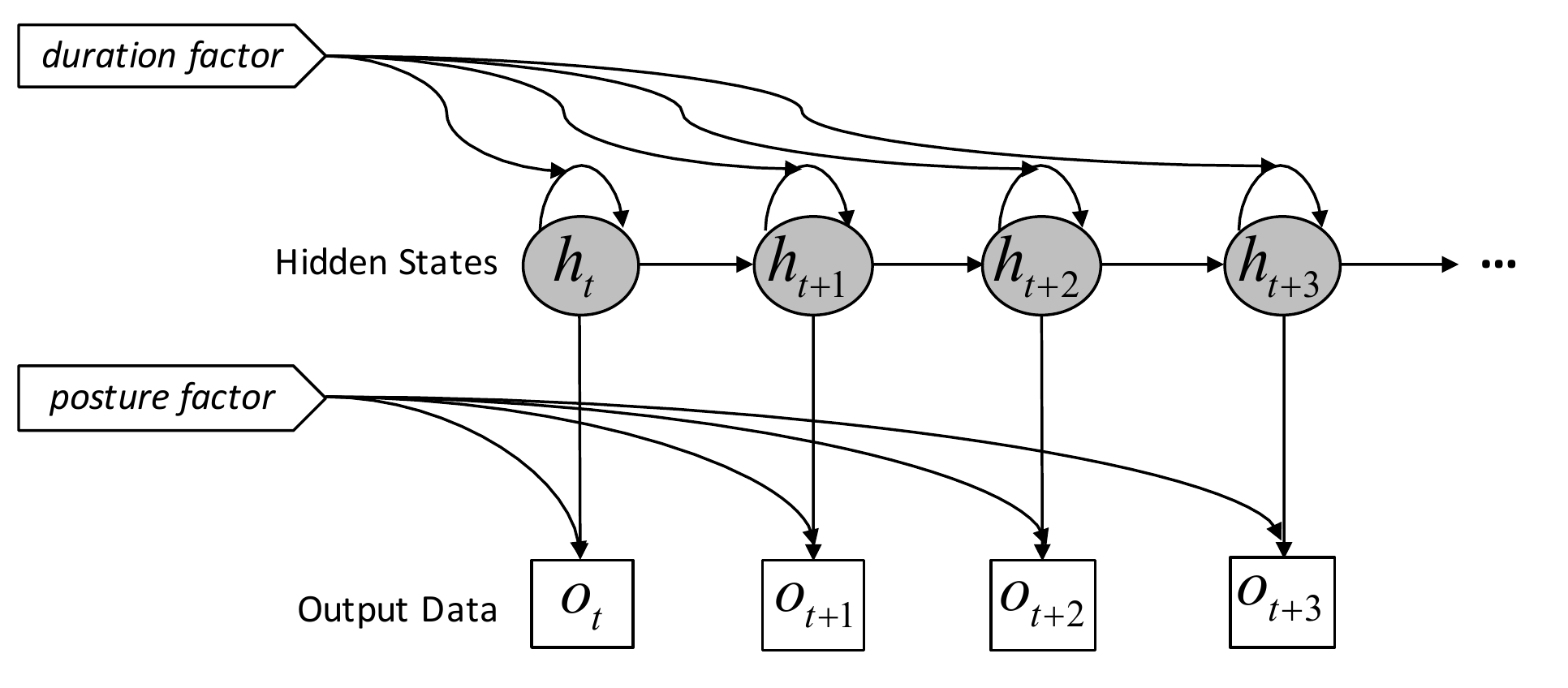}}
\caption{Hidden Semi-Markov Model with two factor variables: one affects the duration of hidden states and one affects the output data.}
    \label{fig:hsmm}
\end{figure}
\citet{yamazaki_human_2005} \circledone{4} use  Hidden Semi-Markov Model (HSMM)  to control the pace and the stride length of the generated movements.  Compared to the classic HMMs, HSMMs also capture the dwell-time of each of its hidden states, which makes them suitable for modelling the variations in the duration of movement.  To learn the state dwell-time parameters of an HSMM, the first and second derivatives of the joint rotations must also be added to the training data.

Unlike the Style Machines in which movement primitives are learned in an unsupervised manner, this approach is based on learning movement primitives based on manually segmented and labelled data. In addition, each movement primitive is modelled by a separate HSMM with fewer states than the Style Machines.  The model is based on decomposing walking cycle into four primitives. The L-step, which is the back-to-front movement of the left leg, and the R-step, which is the back-to-front movement of the right leg. In addition, two primitives for the beginning and the end of a walking cycle are also considered.

Generating movements with an HSMM is similar to the classical HMMs. First, for each HSMM, the most likely state sequence is determined by using Maximum Likelihood. Next, poses are sampled for each state \citet{yamazaki_human_2005} \circledone{4}. The pose sequences from all of the HSMMs are then concatenated in the correct order to create a movement. 

Another system introduced by  \citet{tilmanne_stylistic_2012} \circledone{23} is inspired by style-adaptive speech synthesis techniques. 
Similar to the approach used by \citet{yamazaki_human_2005} \circledone{4}, the walking sequences are manually segmented into shorter ones based on five stages that they define for each walking cycle.
In this method, a generic (average) walking model is learned from a relatively large number of training sequences.  
The style-specific HSMMs are created by transforming the parameters of the generic model to produce a particular walking style using linear transformations that are borrowed from speech synthesis applications \citep{Gales1998MaxLike,Yamagishi2009CSMAPLR}.

With the style-adaptive HSMM, the generation is performed through the HMM-based Speech Synthesis System (HTS) framework\footnote{ \url{http://hts.sp.nitech.ac.jp}}, which is designed for speech synthesis. 
First, HSMMs that correspond to a desired series of primitives are concatenated.   After determining the most likely state sequence for each HSMM, the poses for each frame is sampled using the Cholesky decomposition from the model parameters and the output features of the HSMM. The generation is completed by scaling back the data based on the data's global variance to avoid overly smoothed movements. 
The output of the model does not include the global displacement of the agent (i.e., the root location). Thus, the trajectory of the agent is calculated by identifying the points in time when the feet contact the ground. 

In a later work, \citet{Tilmanne:2014tx} \circledone{26} extend the previous approach. For the generation, the HSMMs are `unwrapped', i.e. the transition matrix is replaced with an explicit model. Next, to ensure the smooth trajectory of the output, the Maximum Likelihood Parameter Generation (MLPG) algorithm \citep{Tokuda:2000ew} is used over the strict Maximum Likelihood criterion.
This model allows for choosing the factors of the generated movements (styles), as well as interpolating models based on a weighted sum of the model parameters in order to blend, inhibit, exaggerate, or inverse the movements factors.

The HMMs described above only learn a single-layer model. In the following, we look at the methods that use hierarchical architectures to capture the hierarchical nature of human movement. 
The algorithm proposed by  \citet{tanco_realistic_2000} \circledone{2} builds a two-layer model of movement. First, the training mocap data is transformed into 15 principal components that are determined by PCA.
In the bottom layer, the data is clustered into multiple groups using the K-means algorithm. The clusters are then modelled using a Markov chain to learn the temporal relationships between the clusters. 
In the top layer, a discrete-output HMM models the higher level relationships. 
For generating new movements, first the two keyframes of the starting and ending poses are specified by the user and quantized using the K-means classifier. These keyframes correspond to the initial and final states of the Markov chain. Then a sequence of states between the initial and final states is determined by forming a synchronous sequential decision problem solved using dynamic programming. Finally, given the sequence and using the top layer HMM, the most likely hidden state sequence, and thus the set of consecutive movement segments are calculated using the Viterbi algorithm.

In a different approach, \citet{Kulic:2011hw} \circledone{19} model movement by segmenting the training data into movement primitives, clustering the primitives, and concatenating them into longer sequences for generating new movements.  Unlike other techniques which take an off-line learning approach, this work uses a method for learning movements from observations during an on-line, continuous process.   

In this method, movement is modelled in a hierarchical way. 
The data is segmented using a stochastic segmentation technique. The segments are incrementally clustered and organized into a hierarchical tree structure which represents movement primitives. If a new movement primitive is introduced to the model at any stage, a new cluster is formed to represent the new primitive. 
Each movement primitive is modelled with an HMM. At the top layer, the temporal relationships between movement primitives (i.e. their transition matrix) are learned through a hierarchical graph structure, called the ``movement primitive graph''. 
By performing walks on the movement primitive graph, and sampling primitives from the HMMs of the nodes in the walk,  one can generate continuous streams of movement.

\citet{wang_learning_2005} \circledone{5} use hierarchical hidden Markov models with non-parametric output distributions (NPHHMM) to create a hierarchical movement model: the top layer acts as a state machine describing the relationships between movement primitives, while the bottom layer models the sequences of poses that represent each primitive. Compared to the classical HMM which is based on a first-order Markov model, NPHHMM can capture longer temporal dependencies within the data.  

During the training process, first, the training sequences are segmented into movement primitives. Next, the segments are clustered using the EM algorithm, which also learns the transitions between each primitive cluster. In the final step, to movement in each primitive is learned by a hidden Markov model. Similar to the HMM/Mix-SDTG model, the output densities of the HMMs are modelled by the decomposable triangulated graphs (DTG).  In contrast, NPHHMM only models the functional factors of movement and does not support controlling the expressive or planning factors. 

With the NPHHMM, for any given user constraints, new sequences are generated by first synthesizing a path for the top layer of the model. This path represents the most likely behaviours with the given constraints. Next, for each behaviour defined in the top layer, a movement segment is generated. To ensure that the joint rotations in each frame are consistent with the adjacent frames velocities, the position of the joints are calculated by interpolating the frames.

\subsection{Artificial Neural Networks}
\label{sec:synth:ann}
In the following, we review studies that use artificial neural networks to learn and generate movement, discussing different types of the units and architectures they use.

\subsubsection{Feed-Forward Networks }
\begin{figure}[t]
\centerline{\includegraphics[width=0.7\linewidth]{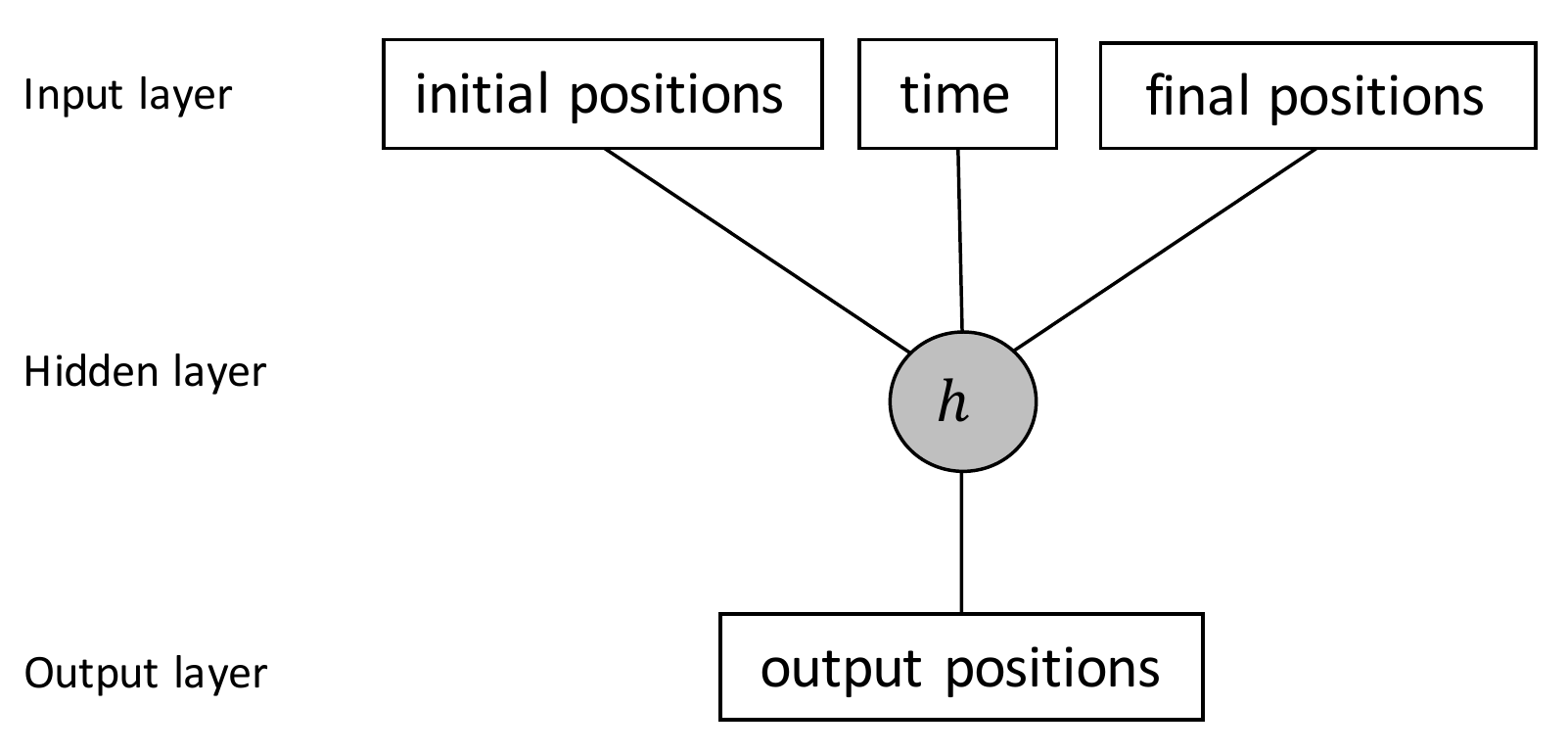}}
\caption{The architecture of the perceptron-based neural network using by \citet{lin_applying_2008}. } 
    \label{fig:lin_ann}
\end{figure}

\citet{lin_applying_2008} \circledone{11} use feed-forward neural networks in combination with optimization techniques for learning and generating the trajectory of a humanoid arm lifting objects. This perceptron-based neural network, as shown in Fig.~\ref{fig:lin_ann}, works as a function approximator for the angular positions. The network has one hidden layer with ten hidden units.  Its input layer consists of the frame number (time), initial joint positions, final joint positions, and the total number of frames. The output layer contains the angular position of the joints at a particular time in the lifting movement. 
The output is then applied to an optimization model to ensure that the initial and final positions match the desired values.   The network is trained using the back-propagation algorithm. 


\begin{figure}[t!]
    \begin{center}
    \begin{subfigure}[b]{0.35\linewidth}
     \includegraphics[width=\textwidth]{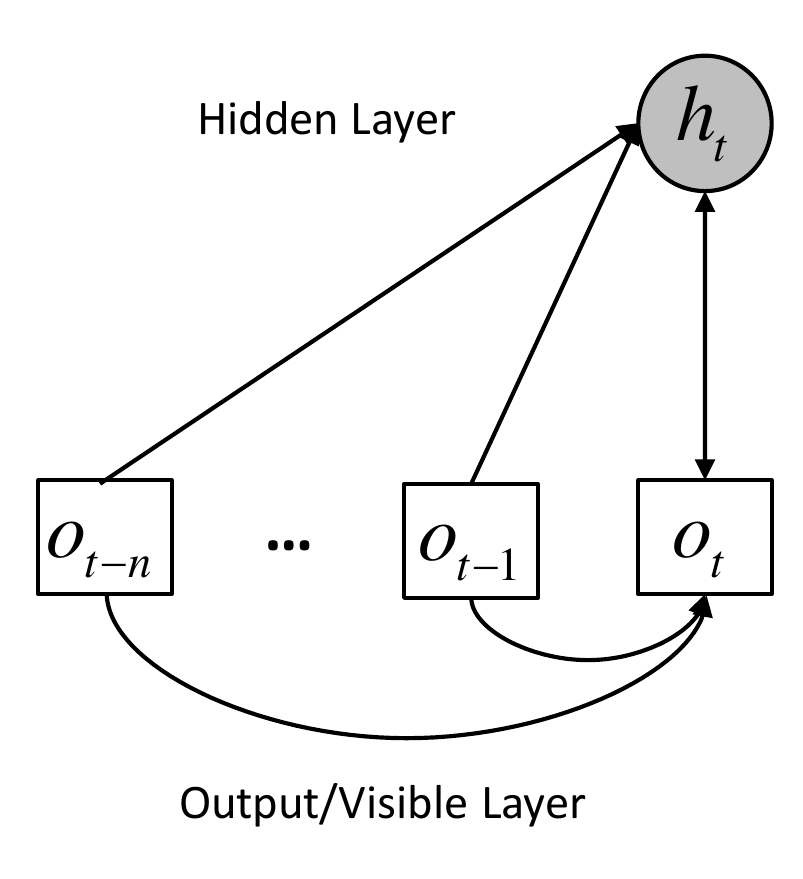}
     \caption{}
    \end{subfigure}
    ~
    \begin{subfigure}[b]{0.35\linewidth}
    \includegraphics[width=\textwidth]{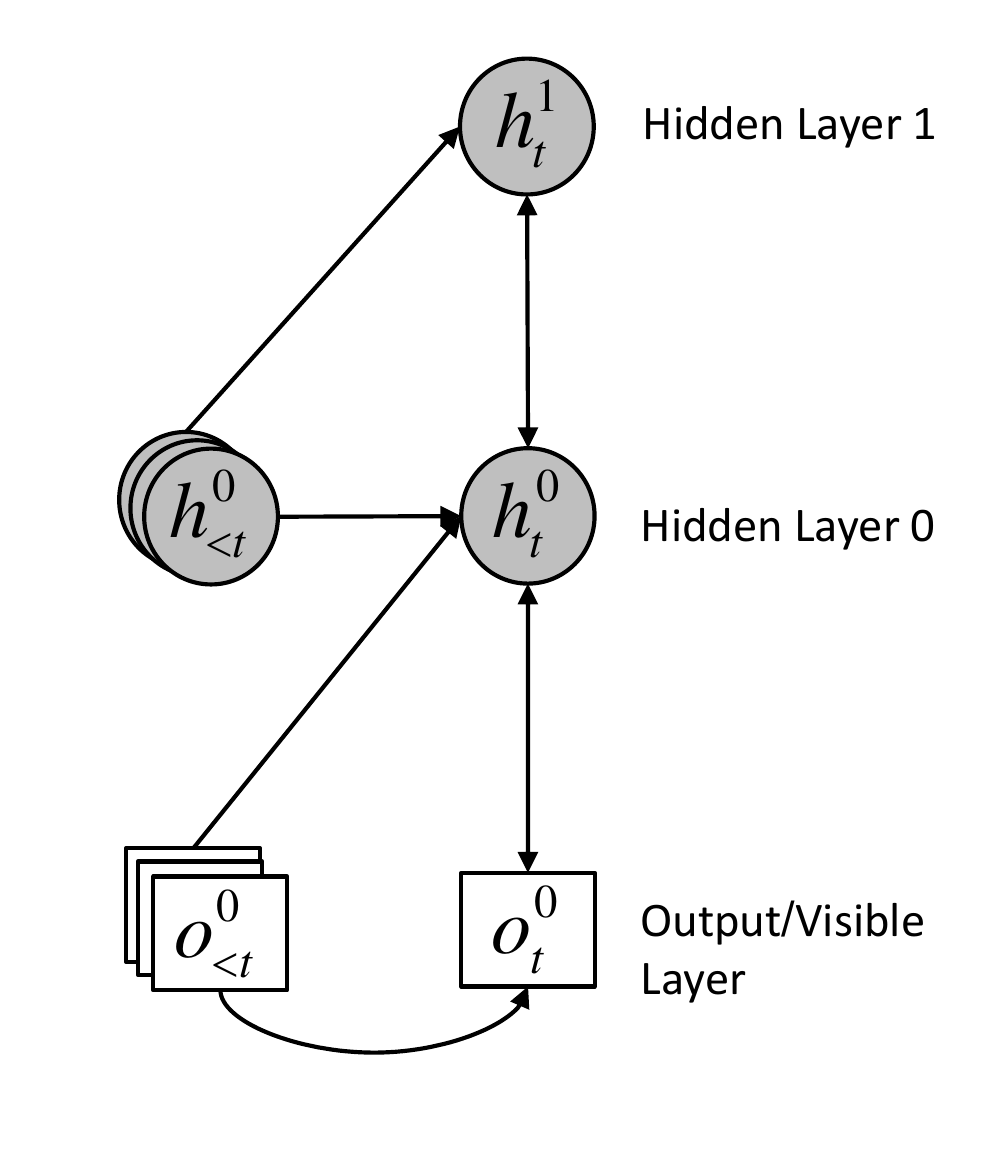}
     \caption{}
    \end{subfigure}
    \caption{(a) The architecture of a single-layer Conditional Restricted Boltzmann Machines (CRBM) with $n$ previous time steps as the conditional inputs. (b) A Conditional Deep Belief Network (CDBN) built from two CRBMs.  The $o_{<t}$ and $h_{<t}$ represent the data history vector $t-1,...,t-n$ where $n$ is the number of past time steps that are connected to the units at the current time step.}
    
        \label{fig:crbm}
   \end{center}
\end{figure}

\subsubsection{Self-Organizing Mixture Networks}
Self-Organizing Mixture Network or SOMN \citep{Yin:2001cg} is a type of artificial neural network that estimates mixture distributions using a self-organizing, unsupervised approach.
\citet{wang_key-styling:_2006} \circledone{6} use a SOMN of parametric
Gaussians and introduce an approach called \textit{key-styling} to  generate movement animations. Unlike the original SOMN, this model uses a conditional probability distribution to learn the effects of different values of movement factors. 
The model learns a probabilistic mapping from the low-dimensional space of factors to the high-dimensional pose space, while the mapping is controlled by a  style (factor) variable. 
The value of each dimension of the  variable is determined in a supervised manner via the annotations of the training data. 

For movement generation, one can specify a sparse sequence of key-style values (as opposed to keyframes).  The algorithm then interpolates the key styles into a dense sequence of style values, as it is expected to be more robust than interpolating the rotations in the pose space. Next,  a pose is generated for each style value using the distribution learned by the SOMN.


\subsubsection{Boltzmann Machines}

\citet{taylor_modeling_2007} \circledone{8} introduce a generative model for human movement based on Conditional Restricted Boltzmann Machines (CRBM). CRBM extends the Restricted Boltzmann Machine by adding conditional inputs to the model to capture temporal dependencies in the data (Fig.~\ref{fig:crbm}). 
CRBM can use an adjustable number of past data frames as conditional inputs at each time step and uses two extra sets of weights compared to the visible-to-hidden weights in the standard RBM: 
Autoregressive weights connecting the conditional inputs (past frames) to current visible units, which model the linear, temporally local structures, and weights connecting the conditional inputs to the hidden units, which model the non-linear and higher-level structures \citep{taylor_modeling_2007} \circledone{8}. 
CRBM learns the weights in an unsupervised manner using an adapted version of the Contrastive Divergence (CD) algorithm \citep{Hinton:2002ic}. 


It is possible to use CRBMs in layered architectures similar to Deep Belief Networks (DBN) \citep{Hinton_2006} and form a Conditional DBN to achieve more representational power (Fig.~\ref{fig:crbm}.b).

CRBM does not support an explicit representation of movement factors. A single model can learn and generate different movement functions (e.g., walking and running) if they exist in the training data. In this setting, the type of the movement to be generated is specified by seeding the model (a small number of frames used as the first set of conditional inputs). For example, if frames from a walking movement are used to initialize the model, it will generate walking movements and if frames from running are used, it will generate running movements. 

An updated version of CRBM \citep{Taylor_PhD_2009} uses soft-max labels as extra inputs to control the factors of movement during the generation. However, this technique determined to be inefficient as each hidden unit also receives many connections from the past and the current visible units, which diminishes the influence of the soft-max labels. 

\begin{figure}[t]
    \centerline{\includegraphics[width=0.5\linewidth]{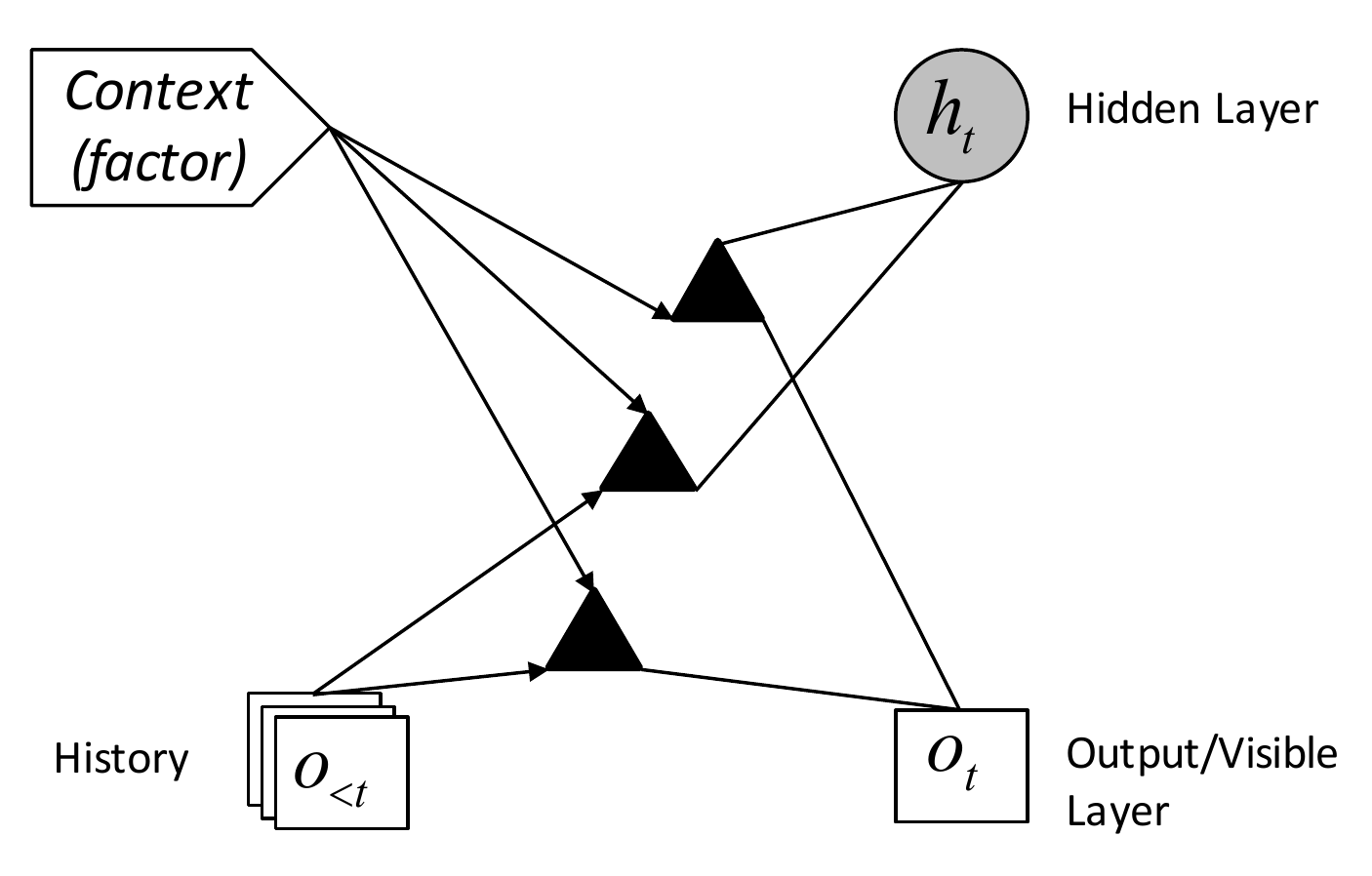}}
    \caption{The architecture of a Factored CRBM with interactions gated by real-valued
    stylistic features.} 
        \label{fig:fcrbm}
    \end{figure}

Further research in explicitly controlling movement factors using CRBM resulted in Factored CRBM (FCRBM) \citep{taylor_factored_2009} \circledone{14}, a model which supports  more representational capabilities and effective control over the generation. As depicted in Fig.~\ref{fig:fcrbm}, FCRBM uses three-way connections that allow a third unit (the context unit) to control the interactions between the visible and hidden units. Thus, the user can control the factors of the movements being generated \citep{taylor_factored_2009,Alemi:2015kg} \circledone{14} \circled{28}.

FCRBM supports a multidimensional discrete or continuous variable as the context unit, which allows it to capture and represent different factors of human movement. The interaction of the factors with the model is learned in a supervised manner using annotated data.
In addition, one can interpolate or extrapolate the factors to create new characteristics that did not exist in the training data. This generalization is demonstrated by \citet{Alemi:2015kg}  \circled{28} with generating movements with a full spectrum of affective states and transitions by only training the model on nine discrete states.

\begin{figure}[t]
    \begin{center}
 \begin{subfigure}[b]{0.4\linewidth}
  \includegraphics[width=\textwidth]{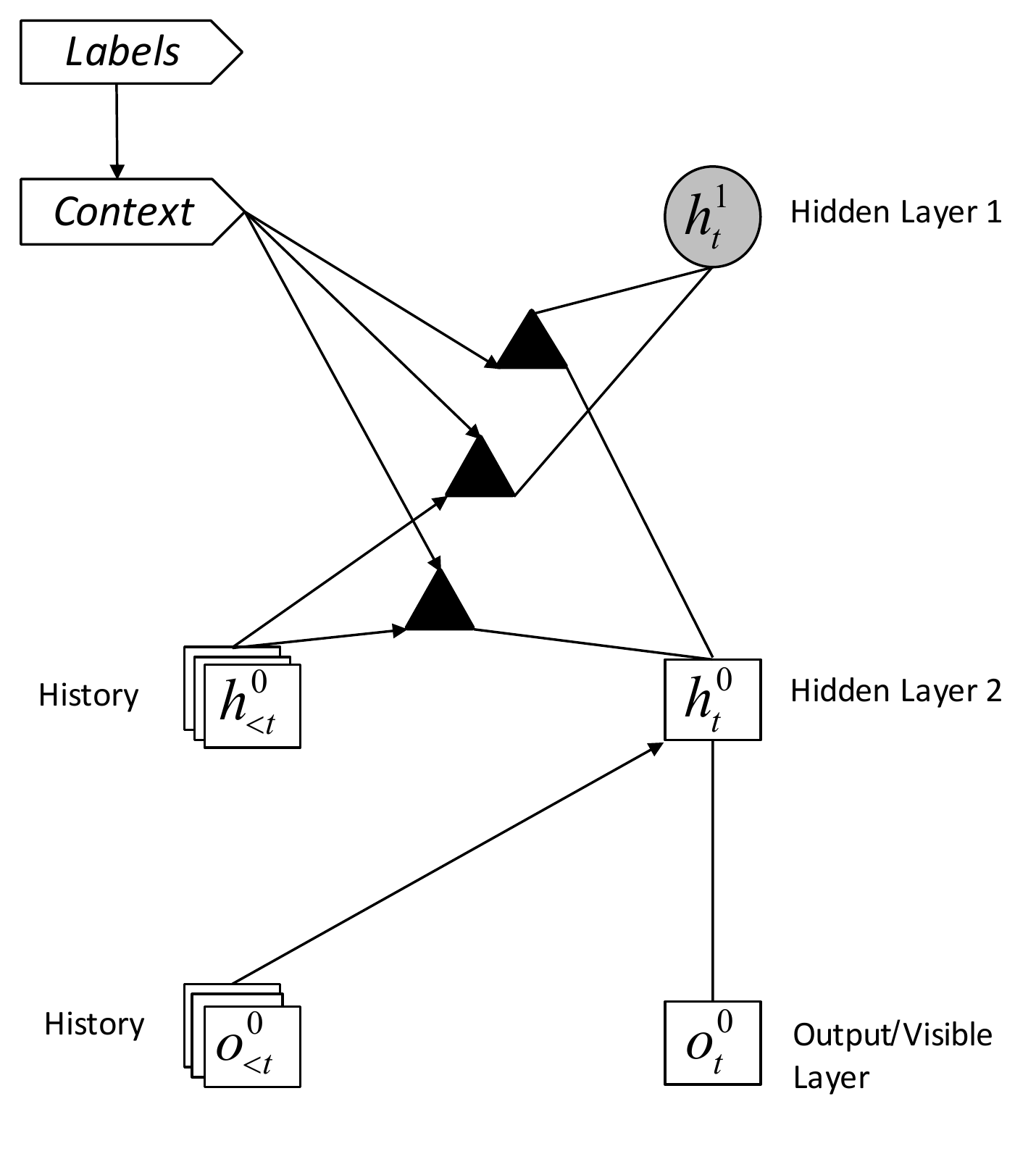}
  \caption{}
 \end{subfigure}
 ~
 \begin{subfigure}[b]{0.4\linewidth}
 \includegraphics[width=\textwidth]{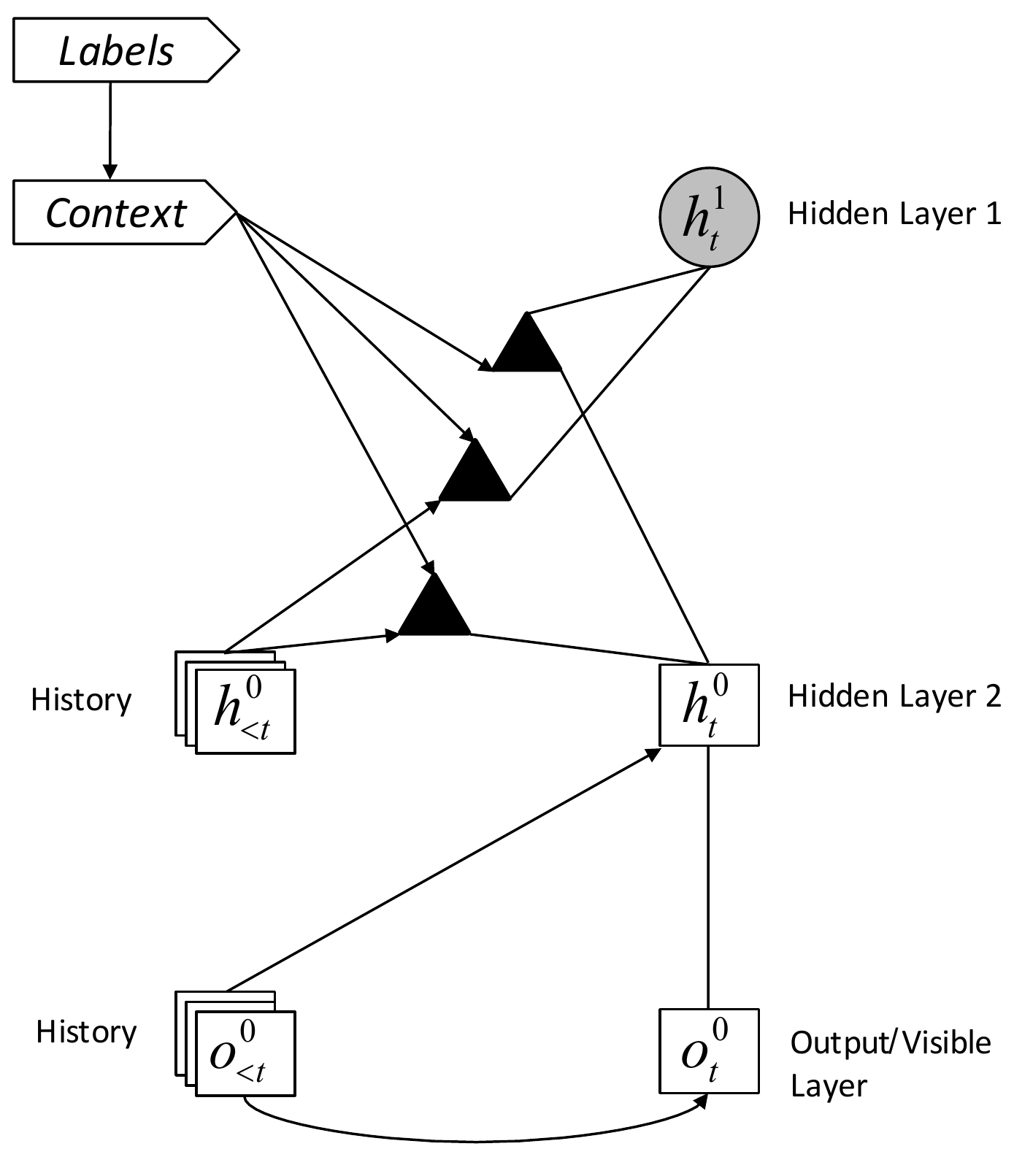}
  \caption{}
 \end{subfigure}
\caption{(a) The architecture of a Hierarchical FCRBM (HFCRBM) with a Reduced CRBM as the first layer and an FCRBM as the top layer. (b) A modified HFCRBM with a CRBM as the first layer and an FCRBM as the top layer.}
 
     \label{fig:hfcrbm}
\end{center}
 \end{figure}

A model for blending the factors of different movement segments is introduced by \citet{Chiu:2011uv} \circledone{21}. 
The model uses an extension of the CRBM called the Hierarchical Factored CRBM (HFCRBM).
An HFCRBM consists of a Reduced CRBM as its bottom layer and an FCRBM as the top layer (Fig.~\ref{fig:hfcrbm}.a). 
A Reduced CRBM is the same as a CRBM except that it does not include the autoregressive connections. In an HFCRBM, the input visible data are fed into the Reduced CRBM.  Once the Reduced CRBM is trained, an FCRBM is trained on the features discovered by the hidden layer of the Reduced CRBM as its input. This way of stacking the models together without the autoregressive connections ensures that during the generation, the visible data are only affected by the top layer and its labels and not the past visible data.

This model is designed specifically to interpolate different values of a factor using a procedure called the multi-path method.
To interpolate between two factor values, we first generate a sample for each value from the FCRBM. The generated samples are effectively two representations of the hidden layer of the Reduced CRBM. We then create a weighted sum of these two representations, and generate the corresponding new sample from the Reduced CRBM.

This type of factor interpolation is not as robust as directly interpolating the labels or the movements as it is not guaranteed that the Reduced CRBM would always produce a plausible movement. However, this method has a better chance of generating a novel factor value as the interpolation occurs in the latent space, and the final sample is generated through a non-linear process.


\citet{Chiu:2011do} \circledone{20} use a modified version of the HFCRBM (Fig.~\ref{fig:hfcrbm}.b) to generate gestures using the prosody of speech as the controlling factor. 
Using a set of training data that includes motion capture recordings of gestures accompanied by the voice recordings of the actors, 
the model learns the relationship between the prosody of speech and the movement.

Rather than directly learning the relationship between the audio features and the joint rotations, this approach uses the two layer architecture of the HFCRBM. The FCRBM portion of the model is trained on the features discovered by the hidden layer of a Reduced CRBM that is self is trained on the joint rotations of the upper-body.
The use of this two-layer architecture adds more non-linearity to the model,  and learns a more robust relationship between the audio features and movement. 

\subsubsection{Recurrent Neural Networks}


\begin{figure}[t!]
    \begin{center}
    \begin{subfigure}[b]{0.5\linewidth}
     \includegraphics[width=\textwidth]{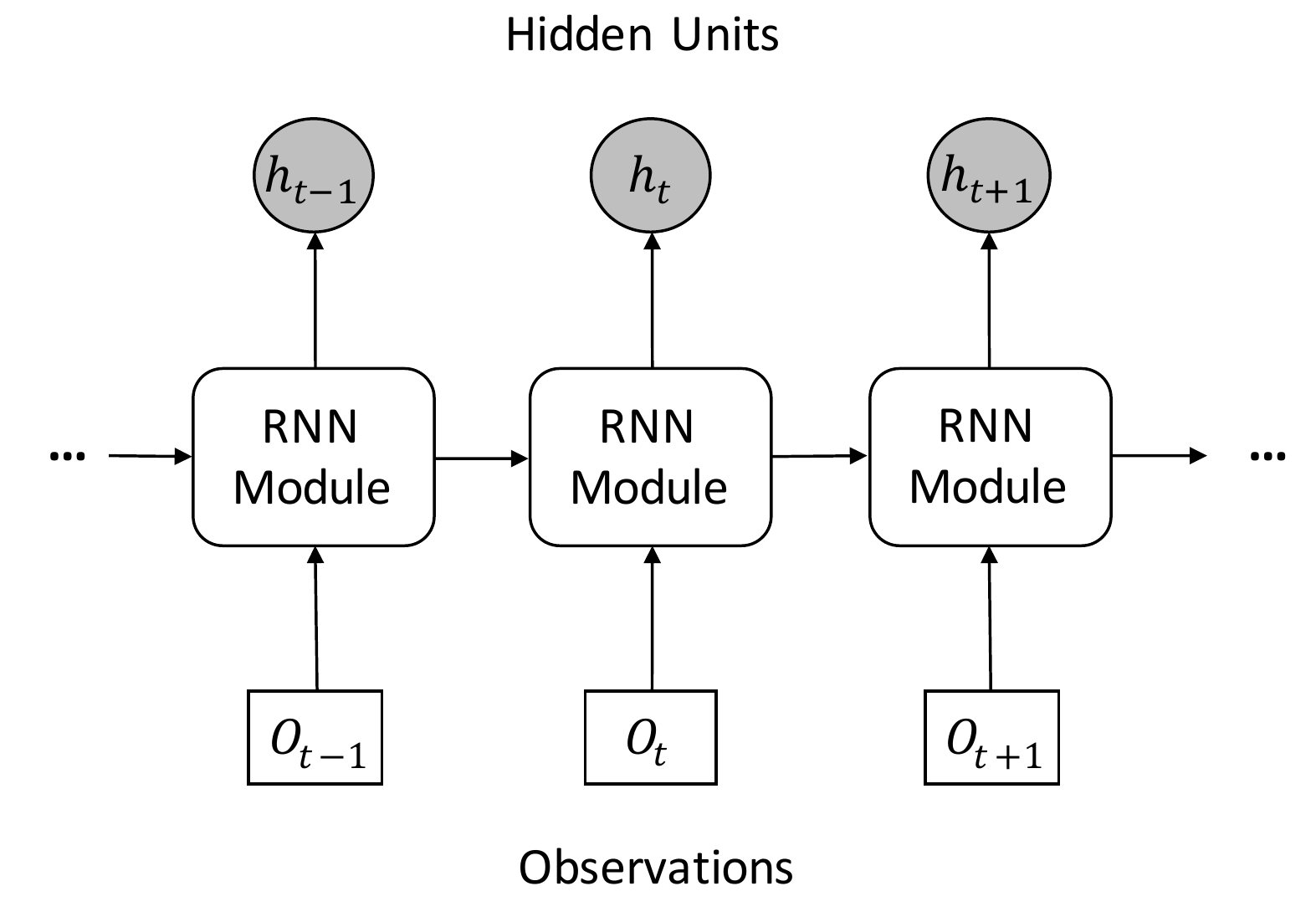}
     \caption{}
    \end{subfigure}
    ~
    \begin{subfigure}[b]{0.45\linewidth}
    \includegraphics[width=\textwidth]{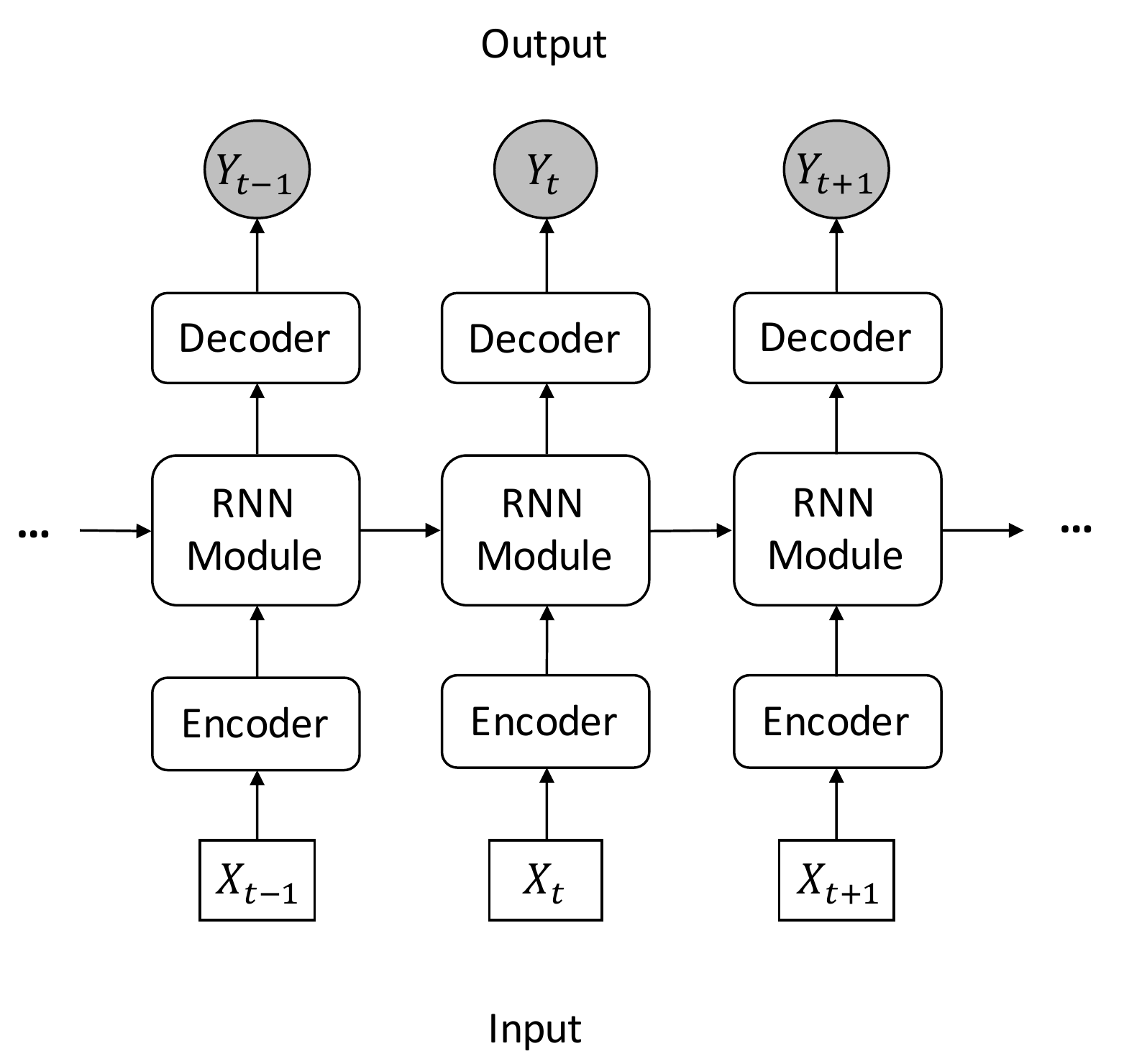}
     \caption{}
    \end{subfigure}
    \caption{Left: The architecture of a Recurrent Neural Network (RNN). In RNNs, a sequence of neural network modules repeat themselves. In standard RNNs, the module has a simple structure with a single neural network, while in an LSTM, the module contains four different neural networks interacting with each other and with the input/output. Right:
    The Encoder-Recurrent-Decoder architecture used by \citet{Fragkiadaki:2015vx},  }
    
        \label{fig:rnn-erd}
   \end{center}
\end{figure}


Recurrent Neural Networks (RNNs) are neural networks that apply the same operation on every input vector,  (Fig.~\ref{fig:rnn-erd}), with the output depending on previous operations and input vectors. 
By processing the input data sequentially and updating the weights accordingly at each step, RNNs create a form of `memory', which makes them suitable to model sequences with short and long-term dependencies. At the time of this writing, RNNs are the state-of-the-art technique for modelling speech recognition and natural language translation \citep{Greff:2015wv}. 


\citet{CrnkovicFriis:2016vx} \circledone{29} train an Long Short-Term Memory (LSTM) on 3D joint positions of a dancer. LSTM is a special type of RNN that is capable of learning high-order dependencies \citep{Hochreiter:1997fq} more effectively than the classic RNN. The authors train an LSTM with 3 hidden layers, each with 1024 neurons and at each frame, the model unrolls in time for 1024 previous frames.  In order to output real-valued motion capture data, a Mixture Density Network (MDN) \citep{Bishop:1994vb} is attached to the output of the LSTM. By using an MDN, the LSTM learns to output a probability density function for each DOF of the joint positions, from which the actual values of the joint positions can be sampled. 

For the generation, one can sample random sequences from a trained LSTM. The sampling is done by providing a short initial sequence to the model, followed by iteratively producing a probability density function, which in turn is used to determine the next pose by the MDN. 


\citet{Fragkiadaki:2015vx} \circledone{27} use RNNs with the addition of an encoder network before the input of the RNN and a decoder network after the output of the RNN. The proposed  architecture is used for the both tasks of modelling the motion capture data for movement generation as well as learning to recognize activities from videos.

The proposed Encoder-Recurrent-Decoder (ERD) architecture, as shown in Fig.~\ref{fig:rnn-erd},  extends the typical RNN architecture by jointly learning  representations of posture using the encoder-decoder networks,  as well as the dynamic qualities of movement using the RNN.  The motivation behind the this architecture is to first learn representations of the input data that would make learning their dynamics easier by the recurrent network. In particular, the authors use fully-connected networks for the encoder and decoder modules,  and two stacked LSTMs each with 1000 hidden units for the recurrent module.


\begin{figure}[t]
\centerline{\includegraphics[width=1.0\linewidth]{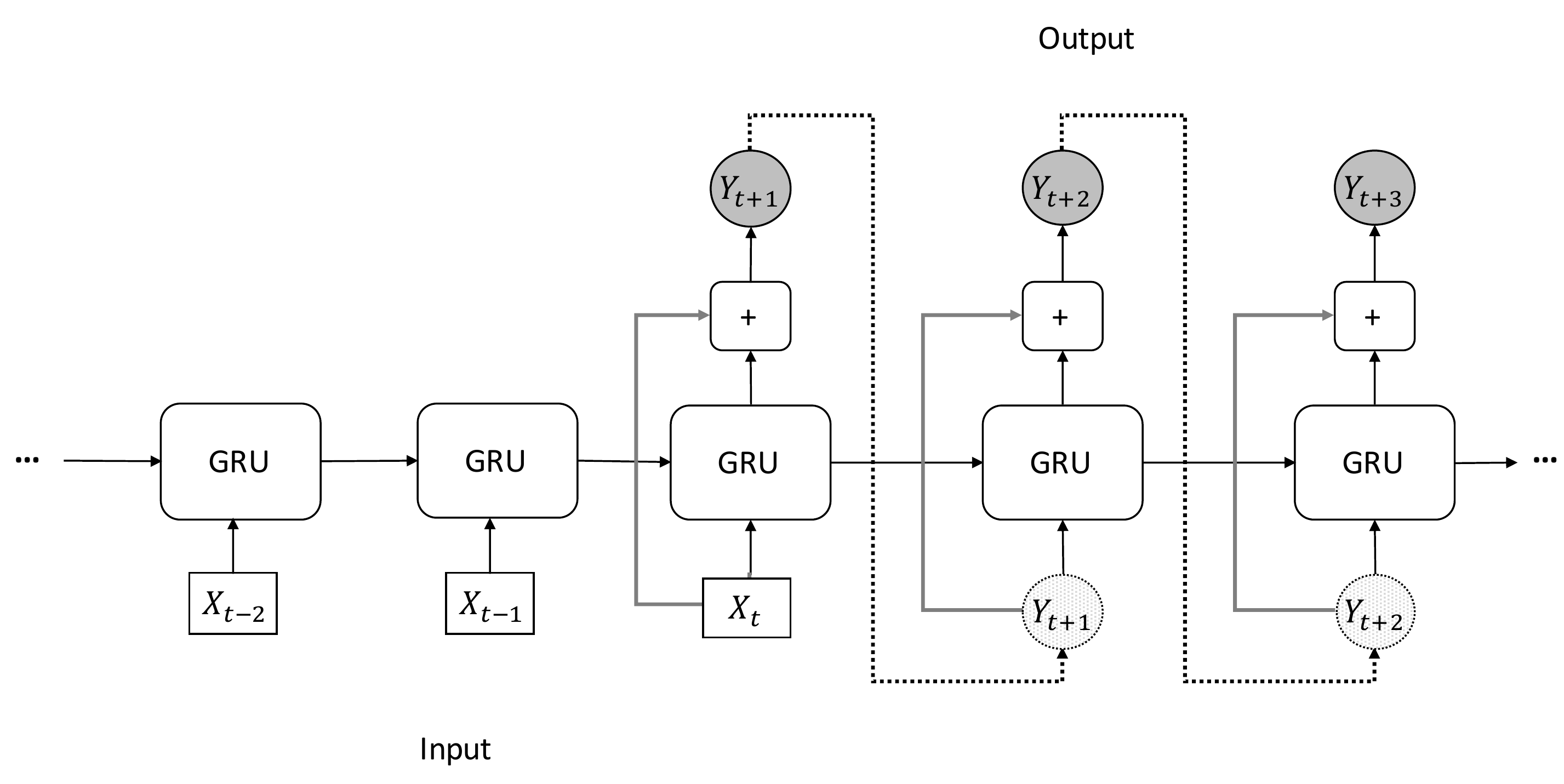}}
\caption{The residual recurrent architecture used by \citet{Martinez:2017ta}.} 
    \label{fig:martinez_arch} 
\end{figure}

\citet{Martinez:2017ta} \circledone{33} train a single-layer Gated Recurrent Unit (GRU) on the complete Human 3.6M dataset. As shown in Figure~\ref{fig:martinez_arch}, the proposed network follows a sequence-to-sequence modelling architecture that consists of an encoder RNN and a decoder RNN that share the same set of weights, as well as residual connections for the output of the decoder network.

This approach differs from similar approaches in two ways. First, although the input of the model contains the joint angle data, the network is trained to predict the joint velocities instead. Using the residual connections on the output, the predicted velocities are then added to the joint angles from the previous frame to calculate the joint angles of the output frame.   The residual connections implicitly push the network to internally model the velocities from the joint rotations and the authors argue that this approach reduces the amount of drifting in the predictions.
Second, the authors point out that since it is shown that RNNs cannot learn to recover from the incorrect predictions that they make if they are only exposed to the ground truth data, they propose a learning approach in which they feed the predictions of the network as the input to the decoder RNN instead of the ground-truth data during the training.

\citet{Wang:2017bs} \circledone{31} propose a model called Sequential Adversarial Auto Encoder (SAAE) based on a sequence-to-sequence architecture and is trained using an adversarial approach. They include the factor information as extra input to both the encoder and decoder networks, enabling the model to learn the relationships between the factor values to the output sequences.



\subsubsection{Convolutional Networks}
\citep{Holden:2016bv} \circledone{30} propose a deep learning framework that combines  convolutional autoencoders  \citep{Vincent:2010vu} with a feed-forward convolutional network for generating movements controlled by high-level parameters. 

First, a representation of movement is learned based on the approach for learning human movement manifolds using convolutional autoencoders \citep{Holden:2015fb}. In this approach, the convolutional autoencoders are trained over a large dataset of movement with the objective of learning a manifold that can be used to reconstruct the movement it is given as input. 
The convolutional network performs a one-dimension convolution over the temporal domain to learn an unsupervised feature map of movement. As a result, 
instead of learning a dynamic model of movement, the network is trained on windows of movement, learning the temporal features the same way it learns the spatial features.  

Next, a feed-forward convolutional neural network is trained based on the representation learned by the convolutional autoencoder.  This network creates a regression model that maps from high-level parameters (factors) to movement in the output pose space. While the autoencoder network is trained over the whole database, this network is trained over a subset of movements that is desired to be generated. Since the network is trained independently from the autoencoder, it allows for learning different control networks for different movement types using the same representation learned by the autoencoder. 

Finally, another convolutional network is trained on top of the feed-forward control network in order to resolve possible ambiguities in the control, such as foot-step timing. The details of this network is specific to the type of the movements being generated.

For generation, the characteristics of the desired movement is fed to the top-most-layer and  the network outputs the movement data. 
As opposed to other approaches reviewed here, which generate movements frame-by-frame, the proposed framework generates a movement segment at once. 

\subsection{Other Techniques}
\label{sec:synt:other}
Other than the aforementioned mainstream machine learning models, a few studies use other techniques to learn and generate movement.

\citet{li_motion_2002} \circledone{3} introduce a technique called \textit{motion texture} for editing motion capture data using a two-level statistical model. This approach overcomes the limited ability of linear systems in capturing highly nonlinear complex movements by introducing a hierarchical approach for modelling non-linearities.  A set of \textit{motion textons}, representing 
movement primitives, is learned using a Linear Dynamic System (LDS) at the bottom level. The top-level model then learns the distribution of the textons using a transition matrix and thus captures the global dynamics of the movement, as an entire sequence. 

 Multilinear Independent Component Analysis (MICA) is a generalization of Independent Component Analysis (ICA) and N-model Singular Value Decomposition (SVD), which models higher order dependencies for each factor. 
\citet{liu_human_2011} \circledone{18} use MICA and decompose the training data into multiple factors. With the assumption that the factors are statistically independent, different states of the factors are arranged in a tensor.  
In this approach,  time-warping is applied on the training data to achieve structurally similar movements. Furthermore, the dimensionality of the data is reduced using PCA.

\section{Evaluation Methods}
\label{sec:eval}

A review of the literature on movement generation systems reveals a lack of emphasis on  evaluation of the system itself, as well as the quality of its output movements. 
The majority of the publications in statistical movement generation provide no formal assessment and rely on the authors’ informal inspection of a small subset of hand-picked movements, generated by the proposed system, and reported via 2D sketches or supplementary videos. Furthermore, only a few works compare their proposed methods with alternative approaches  \citep{Fragkiadaki:2015vx, Martinez:2017ta} \circledone{27} \circledone{33}. 

The systems that control the planning factors of the generated movements (e.g., pointing position, stride length, etc.) use the Root Mean Square Error (RMSE) between the generated data and the target as an indicator of the performance of their systems \citep{herzog_recognition_2009,yamazaki_human_2005} \circledone{13} \circledone{4}. While this evaluation gives a good measure of how accurately the model is able to generate the movements that satisfy the given planning constraints,  they do not measure the believability of the movements. 
For example, a generated  movement for picking up an object might precisely put the hand in the location of the target object, but do so by introducing artifacts in the movement that render the movement unrealistic. 

In another group of studies, the authors quantitatively evaluate the prediction precision of their system \citep{wang_multifactor_2007,taylor_factored_2009, Fragkiadaki:2015vx, Martinez:2017ta} \circledone{9} \circledone{14} \circledone{27} \circledone{33}. A test segment as set aside, and a portion of it is fed as an input to the model, with the task of predicting (generating) the consecutive frames. The precision of the prediction is then quantitatively assessed by using the RMSE of the the generated movement and the ground truth data. As we discuss in more details in the discussion section, using the RMSE to evaluate a movement generation system fails to take into account the stochastic and creative nature of human movement.

A number of works conducted studies involving human subjects assessing the perceptual qualities of the generated and recorded movements  \citep{Chiu:2011do,Chiu:2011uv,tilmanne_stylistic_2012, Alemi:2015kg} \circledone{20} \circledone{21} \circledone{23} \circledone{28}. The main motivation behind these studies is to evaluate the believability of the generated movement, as well as to validate the generation of the intended expressions. 
In such studies, human subjects are presented with one or more movement animations and are asked to either compare them, for example based on which one is more realistic, or to categorize, rate, or rank them based on a given metric such as the valence dimension of affect. 

\section{Summary and Discussion}
\label{sec:discussion}

In this section, we summarize our findings from reviewing the literature on statistical movement generation. First, we look at the types of  movement that are modelled, as well as the dimensions of generated movements that the model can control. We then discuss the limitations and challenges in acquiring training data, followed by summarizing the approaches to learn, generate, and control movement. Finally, we make the case for better evaluation methods for movement generation systems.

\subsection{The Choice of Movements And Scenarios}
\label{sec:disc:move}
Humans are able to perform  a broad range of movements with intricate modulations that come from various factors such as the  physical characteristics of the mover, her or his affective state, intentions, and plans. No computational model is yet capable of learning and generating all types of movements with every possible modulation.  Research on movement generation therefore is done on relatively small and constrained subsets of all possible  movements and scenarios.

Although in many of the reviewed literature it is not clearly stated why certain movements and scenarios were chosen for learning and generation, we point out a number of elements that might play a role in choosing what to model: meeting the demands of a certain application (e.g., the movement repertoire  of a video game character),  the simplicity or complexity of the movement pattern and thus different challenges in modelling them (e.g., modelling walking versus grand pas de chat), or the availability of the training data for certain movements (e.g., there are more training data for walking movements compared to the data available for writing with a pen on paper). In addition, the focus of a group of works is mainly on introducing a new machine learning model and generating movement is used to demonstrate the capabilities of the new model, as in the work of \citet{taylor_modeling_2007}.
In the following, we provide a brief discussion of different aspects of what has been chosen to be generated in the literature and the type of problems that needs to be addressed. 

\textbf{Scenarios:}
Table~\ref{tbl:stylesum} and Table~\ref{tlb:data} provide some insights into the movement types and scenarios that are the subject of the reviewed works. Walking is arguably the most commonly modelled form of movement \citep{wang_learning_2006,taylor_modeling_2007,wang_multifactor_2007,tilmanne_expressive_2010,Kulic:2011hw,tilmanne_stylistic_2012,Tilmanne:2014tx,Alemi:2015kg}  \circledone{7} \circledone{8} \circledone{9} \circledone{16} \circledone{19} \circledone{23} \circledone{26} \circledone{28}.  The prevalence of walking  can be explained by the short and cyclic nature of walking patterns, the large availability of training data, and its application in video games. While most works generate arbitrary walks, a few have addressed the problem of character navigation, which requires the model to provide a way to control the direction of the movement such as minimizing the divergence of the generated path from a target path 
\citep{Holden:2016bv}  \circledone{30}, or continuously adjusting  the heading direction of the character as movements are being generated \citep{Alemi2017_WalkNet}  \circledone{34}.

Arm movements are also commonly modelled as they have various applications in robotics, character animation, and non-verbal communication \citep{herzog_parametric_2008, Taubert:2012gl} \circledone{13} \circledone{22}. 
In reviewing the works on modelling arm movements, we highlight two open problems in modelling generative models of movement.
First, some forms of movement performed mainly by a subsection of the body can be combined with movements in other parts of the body. For example, one can wave her hands while standing, walking, running, or biking. Second, in some applications it might be needed that the movement must satisfy certain constraints. For example, reaching a specific location in space to interact with an object. 

Dance movements, with applications in art installations and video games,  are also explored in a number of works \citep{brand_style_2000,li_motion_2002, qu_motion_2008,CrnkovicFriis:2016vx, Alemi2017_GrooveNet} \circledone{1} \circledone{3} \circledone{12} \circledone{29} \circled{35}. 
Dancing implies  precise timing and positioning rules for different body parts, especially in the case of dancing with a partner. In most cases, the dance is accompanied by music and the choices of movements and their timings are influenced by the music, as well as by the particular choreography of the dance. Addressing these challenges remain open to the research community. 


A few studies consider modelling sports scenarios \citep{wang_key-styling:_2006,qu_motion_2008,Wang:ia,Matsubara:bp}. Similar to the challenges described in modelling arm movements, certain sports movements are often expected to satisfy constraints and interactions with other movers or objects. For example, kicking the ball in its exact position in space or ducking at the right time and position to avoid being hit by an opponent. While none of these problems are addressed in the reviewed literature, they bring interesting challenges in controlling the generated movements that can be made the focus of future research.


\textbf{Diversity of Movements:}
The level of diversity in the training data also plays a role in the choices involved in building generative movement models.  A machine learning model can be trained on a dataset that contains samples from the same form of  movement with no variations across the function, planning, expression, or personal movement signature dimensions. This model learns a specific pattern and generates movements similar to that pattern. On the other hand, models that are trained on a dataset that contains movements that vary across one or more dimensions learn multiple patterns or different modulations of the same pattern. When it comes to generating new movements, only a few of these models provide ways to control the characteristics of the generated movements, which will be  discussed in Section~\ref{sec:disc:control}. 

Creating models that can generate a diverse repertoire of movements has two requirements. First, a machine learning model that has the capacity of learning all such variations (e.g., neural networks versus HMMs), or using a hierarchical architecture that allow breaking down the training task into multiple subtasks. Second, a diverse training dataset.
The recent availability of  large datasets and computational power have allowed training models on a wider variety of movements than before \citep{Martinez:2017ta}. Yet, large datasets such as Human 3.6M are tailored for human activity recognition use cases. They mostly contain variations across the functional dimension, corresponding to everyday movements, and may not contain variations of  the same function in the planning or expressive dimensions needed for many generative applications.

\textbf{Factors Used for Control:}
A group of the reviewed works allow for controlling the characteristics of the generated movements. 
We now summarize and discuss which movement factors in each movement dimension (Section~\ref{sec:char_agent}) are used for control.


\textit{Function:}
Controlling the functional factors of movement allows the user to choose the function (action) and ideally make transitions from one function to another. For example, one can ask the model to generate walking movements, followed by jumping over an obstacle, and and then grabbing an object.
The majority of the systems only model a single function and only a few works address controlling the functional factors of movement \citep{wang_learning_2005} \circledone{5} and \citet{Kulic:2011hw} \circledone{19}. 

The main challenge in modelling the functional factors comes from the broad variations in how different functions are executed (e.g., the differences between walking and shaking hands).  Controlling the functional factors, compared to controlling the planning and expressive factors,  requires employing machine learning models with higher learning capacities that can accommodate the larger repertoire of movement patterns, or designing hierarchical systems that consist of individual models for each function. Another challenge comes from the need to generate transitions from one function to another,  which requires performing so in a plausible manner,  even if samples of such transitions do not exist in the training data.

\textit{Planning}:
The execution of a given movement function can be planned by one or more planning factors. Considering walking as an example, one can plan the walk by specifying its trajectory or by setting the stride length.
In modelling the planning factors, the reviewed works mainly investigate controlling the trajectories of the hands \citep{lin_applying_2008,herzog_parametric_2008,herzog_recognition_2009} \circledone{11} \circledone{13}, or the trajectory of the agent on the ground plane \citep{Holden:2016bv, Alemi2017_WalkNet}.

The challenges in controlling planning factors often come from the need to satisfying the given constraints to a desired level of precision as defined by the plan. Depending on the application, the movement might have to follow an exact trajectory or stop at an exact location in space to follow the plan. In most cases, the plan can be described formally through a set of constraints, which allows calculating the error the agent is making with respect to the given constraints. The movement can then follow the plan by minimizing this error. The minimization can be done through an offline optimization process, as done by \citep{Holden:2016bv}, by learning to perform the movements that cause a reasonably small error, or by designing  a sensorimotor loop for the agent, and feeding back the perceived error the agent has made from the target and adjusting the movements as they are being generated. 



\textit{Expression}: 
A variety of affective expressions can be conveyed by modulating human movement. 
While reviewing the literature, two issues come up in designing generative models that can control the expressive factors of the generated movements. 

First, as opposed to the planning factors, there is no perfect execution pattern for an expression. One can express the same affective state through movements that differ in the way they are executed, which can be influenced by various factors such as the characteristics of the mover or her cultural background. Therefore, unlike the planning factors, one cannot directly measure expressive qualities of a movement,  and use such measurements to control the desired expressive modulation. Consequently, supervised machine learning techniques are the common approaches to control the expressive factors. 

Second, to control the expressive factors, one has  to, explicitly or implicitly, choose a method to describe the affective state or quality. Therefore, systems that support controlling expressive factors differ on their choice of expressive factors and  the way the factors are described. Although the majority of the studies use informally described walking gaits (e.g., chicken walk, drunk walk, etc.) as the expressive factor, more recent studies use categorical \citep{Taubert:2012gl,Samadani:2012ki} \circledone{22} \circledone{25} and dimensional \citep{Alemi:2015kg} \circledone{28} representations of affect as the expressive factor.


\textbf{Modelling Interactions:}
The majority of the studies do not consider modelling the interactions between one agent (mover) and objects or other agents.  Only one study explores modelling the interactions of two or more agents  \citep{Taubert:2012gl} \circledone{22}. 

There are two main challenges in modelling interactions.
First, there is a lack of publicly available motion capture data of agent-agent and agent-object interactions.
Second, interactions with an object or another agent introduce hard constraints that the generated movements have to satisfy, such as the exact timing and positioning of different body parts. Both of the aforementioned challenges have left creating machine learning models for generating interactions a widely open area.

\textbf{Moving Forward:}
To summarize our findings with respect to the current gaps  in the choice of movements and scenarios, we highlight two areas for further consideration by the research community:

\textit{Exploring a wider variety of movement scenarios.} As humans have a large  repertoire of movements, the movements chosen in the literature at the time of this writing only incorporate a small subset of what humans are able to do in various different scenarios. This includes both the scenarios in which only one mover is involved and  scenarios involving agent-agent and agent-object interactions. Each scenario entail different set of challenges, and addressing each challenge contributes to stronger generative models. 

\textit{Work towards better integrations with an agent model.} Controlling the generated movements based on factors that directly map to the internal state of an agent, such as intentions, plans, and affective state makes generative models more suitable for integration into real-world applications such as video games. Using such direct mappings, one can define the desired behavior of an agent and the appropriate movements will be generated. To integrate a generative movement model into an agent model, one needs to address two problems. First, the internal state of the agent needs to be formalized in such a way that it can be translated to movements, across the dimensions of function, planning, expression, and personal movement signature. Second, the generative model should support controlling the movement based on these formalized characteristics.

\subsection{Training Data}
\label{sec:disc:data}
Although the training data is one of the fundamental components of  machine-learning-based solutions, the field of statistical movement generation faces a number of challenges when it comes to finding a desired set of  training data to address a particular research problem. Lack of available training data arguably limits the scope of the problems that can be tackled.

There is a shortage of publicly available motion capture databases. The vast majority of motion capture data are owned by film and video game industries, or are  captured by independent research groups that do not publish them to the public. As mentioned in Section~\ref{sec:data:base}, few publicly available databases are well curated towards particular research questions, and the ad hoc characterization of the movements in the rest of the database makes them less applicable for many research projects. 

New databases can be curated to provide a wider variety of movements to support the problems described in Section~\ref{sec:disc:move}. This includes training data that contain movements with variations across the five movement dimensions to allow for creating models that are able to generate such variations. Moreover, to fully take advantage of the supervised and semi-supervised learning algorithms, the research community needs more annotated databases. Annotations allow for controlling movements based on meaningful factors, and supports creating generative models that integrate well with agent models.
There is also a shortage of training data for scenarios in which two or more agents interact with each other, which is necessary to develop movement generation systems that address inter-agent communications as well as agent-object interactions.

Another challenge in building a large training set is the different skeleton configuration that each database uses. Each database uses a different number of joints and bone proportions. As a result, one needs to re-target the skeletons from multiple datasets to a uniform target skeleton before being able to combine them, which is time and labor intensive.



\subsection{Modular Learning}
While in most reviewed works a single machine learning model is used to learn and generate the entire repertoire of the  movements the system is expected to learn, a generative system might instead utilize a group of machine learning models working in connection with one another.
Making the training process modular works by breaking down the structure of the movement data into smaller components and training different models on different, smaller segments of  movement. 

We describe two ways that one can break down the complexity of movements: 1) following the physical structure of human body, and 2) segmenting the time dimension.

We can group together the moving parts of the body in different ways in the context of motion capture data: all of the body joints together, separating the upper and lower body joints, grouping joints belonging to individual limbs (e.g., right arm, left leg), and finally considering a single body joint. 
A system can be designed to follow such groupings and assign different machine learning models to different parts of the data.  For example, \citet{SukhbaatarMAC11} use a two layer design: at the bottom layer, they train individual models for the right arm, left arm, right leg, left leg, and the trunk at the frame level. On the top layer, a CRBM is then used to coordinate the movements of individual limbs.

Another approach to break down the mocap data is to split the longer sequence of frames into smaller segments, and train each segment by a separate model.  
For example, an approach that is common among HMM-based works is  to split a walking cycle into a few movement primitives (e.g., right leg lifted, …), and use a separate HMM to lean each primitive. During the generation, the output of the HMMs are then concatenate in the right order. 



\subsection{Loss Function}
  In training artificial neural networks, the loss function directs what the model does and does not learn. In the reviewed works, it is common to use the Mean Square Error (MSE)  between the generated  joint rotations or positions and their ground truth counterparts. However, because of  the highly variable nature of movement, using MSE has some implication on the generative and creative performance of the model. Considering the rotations of individual joints in the 3D space, there are many different ways the movement can unfold, resulting in different joints configurations. Therefore, using MSE as the loss function restricts the model to only a single \textit{numerically correct} prediction while there might be many more \textit{perceptually correct} predictions that might result in large MSEs.
Further research is needed to devise more effective loss functions that consider the natural variabilities in human movement.

\subsection{Modelling the Time Dimension}
Movement unfolds through time and likewise, motion capture data is in the form of sequences of frames. The generative models are trained on these sequences and are expected to create new sequences, directly or indirectly.
In the reviewed literature, modelling sequences is handled differently depending on what machine learning model is used. Notably, we can refer to flattening the time dimension, as done in most of the dimensionality-reduction-based models, sequences as conditional inputs as in CRBM and FCRBM, recurrent connections in RNNs, and convolution over time in CNNs. 
Further investigation is needed to compare the pros and cons of each technique. 
The approach to modelling sequences also has an impact on the memory and time complexities of both the learning and generation algorithms. As a result, the length of the sequences that can be modelled is limited by available computational resources.

\subsection{Generation Algorithms}

The mechanics of the generation algorithms depend on the machine learning model. An algorithm that is suitable for one group of models may not work for others. As a result, it is not easy to compare the algorithms with on another. In the following, we summarize algorithms used in the literature for each family machine learning models.

Dimensionality reduction (DR) techniques generate movements by choosing a point in the DR space, and projecting the data from the DR space back into the mocap space. This process is often followed up by post-processing procedures to re-sample the data into a sequence of frames as the time dimension is often flattened in DR models.

HMM-based models use the   Viterbi algorithm \citep{Rabiner:1989hs} to sample form the distribution learned by the HMM. In cases where more than one HMM is used for different portions of the movement, first the desired order of the multiple HMMs is determined manually or by sampling from another model. Next, mocap frames are sampling from each HMM and concatenated to create the final sequence.

A group of neural networks including feed-forward nets, CRBM, FCRBM, and RNNS, use iterative sampling. Such models are trained to predict the next frame from an input sequence of previous frames. By iteratively performing this sampling operation, while shifting the input sequence to include the newly predicted frames, a sequence is generated. 

To sample from Convolutional Neural Networks (CNNs), one has to reverse the flow of information in the network. While during the training the flow of information is from the mocap data to the labels, during generation the network is fed with a set of desired values for the labels and the connections are followed back to reach the input layer of the network which represents the generated mocap data. Since the CNNs use convolution over time, a whole sequence is generated at once, as opposed to the iterative sampling of other neural networks.

\subsection{Control Techniques}
\label{sec:disc:control}
One of the challenges in statistical movement generation is controlling the qualities of the generated movements. As described in Section~\ref{sec:charac}, the term \textit{factor} refers to the sources of influence on movement and we call the domain of possible values for each factor  the \textit{factor space}. 
In Section~\ref{sec:disc:move}, we discussed what factors are used to control the movement in the literature. In the following, we summarize the key techniques that are used to implement the control mechanisms.

\textbf{Supervised versus Unsupervised Learning}. 
Most works use a supervised learning approach, in which the model learns a mapping between the input data and the labels. 
Supervised approaches have the benefit of allowing the researchers to explicitly convey to the model what factors they want to be controlled. However, the performance of the model depends on the quality and the balance of the labels. Another challenge in using supervised techniques is the lack of annotated mocap datasets as mentioned in  Section~\ref{sec:disc:data}.

 In unsupervised techniques, the factors of variation in the training data are determined automatically, without any prior knowledge of what they might correspond to semantically.  The result is often a low-dimensional representation of movement that could be interpreted by means of experimentation,  e.g., \citep{brand_style_2000,wang_multifactor_2007} \circledone{1} \circledone{9}. 

Using unsupervised methods eliminates the need for labeled datasets. However,  the discovered factors are not defined by the researchers and highly depend on the variations that exist in the training data, which may or may not directly correspond to semantically meaningful factors. While this can make it more challenging to design a system for a particular application that requires controlling certain factors, such unsupervised methods can be used as pre-trained models for creating supervised models.

\textbf{Individual Models}. 
The simplest way to create a control mechanism is to train a separate model for each point in the factor space and switch between models to control the generation. Each model is trained only on the data that correspond to that particular point, thus only imitating the same factor value. 
For example, one can train a model on a set of training data containing only walking movements, and train another model on a set of training data containing only jumping movements. The former model will only generate walking movement while the latter will only generate jumping movements.

Some studies combine the outputs of the individual models to create movements for other points in the factor space. 
For instance, \citet{herzog_parametric_2008} \circledone{13} and \citet{herzog_recognition_2009} \circledone{13} train individual Hidden Markov Models (HMMs) for learning pointing movements that vary in the position the hand points at.
To generate a movement that points to a target position, a set of local HMMs with aiming positions closer to the target position are selected. Next, the parameters of a new HMM for the target position are determined by interpolating the chosen HMMs, and a new output is generated from the newly constructed HMM. 

Similarly, \citet{Tilmanne:2014tx} \circledone{26} train a Hidden Semi-Markov Model (HSMM) for each variation that exists in their training dataset. To generate movements for a given new point in the factor space, the model parameters of the individual HSMMs are interpolated or extrapolated. 

In another study, \citet{tilmanne_stylistic_2012} \circledone{23}   train an HSMM on a large set of neutral walking sequences and use a linear regression transformation technique to adapt the parameters of the HSMM to a particular walking style. The adaptation algorithm uses the data from a small set of walks with that particular style.


Other models in this category learn one model per factor state, e.g., \citep{tilmanne_expressive_2010} \circledone{16}, or learn a range within the factor space using the same model, e.g., \citep{taylor_modeling_2007} \circledone{8}, but provide no method for controlling the generated movements. 

\textbf{Parametric Probability Distributions}. 
One way to learn factor variations is to use a parametric probability distribution to model the data. In  a parametric probability distribution, the mean of the distribution is a function of the factor(s). As a result, the value of the factor influences the mean of the distribution and thus controls the characteristics of the movements sampled from the distribution. 

This method is commonly used among the HMM-based studies, e.g., \citep{herzog_parametric_2008,herzog_recognition_2009,yamazaki_human_2005} \circledone{13} \circledone{4}. \citet{wang_key-styling:_2006} \circledone{6} use Self-Organizing Mixture Network (SOMN) of parametric Gaussians to create a probabilistic mapping from the factor space to the pose space. 
 In another approach, \citet{wang_learning_2006} \circledone{7} use parametric Gaussians to build Stylistic DTGs \citep{song_2003}.

\textbf{Labels as Extra Model Input}.

Another technique to control what the model generates based on given labels is to feed the labels as extra inputs to the model alongside the training data. In this way, the model learns the correlations between the labels and the training data.
During the generation, we can set the label inputs to our desired values and perform the sampling procedures as usual to control the generated data, e.g., \citep{Wang:2017bs, Taylor_PhD_2009}.

\textbf{Built-in Support for Control}.
A machine learning model can be designed in a way that provides a mechanism for a factor variable to control the characteristics of the generated movements through its internal connections. 
For instance, Factored Conditional Restricted Boltzmann Machine (FCRBM) uses a context variable that controls the behaviour of the network through gated connection between the observations and the latent variables \citep{taylor_factored_2009}  \circledone{14}. 

\citet{Holden:2016bv} \circledone{30} use a feed-forward convolutional neural network dedicated to controlling the behaviour of another machine learning model that is trained on movement data. The control network learns a regression model from high-level parameters (factors) to the hidden layer of the main machine learning model. In this approach, a different control network is trained for different applications (e.g., controlling the walking direction versus controlling the affect), while  the same main network is reused.


\subsection{Machine Learning Family}

As presented in Section~\ref{sec:synthesis}, different families of machine learning models such as dimensionality reduction (DR), Gaussian Processes (GP), Hidden Markov Models (HMMs), and Artificial Neural Networks (ANNs), are used to learn and generate movement.
While we point out the strengths and limitations of each family, we acknowledge that further investigations are needed to discuss which types of movements each machine learning model is capable of learning and generating. For example, while most reviewed works that are based on DR techniques model walking movements, one needs to apply the same approaches to other types of movement for comparison. 
This would be challenging since it is not always easy to replicate the approaches described in the literature.

DR techniques have two limitations in learning human movement. 
First, DR  models map static poses to a DR space and do not explicitly consider the dynamics of movement. This is overcame by using a dynamical model such as LDS  \citep{qu_motion_2008} \circledone{12} to model the temporal characteristics. 
Second, DR techniques rely on pre-processing steps such as sequence alignments and fixed-length representation of the data, which could require extensive manual labor or limit the variety of movements that can be modelled beyond short and cyclic movements such as locomotion \citep{tilmanne_expressive_2010,Samadani:2012ki} \circledone{16} \circledone{25}.  

The GPLVM family (Section~\ref{sec:synth:gp}) are known for their abilities to generalize well from a relatively small training data, which is demonstrated by \citet{wang_multifactor_2007} \circledone{9}. This capability makes them suitable for applications in which one is interested in generating movements that there is not pre-existing recordings for them. 
However, GPLVMs are not inherently dynamic models and do not capture the temporal structures of the data. This is overcome by integrating them with a dynamic model such as HMM \citep{Taubert:2012gl} \circledone{22}, or by the introduction of dynamic GP models such as GPDM \citep{Wang:ia} \circledone{10} and MF-GPM \citep{wang_multifactor_2007} \circledone{9}.  
Another limitation of the GP-based models is that they are computationally expensive to train and draw samples from, and they require maintaining the complete training dataset for generation, which make them less efficient for realtime and interactive applications.

HMMs (Section~\ref{sec:synth:hmm}) were designed to learn and generate temporal data such as human speech.
Unlike GP-based models, they do not require retaining the training data, and their generation algorithms can be used in real-time applications. 
However, the learning and expressive capacity of HMMs is limited. 
To keep the computational cost manageable, most HMMs are trained with the assumption that the data follows a first-order Markovian dependency, meaning that the next frame of data only depends on the current frame. 
While the first-order assumption might be sufficient in modelling short and cyclic movements such as walking, it has limitations in modelling more complex movements.  One way that to overcome this limitation, as applied in the literature, is to train a group of individual HMMs to capture different movement primitives or different factor variations as demonstrated in most HMM-based studies \citep{yamazaki_human_2005,herzog_recognition_2009,tilmanne_stylistic_2012} \circledone{4} \circledone{13} \circledone{23}.  However, it is still possible to learn longer movements with a single HMM as demonstrated by \citet{brand_style_2000} \circledone{1}.

There are a variety of artificial neural networks (Section~\ref{sec:synth:ann}) used for movement generation, each with different characteristics and applications.
Shallow, perceptron-based neural networks can only learn basic movements with few degrees-of-freedom.  However, the more complex versions of neural networks that are discussed here provide  more expressive power than HMMs. 
While RBM-based models are successfully applied in generating mocap data, the need for layer-wise training rather than back propagation makes it harder to train deeper RBM-based networks.  On the other hand, over the past few years Recurrent Neural Networks (RNNs) and Convolutional Neural Networks (CNNs)  have shown promising results in learning long-term dependencies well beyond a first-order Markovian assumption in sequential data. It is easier to train deep RNN or CNN  networks compared to RBMs. These properties make them suitable for learning movements that in nature have longer-term dependencies and might follow complex hierarchical characteristics. For example, while a walking cycle can be modelled with a single-order Markov process, many dance pieces contain phrasings that need to be defined over longer windows of frames rather than a single past frame.

\subsection{Evaluations}
\label{sec:disc:eval}

As presented in Section~\ref{sec:eval}, a major gap in the field of automatic movement generation is the lack of a widely-accepted evaluation procedure for the proposed systems and most of the studies merely rely on the informal inspection of their authors. 

Evaluation of movement generation systems is a challenging task. 
First, movement generation is highly function-dependent. 
Each movement generation system models only a subset of possible human movements, such as walking and running, or picking up an object. Such a model is therefore only capable of generating  movements functionally similar to the ones it has seen in the training data and can be expected to be evaluated for those movements only. For example, model that can successfully learn and generate walking movements may not perform well in learning and generating more complex movements such as dancing. This makes it difficult to compare alternative approaches to movement generation if they do not model the same set of movement functions.

Second, each system targets a specific application and has to be evaluated towards meeting the specifications of that application. For example, a model aiming for real-time generation has to be evaluated towards its space and time complexities, whereas an offline generation system might prefer better quality over faster generation. As another example, a system that supports controlled generation has to be evaluated based on its control abilities. As a result, not all systems can be evaluated in the same way. 

Third, movement generation is a creative task, which requires a different evaluation approach than rational problem solving tasks.
Although a movement generation system can be evaluated based on memorizing and regenerating the movement in the training data, thus using perfect recall as a measure of evaluation, one can evaluate the system based on its creativity and generalization capabilities. For example, evaluating the quality of the system’s output in generating movements that do not exist in the training set.

In this section, we discuss the challenges we face in evaluating movement generation systems. We then highlight the lack of comparisons between alternative approaches and make the case for building a stronger community for movement generation and its role in evaluations. 





\textbf{Evaluation of Generative Systems}
For any generative system, there are two dimensions that can be evaluated: 1) the performance  of the software or the algorithm, and 2) the system's quality at its generative task (output).
These evaluations can be exploratory, to identify any issues with the system or to determine the characteristics that can be measured in later evaluations, or they can be descriptive, to asses the quality of the system according to some standards, metrics, or requirements.  
 
For evaluating the software and the algorithmic aspects of a generative system, we are mainly  interested in the computational and memory (time and space) complexities. One can therefore use the common space and time complexity analysis to evaluate the performance of a system. This becomes more important if the system aims at performing real-time generation. 

Validating the creative quality of the system can be difficult. First, as opposed to rational problem solving, creative tasks are those for which there is no such a thing as a well-defined preference relation or utility measure. 
This is where most attempts at evaluating movement generation systems presented here fall short while using the mean squared error (MSE) between the generated movements and the ground truth data. Using MSE implies that there exist a single correct prediction and that is the one closest to the ground truth. However, in human movement, there exist many possible poses that can proceed from a given sequence of previous poses. This results in a \textit{space} of correct and plausible poses rather than a single correct \textit{instance}. 

It is worth mentioning that one can evaluate a movement generation systems’ short-term and long-term prediction errors using MSE between the generated movements and the ground-truth data as a proxy to ensure that the model continues to generate plausible movements over time and the output does not drift to  implausible and unrealistic movements.

Another aspect of evaluating generative systems is to browse the variety of possible inputs and observe the system's output to make sure the system produces the correct output.
For example, if we ask the model to point at a particular location in space, we are interested in making sure the system does generate movements that satisfy the given conditions for a variety of inputs. While this can easily be done if the conditions are quantified (such as locations in space), it is more challenging if we are evaluating in terms of more qualitative factors such as emotional states. 

For the qualitative aspects of generative systems, evaluation studies involving human subjects are designed and conducted. In such studies, one has to consider that creative tasks can be subjective and the cultural biases, as well as  any individual judgments in the evaluation of the outputs of a generative system as to be taken into account in analyzing the responses from human subjects. 

\textbf{Replicability}
The majority of the reviewed systems are difficult to replicate. First, the training data for many studies are not provided to the public. Second, only a few studies have published the source codes for their experiments. As a result, only a few studies have provided a comparison between their approach and the alternative ones. 

Objective comparisons between alternative approaches based on well-defined metrics, tasks, and applications can speed up the research and facilitate the innovations in any field.  Some fields such as computer vision take advantage of  the availability of widely accepted datasets that come with well-defined tasks and evaluation methods, such as the ImageNet \citep{Deng:jn} dataset and the ImageNet Large Scale Visual Recognition Challenge. While the field of movement generation lacks such datasets and challenges, the Movement and Computing (MOCO) community\footnote{\url{https://www.movementcomputing.org/}} which has been established in recent years can be a place for setting up such datasets and challenges by a group of interdisciplinary researchers and artists.

\section{Conclusion}
\label{sec:conc}

With the increasing demand for dynamic and interactive contents across various media, the need for automatic content generation is more apparent and movement animation is no exception. Meanwhile, the recent advancement in the field of machine learning and the promising results in the domains of audio, vision, and text, machine learning has shown to be a prominent choice for learning generative models of spatio-temporal data. 

In this paper, we provided a  review of the body of literature on using machine learning techniques and  motion capture data for the purpose of movement animation generation. 
We argue that advances in this field lead to a variety of applications in the video game and film industries, as well as in art practices by providing a less expensive, faster, and more flexible way to create movement animation content both in offline and interactive scenarios.

We point out a number of gaps in each aspect of the reviewed systems. Above all, we raise the need for high quality training datasets with diverse and well-curated contents that serve particular research questions. The availability of public-domain datasets, in conjunction with the rapid progress of the field of machine learning, will pave the way to create more powerful movement generation systems.

The works reviewed here have been published in a variety of different communities, depending on the field where the focus of the contributions were. 
Studies with focus on computer animation side of the research have been published in conferences and journals such as SIGGRAPH and ACM Transactions on Computer Graphics. Studies that focus on the affect-expressive movements are published in IEEE Transactions on Affective Computing and International Conference on Affective Computing and Intelligent Inter- action (ACII).
Studies that contribute to the field of machine learning and use motion capture data have been published in machine learning venues such as international conference on machine learning and IEEE Transactions on Pattern Analysis and Machine Intelligence. 
While some fields such as computer music or computer graphics take advantage of strong specialized communities, ISMIR\footnote{\url{http://www.ismir.net/}} and ACM SIGGRAPH\footnote{\url{http://www.siggraph.org}} respectively, a specialized community for human movement and computation has only recently been emerged through Movement and Computing community (MOCO)\footnote{\url{http://movementcomputing.org/}}. 

\acks{This work was funded by the Social Sciences and Humanities Research Council of Canada (SSHRC) through the Moving Stories Project, as well as the Natural Sciences and Engineering Research Council of Canada (NSERC).}

\vspace{1cm}

\bibliographystyle{acmtog}

\bibliography{bib/movement,bib/movement_papers,bib/extrabib}

\end{document}